\documentclass{article}

\usepackage[final]{corl_2026} %
\usepackage{booktabs}
\usepackage{tabularx}

\usepackage[dvipsnames]{xcolor}
\usepackage{multirow}    
\usepackage{booktabs}    
\usepackage{graphicx}    
\usepackage{makecell}    
\usepackage[percent]{overpic}
\usepackage{pifont}       
\usepackage{array}        
\usepackage{rotating}     
\usepackage{subcaption} 
\usepackage{bm}
\usepackage{amsmath}
\usepackage{afterpage}
\newcommand{\cmark}{\textcolor{green!55!black}{\ding{51}}}
\newcommand{\xmark}{\textcolor{red!75!black}{\ding{55}}}
\newcommand{\nmark}{\textcolor{gray!70}{--}}
\newcommand{\rot}[1]{%
  \makebox[1.2em][l]{\rotatebox[origin=lB]{75}{\textbf{#1}}}%
}
\newcommand{\cmarkm}[1]{\cmark\,\raisebox{0.1ex}{\scriptsize\textsf{#1}}}

\newcommand{\srreach}{\ensuremath{\mathrm{SR}_{\mathrm{reach}}}}
\newcommand{\srinsert}{\ensuremath{\mathrm{SR}_{\mathrm{insert}}}}
\newcommand{\pizero}{\ensuremath{\pi_{0.5}}}
\newcommand{\hybrid}{\textsuperscript{H}} %

\title{WireCraft: A Simulation Benchmark  \\for Industrial DLO Manipulation}

\author{
  \textbf{Chongyu Zhu}\textsuperscript{1}\thanks{Equal contribution.}\thanks{Corresponding author: \texttt{cyzhu@mie.utoronto.ca}} \quad
  \textbf{Ramy ElMallah}\textsuperscript{1}\footnotemark[1] \quad
  \textbf{Hyegang Kim}\textsuperscript{1} \quad
  \textbf{Zachary Tang}\textsuperscript{1} \quad
  \textbf{Jiachen Rao}\textsuperscript{1} \\[0.3em]
  \textbf{Artem Arutyunov}\textsuperscript{1} \quad
  \textbf{Seungyeon Ha}\textsuperscript{2} \quad
  \textbf{Chi-Guhn Lee}\textsuperscript{1} \\[0.8em]
  {\normalfont\textsuperscript{1}Department of Mechanical and Industrial Engineering, University of Toronto, Canada} \\
  {\normalfont\textsuperscript{2}CREFLE Inc., Seongnam, Republic of Korea}
}

\hypersetup{
  pdftitle={WireCraft: A Simulation Benchmark for Industrial DLO Manipulation},
  pdfauthor={Chongyu Zhu, Ramy ElMallah, Hyegang Kim, Zachary Tang, Jiachen Rao, Artem Arutyunov, Seungyeon Ha, Chi-Guhn Lee}
}

\begin{document}
\maketitle

\newcommand{\chongyu}[1]{\textbf{\textcolor{Plum}{Chongyu: #1}}}
\newcommand{\ramy}[1]{\textbf{\textcolor{RoyalBlue}{Ramy: #1}}}
\newcommand{\hyegang}[1]{\textbf{\textcolor{ForestGreen}{Hyegang: #1}}}
\newcommand{\zach}[1]{\textbf{\textcolor{DarkOrchid}{Zach: #1}}}
\newcommand{\artem}[1]{\textbf{\textcolor{BrickRed}{Artem: #1}}}
\newcommand{\jiachen}[1]{\textbf{\textcolor{TealBlue}{Jiachen: #1}}}

\begin{abstract} Deformable Linear Objects (DLOs), such as wires and cables, are central to industrial assembly. Unlike rigid objects, whose state is captured by a 6-DoF pose, DLOs have an infinite-dimensional configuration space and deform continuously under contact with grippers, fixtures, and the workspace, making them a demanding benchmark for general dexterous manipulation. Despite their importance, policy development and comparison remain difficult: existing benchmarks are often tied to specific hardware setups, lack modular and customizable task assets, or study generic deformable-object tasks without the fixtures relevant to real-world industrial wire manipulation. Few benchmarks align simulation, real-world data, and shared evaluation protocols. To bridge this gap, we introduce \textbf{WireCraft}, a simulation benchmark for industrial DLO manipulation with configurable difficulty and assets, spanning three task families: connector insertion, clip routing, and channel seating. It supports two complementary DLO physics models, articulated and deformable, and the released trajectories come from both simulation and a physical UR5. We benchmark reinforcement learning (RL), imitation learning (IL), and vision-language-action (VLA) policies under shared metrics. Privileged state-based RL solves a representative setting in each task family with over 82\% success, confirming the tasks are well-posed. Image-based policies remain less successful across the evaluated tasks, with ACT and DiT VLA falling below privileged State PPO. On a physical UR5, supplementing real demonstrations with simulated data improves success by 15--65 percentage points in the best-performing mixture for each task. These results indicate that industrial DLO manipulation, though tractable under privileged state, remains an open challenge for current vision-based learning.
Our project is open-sourced at \href{https://github.com/DORLR/WireCraft}{https://github.com/DORLR/WireCraft}.

\end{abstract}

\section{Introduction}
\label{sec:intro}

Manipulating Deformable Linear Objects (DLOs)~\cite{zhu2022challenges}, particularly wires, is important for industrial wire-harness assembly~\cite{zhang2024harnessing} and cable management~\cite{wilson2023cable, luo2024multistage}. These settings require the robot to do more than grasp and move a flexible object: it must interact with structured fixtures such as clips, connectors, sockets, and channels~\cite{luo2024multistage,  9732654} while controlling the DLO's deformation~\cite{yu2022shape}. Unlike rigid objects, whose state can be captured by their 3D position and orientation, DLOs possess an infinite-dimensional configuration space and deform continuously~\cite{gu2026survey, zhang2021deformable, Li_2024}. Forces applied at the gripper propagate along the body, causing simultaneous deformation and motion across the DLO~\cite{cao2024shape}. DLOs further exhibit complex contact dynamics, self-occlusion, and deformation uncertainty arising from variation in material stiffness and damping~\cite{cao2024shape, yan2020self}. These properties make policy learning difficult. Compounding the challenge, existing efforts often tie simulation, fixtures, and sensing to specific hardware setups, leaving industrial DLO tasks without reusable simulation benchmarks or shared data interfaces.

\begin{figure*}[t]
    \centering

    \includegraphics[
        width=1\textwidth
    ]{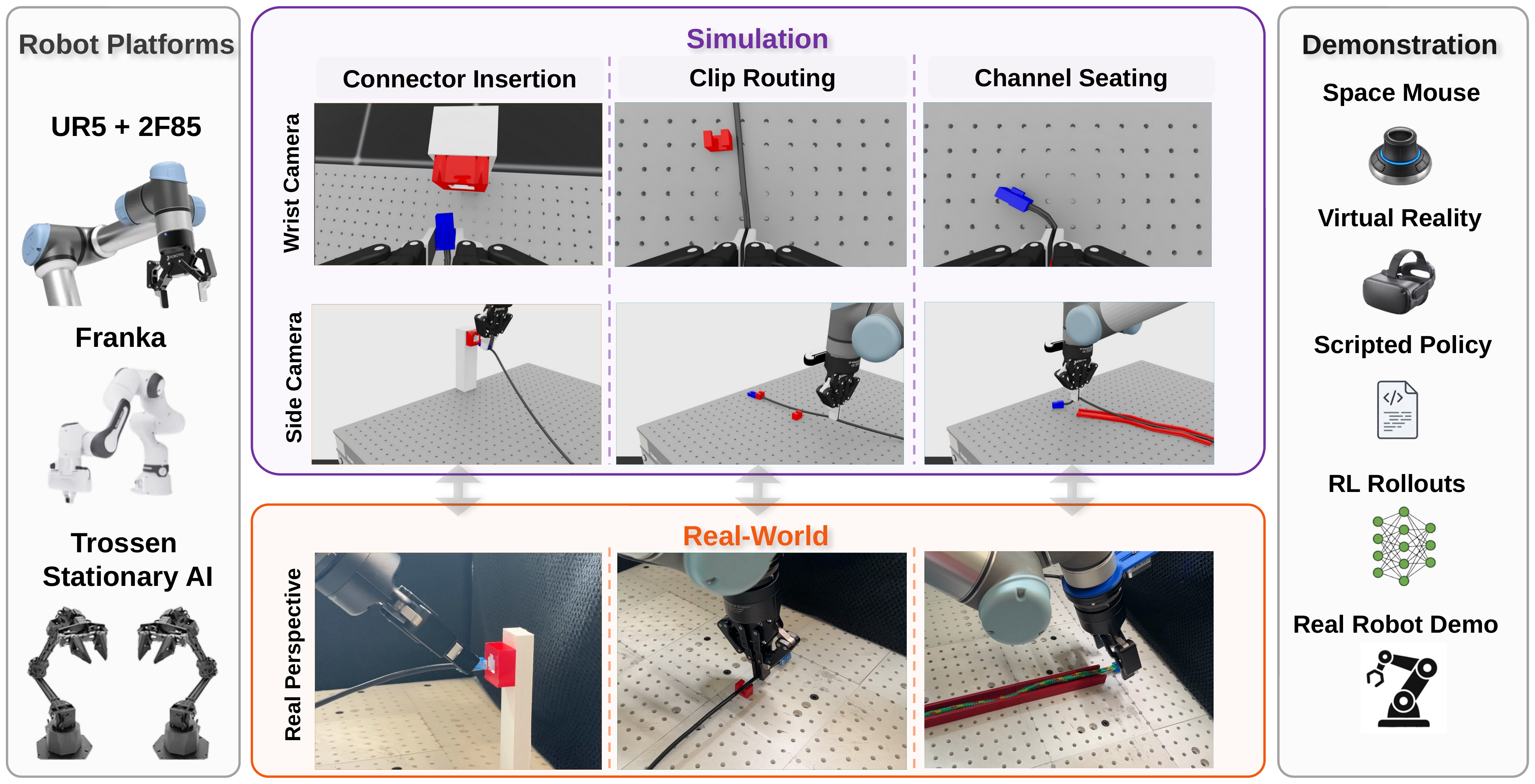}
    \caption{
    \textbf{WireCraft} simulates industrial DLO manipulation across three task families: connector insertion, clip routing, and channel seating with trajectories from multiple sources. Tasks run on the UR5, Franka, and Trossen Stationary AI in simulation, with real-world validation on a UR5.
    }
    \label{fig:wirecraft_ai_pipeline}
\end{figure*}

General robot learning benchmarks such as RLBench~\cite{james2020rlbench} and ManiSkill3~\cite{taomaniskill3} are not centered on DLO manipulation and provide limited DLO-specific data. Deformable-object benchmarks such as SoftGym~\cite{lin2021softgym}, DeformableRavens~\cite{seita2021learning}, and DaXBench~\cite{chendaxbench} include DLOs alongside other deformable categories, while DLO-dedicated environments such as ReForm~\cite{9561766} and DLO-Lab~\cite{cao2026dlolab} target canonical DLO behaviors like shape control and coiling. However, both lack structured industrial fixtures such as connectors, clips, and channels central to assembly tasks. A parallel line of work studies these industrial tasks on physical testbeds~\cite{zhang2024harnessing, wilson2023cable, luo2024multistage, li2025hierarchicaldloroutingreinforcement, tanureza2025industrial, kienle2025ai}, but each effort is tied to specific robots, fixtures, cable geometries, and sensing setups, limiting reuse for new model training or sim-to-real research.
Physical task boards such as NIST~\cite{nist_assembly_task_boards} and the ManipulationNet Cable Routing Benchmark~\cite{chen2026manipulationnet} standardize real-world evaluation but offer no simulation environment. WireCraft combines configurable insertion, routing, and seating tasks with shared policy interfaces and trajectory data from simulation and physical robots.

To address this gap, we make the following contributions:
\begin{itemize}
    \item 
    We introduce WireCraft, a unified simulation benchmark for industrial DLO manipulation across connector insertion, clip routing, and channel seating, with configurable task components, standardized policy interfaces, and two complementary DLO physics models trading off speed and fidelity.

    \item
    We evaluate representative RL, IL, and VLA policies under a shared evaluation protocol, providing a consistent reference for future work on industrial DLO learning. Simulation baseline evaluation and physical UR5 validation cover all three task families.

    \item
    We open-source the benchmark with available trajectory data, data-generation tools, and 3D-printable task-board components. This enables researchers to reproduce the setups, collect demonstrations, and study sim-to-real transfer for industrial DLO manipulation.
\end{itemize}

\section{Related Work}

\textbf{Robot learning and deformable-object benchmarks.}
General benchmarks have standardized rigid-object manipulation. RLBench~\cite{james2020rlbench} provides tabletop tasks with motion-planner-generated demonstrations, while ManiSkill3~\cite{taomaniskill3} extends this to contact-rich manipulation across 12 task categories. However, neither is centered on DLOs, and neither models fixture-constrained DLO assembly. A second line of work targets deformable-object manipulation: SoftGym~\cite{lin2021softgym} provides environments for ropes, cloth, and fluids; PlasticineLab~\cite{huang2021plasticinelab} contributes a differentiable soft-body benchmark for deformation; DaXBench~\cite{chendaxbench} unifies rope, cloth, and liquid manipulation under differentiable physics; and DeformableRavens~\cite{seita2021learning} benchmarks goal-conditioned rearrangement of deformables. MoDeSuite~\cite{zhang2026modesuite} introduces mobile-manipulation deformable tasks in Isaac~Sim. These benchmarks have driven progress on shape control and free-space rearrangement, but their DLO tasks are not industrially grounded. 

\textbf{DLO-specific simulators and learning environments.}
A more focused thread targets DLO modeling and control. ReForm~\cite{9561766} introduces an early sandbox of DLO shape-control tasks. Chen et al.~\cite{chen2023dermujoco} integrate Discrete Elastic Rods (DER) \cite{bergou2008discrete} into MuJoCo, targeting dynamic DLO simulation rather than benchmarked task suites. DexDLO~\cite{sun2024dexdlo} studies goal-conditioned dexterous manipulation of DLOs. DEFORM~\cite{chen2024deform} and its extension DEFT~\cite{chen2025deft} provide differentiable DER \cite{bergou2008discrete} models aimed at perception/control rather than assembly benchmarks. Govoni et al.~\cite{govoni2025dlo} analyze DLO models under domain randomization across Isaac~Sim and Gazebo, focusing on parameter sensitivity. The closest concurrent benchmark is DLO-Lab~\cite{cao2026dlolab}, a differentiable Genesis/Taichi-based simulator paired with DLO tasks and multiple policy-learning baselines. However, DLO-Lab emphasizes generic DLO shape control and material modeling rather than industrial tasks. WireCraft differs by organizing tasks around industrial task families, offering dual wire representations and policy-learning-ready datasets.

\textbf{Industrial DLO manipulation and physical task boards.}
A growing body of applied work tackles the three task families WireCraft unifies. For \textit{clip routing}, prior work learns motion primitives for guiding cables through clips~\cite{wilson2023cable, luo2024multistage}, proposes spatial representations with fixture-aware planning~\cite{9732654, 11127451}, and combines VLM-based planning with RL for long-horizon tasks~\cite{li2025hierarchicaldloroutingreinforcement}. For \emph{connector insertion}, methods generalize across materials and novel connectors via visual flexibility cues or meta-RL~\cite{Li_2024,zhao2022offline}, and optimize cable following, search, and mating~\cite{kienle2025ai, Cao2026RoboticCF}. For \emph{channel seating and constrained placement}, prior work has addressed clamp engagement~\cite{zhang2024harnessing}, plugging/clipping/routing in constrained environments~\cite{tanureza2025industrial,zhou2020dlocableharness}, dual-arm planning under constraints~\cite{yu2025generalizable}, simulated cable laying~\cite{10.1145/3776942.3777006}, and DLO manipulation in benchmark task spaces~\cite{9926677}. Physical task boards provide standardized real-world evaluation: the NIST Assembly Task Boards~\cite{nist_assembly_task_boards,kimble2022performance} and ManipulationNet~\cite{chen2026manipulationnet} standardize protocols for deformable-part assembly, MOTORCYCLE~1.0~\cite{azulay2025motorcycle} automates bimanual cable routing on a reconfigurable board, and Intrinsic's AIC provides a 3D-printable cable insertion board with Gazebo simulation~\cite{intrinsic2026aic}. However, each system is tied to specific robots, fixtures, sensing, or protocols. WireCraft unifies all three task families under shared simulation, industrially grounded assets, and multi-source demonstrations, making it, to our knowledge, the first unified industrial DLO benchmark.

\textbf{Learning algorithms for contact-rich and deformable manipulation.} Recent progress in robot learning has produced several policy classes relevant to contact-rich manipulation, including on-policy and off-policy reinforcement learning~\cite{schulman2017ppo,vecerik2017leveraging}, action chunking~\cite{Zhaoetal2023act}, diffusion-based visuomotor policies~\cite{chi2023diffusion}, transformer-based diffusion backbones~\cite{Peeblesetal2023dit,hou2025dita}, and VLA language-conditioned action prediction~\cite{black2025pi05}. Subsequent work has also advanced deformable-object manipulation through geometry-aware RL and equivariant graph policies~\cite{hoang2025geometry} and through model-based RL with differentiable multiphysics simulation gradients~\cite{xing2025stabilizing}. These methods motivate the need for benchmarks that expose contact-rich deformable manipulation under shared task definitions and evaluation metrics. To this end, we incorporate representative policy families across RL, IL, and VLAs as evaluation baselines rather than exhaustively optimized solvers, establishing WireCraft as a foundation for future specialized algorithms.

\section{WireCraft Infrastructure}

\begin{figure*}[t]
    \centering
    \includegraphics[
        width=1\textwidth,
    ]{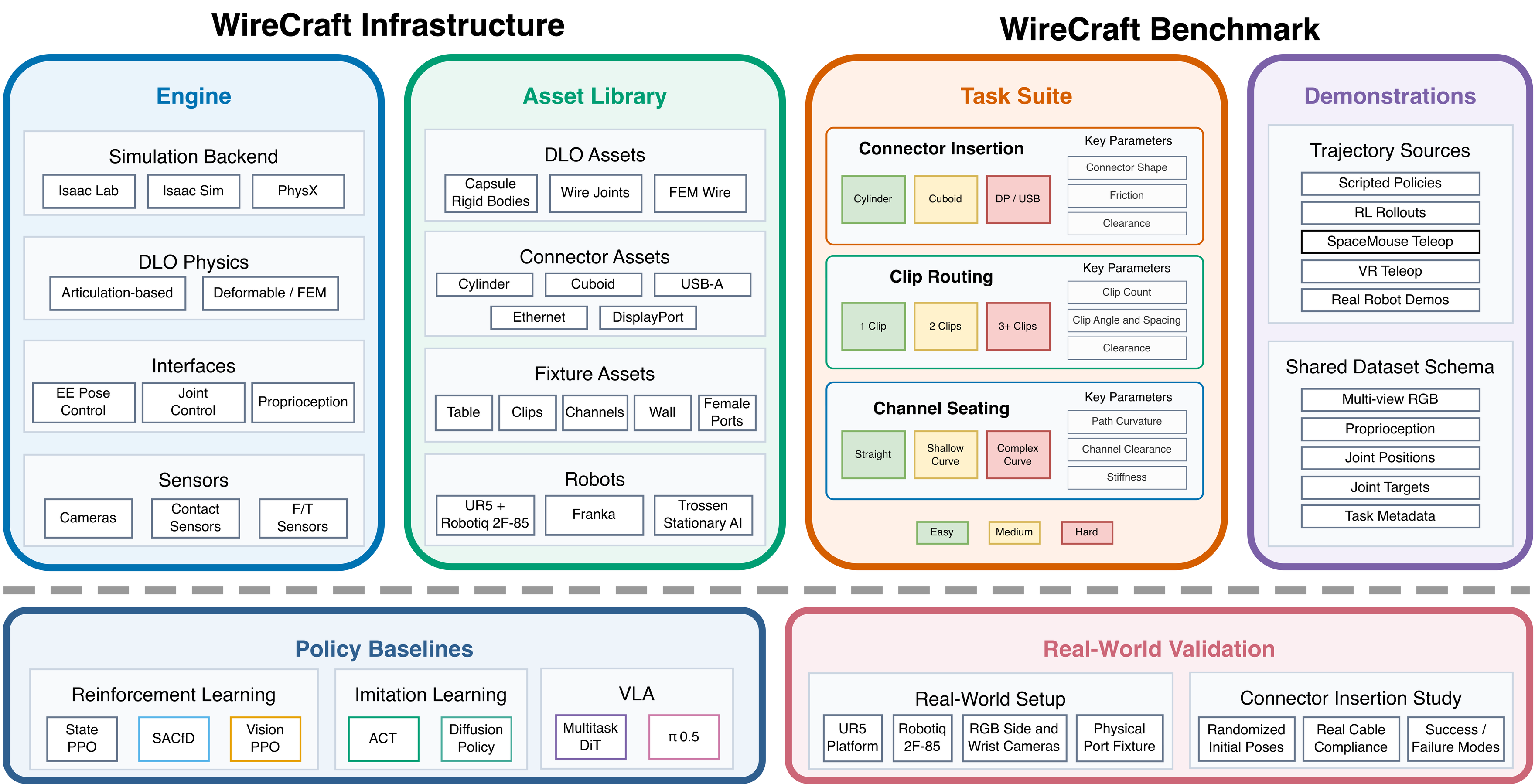}

    \caption{
    \textbf{WireCraft Architecture}, including the WireCraft simulation engine, asset library, task suite, demonstration sources, policy baselines, and real-world validation setup. 
    WireCraft supports articulated and FEM-based deformable wire models, provides configurable industrial DLO manipulation tasks across connector insertion, clip routing, and channel seating, and pairs simulation data with real-world teleoperation trajectories under a shared dataset schema. 
    }
    \label{fig:wirecraft_architecture}
\end{figure*}

\subsection{WireCraft Engine}
\label{subsection:wirecraft engine}

\textbf{Simulation Backend.}
WireCraft is built on Isaac Lab~v2.2.1 \cite{mittal2025isaaclab}, an open-source robot learning framework on top of Isaac Sim~4.5 \cite{isaacsim}; full version and physics details are in Appendix~\ref{appendix:physics}. Isaac Sim provides photo-realistic scenes and fast simulation, while Isaac Lab supports GPU-accelerated, multi-instance training for scalable policy learning and trajectory generation.

\textbf{DLO Physics.}
WireCraft simulates its two wire representations through complementary physics mechanisms. The articulated wire uses rigid-body dynamics: each segment is a rigid body, and neighboring segments are coupled by 6-DoF joints (PhysX D6). This supports high-throughput trajectory generation due to efficient rigid-body contact but can become unstable under adjacent-segment collision or accumulated contact forces. The deformable wire uses Isaac Sim's deformable body API and PhysX FEM deformable body simulation~\cite{isaaclab_deformable_object,physx_soft_bodies}, improving local bending, compression, and contact deformation with higher fidelity, but at higher computational cost. We treat the two wire types as complementary benchmark representations. Detailed wire-model parameters and throughput measurements are reported in Appendix~\ref{appendix:physics}, and a wire--connector physics visualization is shown in Figure~\ref{fig:appendix_wire_with_port}. Details of DLO and scene interaction are provided in Appendix~\ref{appendix:dlo_connector_interaction}.

\textbf{Sensors.} Beyond standard camera RGB data, WireCraft provides built-in interfaces to optional depth channels. To capture physical interactions, the platform integrates contact sensors for precise collision detection and force sensors to monitor grasping force magnitudes and insertion states. Furthermore, real-time robot proprioception, including joint positions and velocities, is natively supported. Observation inputs are selected per baseline, and users can add other sensor types supported by the NVIDIA Isaac ecosystem.

\subsection{WireCraft Assets}
\label{subsection:wirecraft assets}

\begin{figure*}[h]
    \centering
    \newlength{\figgap}
    \setlength{\figgap}{2pt}
    \setlength{\tabcolsep}{0pt}
    \renewcommand{\arraystretch}{0.85}

    \makebox[\textwidth][c]{%
    \begin{minipage}[t]{0.42\textwidth}
        \vspace{0pt}
        \centering
        \begin{tabular}{@{}c@{\hspace{\figgap}}c@{\hspace{\figgap}}c@{}}
            \begin{overpic}[width=0.327\linewidth, trim=100 100 100 100, clip]{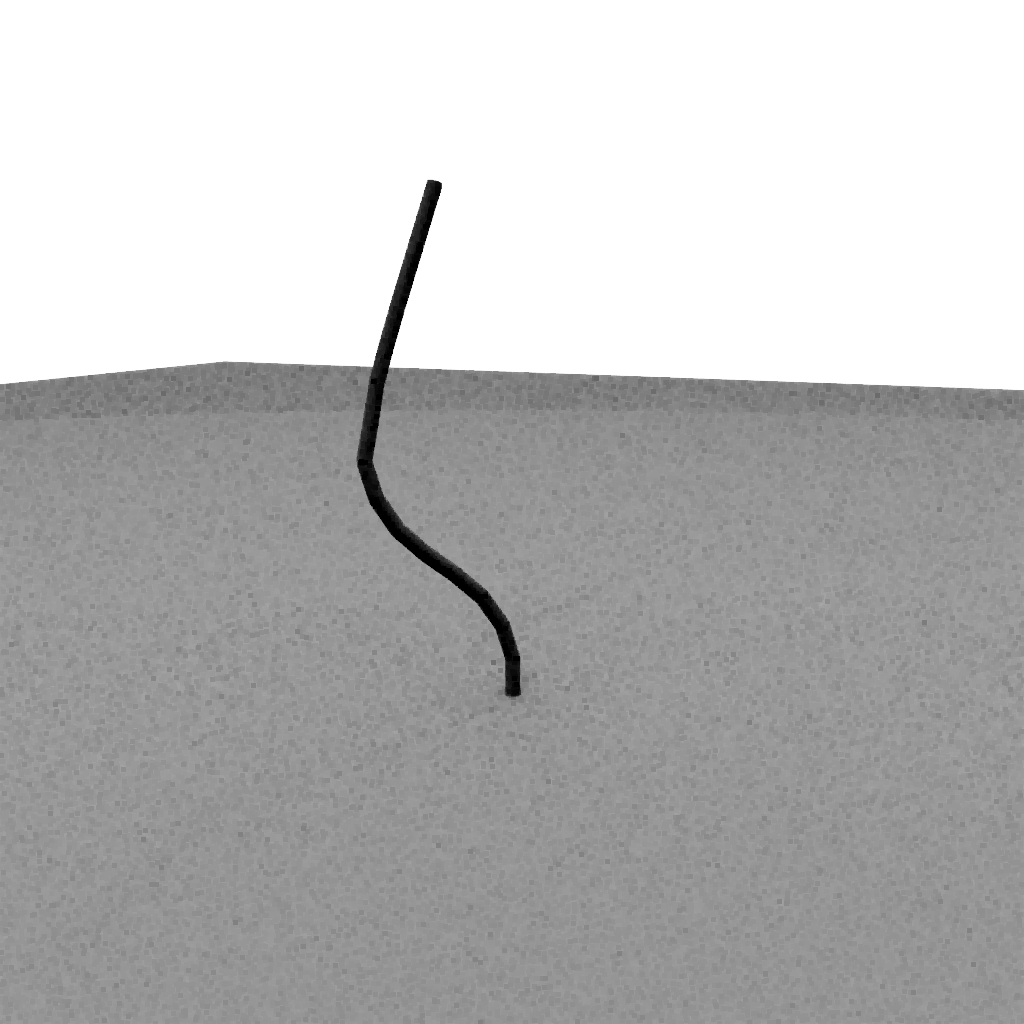}
                \put(4,4){\tiny\textbf{Articulated}}
            \end{overpic} &
            \includegraphics[width=0.327\linewidth, trim=100 100 100 100, clip]{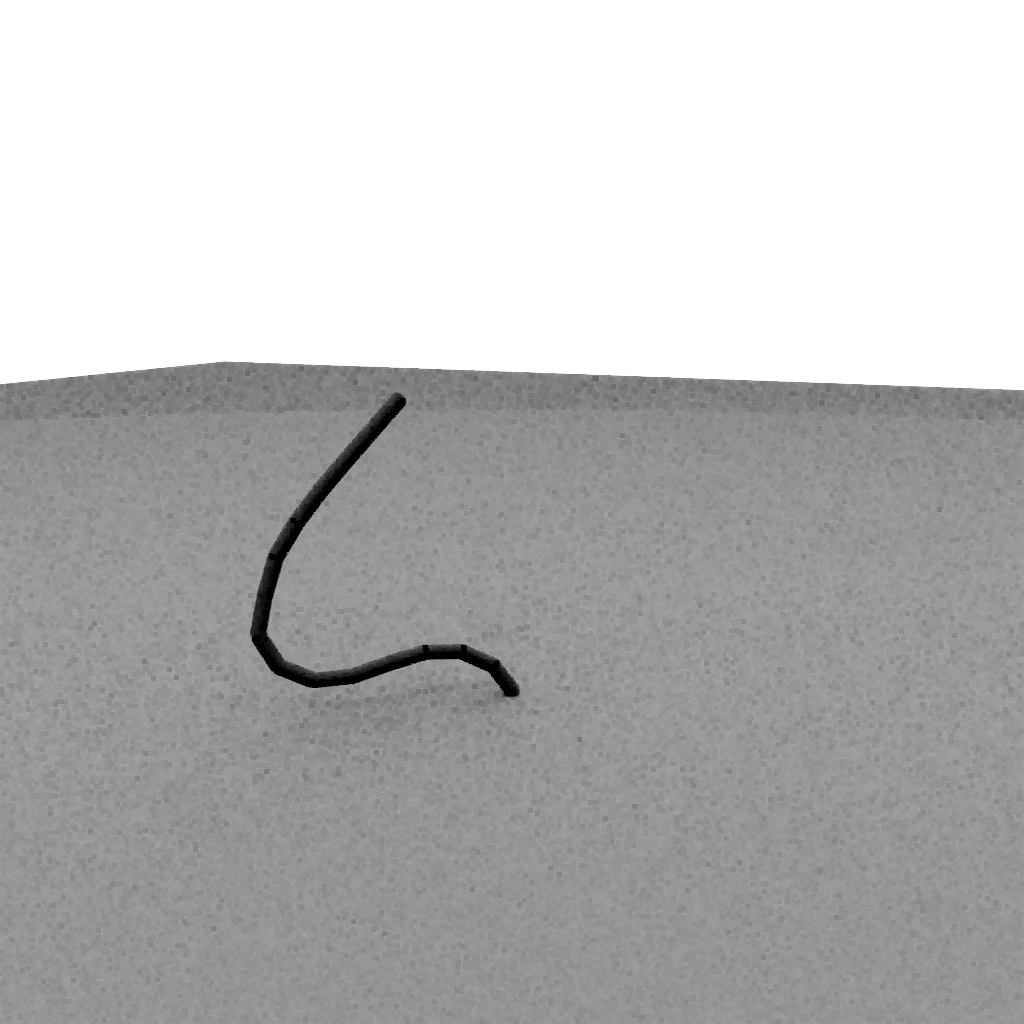} &
            \includegraphics[width=0.327\linewidth, trim=100 100 100 100, clip]{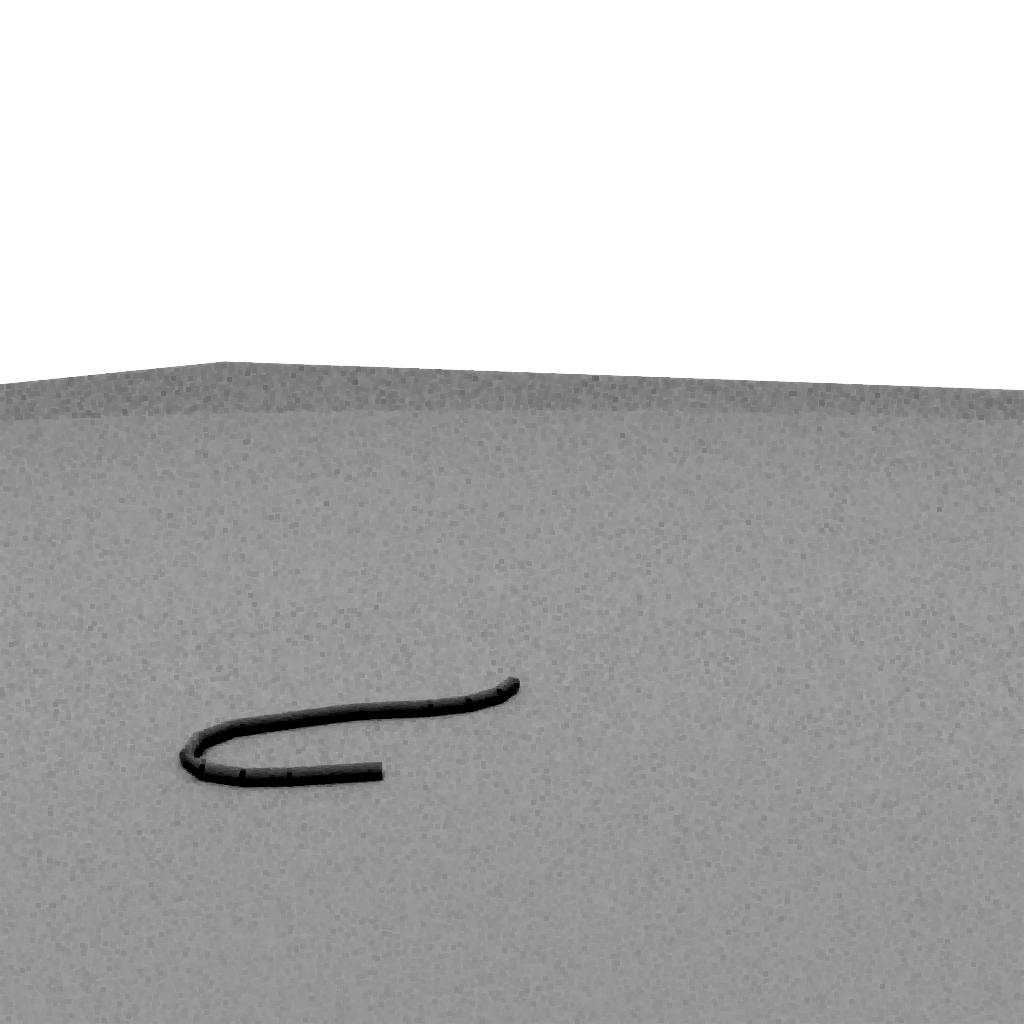}
            \\[\figgap]
            \begin{overpic}[width=0.327\linewidth, trim=100 100 100 100, clip]{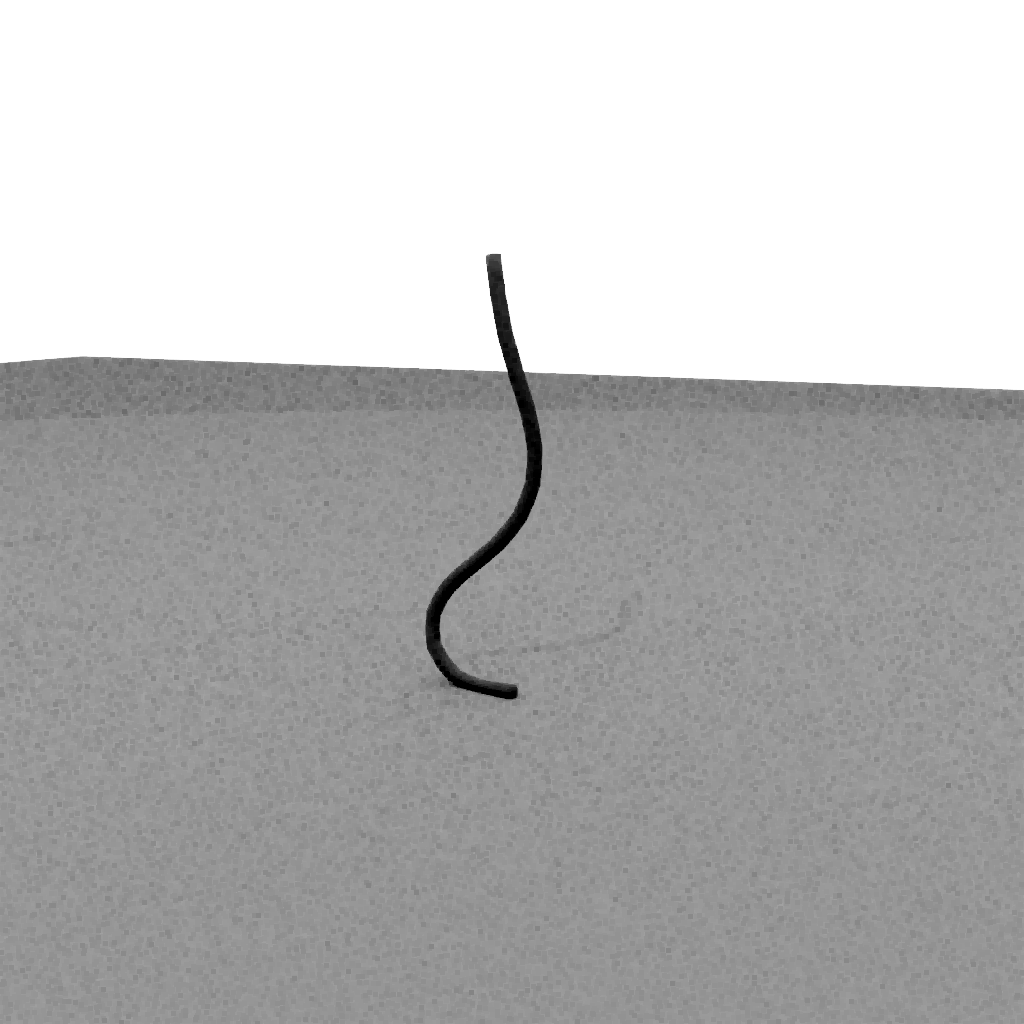}
                \put(4,4){\tiny\textbf{Deformable}}
            \end{overpic} &
            \includegraphics[width=0.327\linewidth, trim=100 100 100 100, clip]{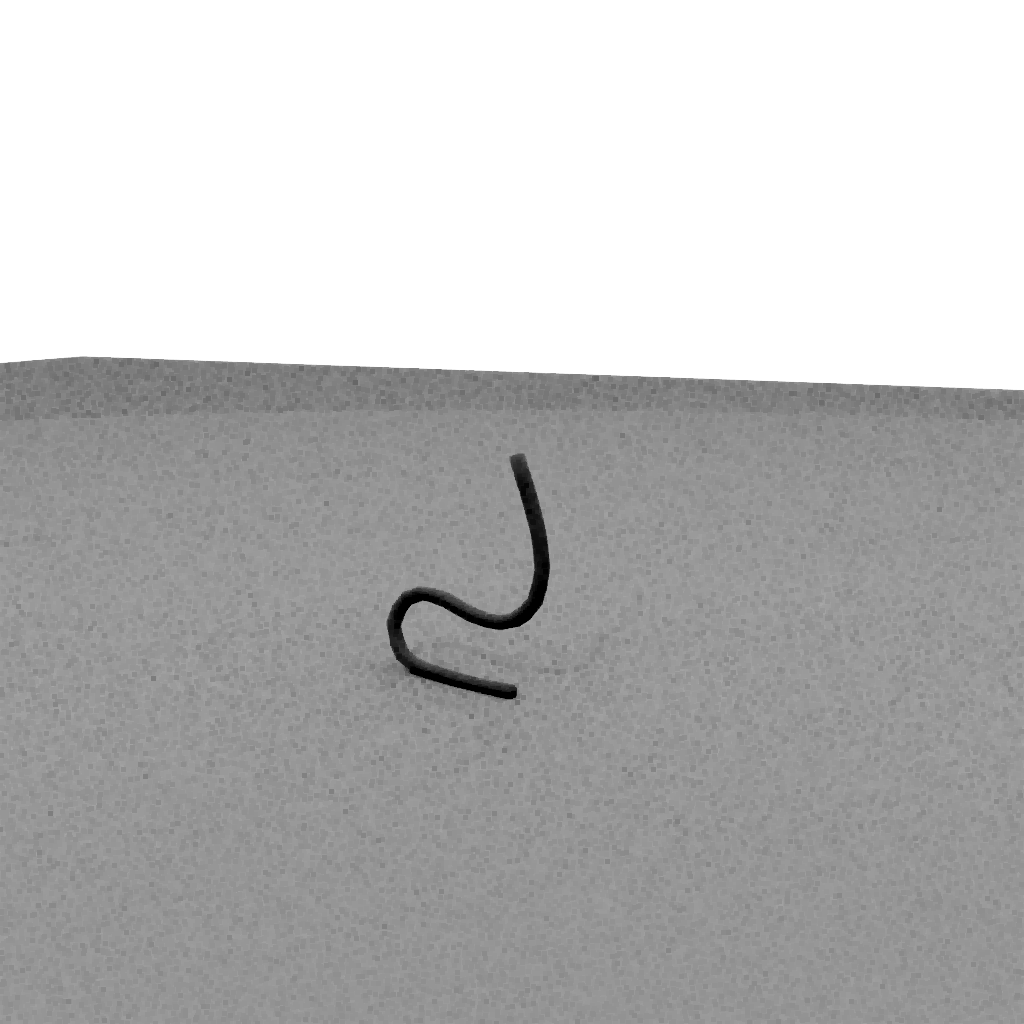} &
            \includegraphics[width=0.327\linewidth, trim=100 100 100 100, clip]{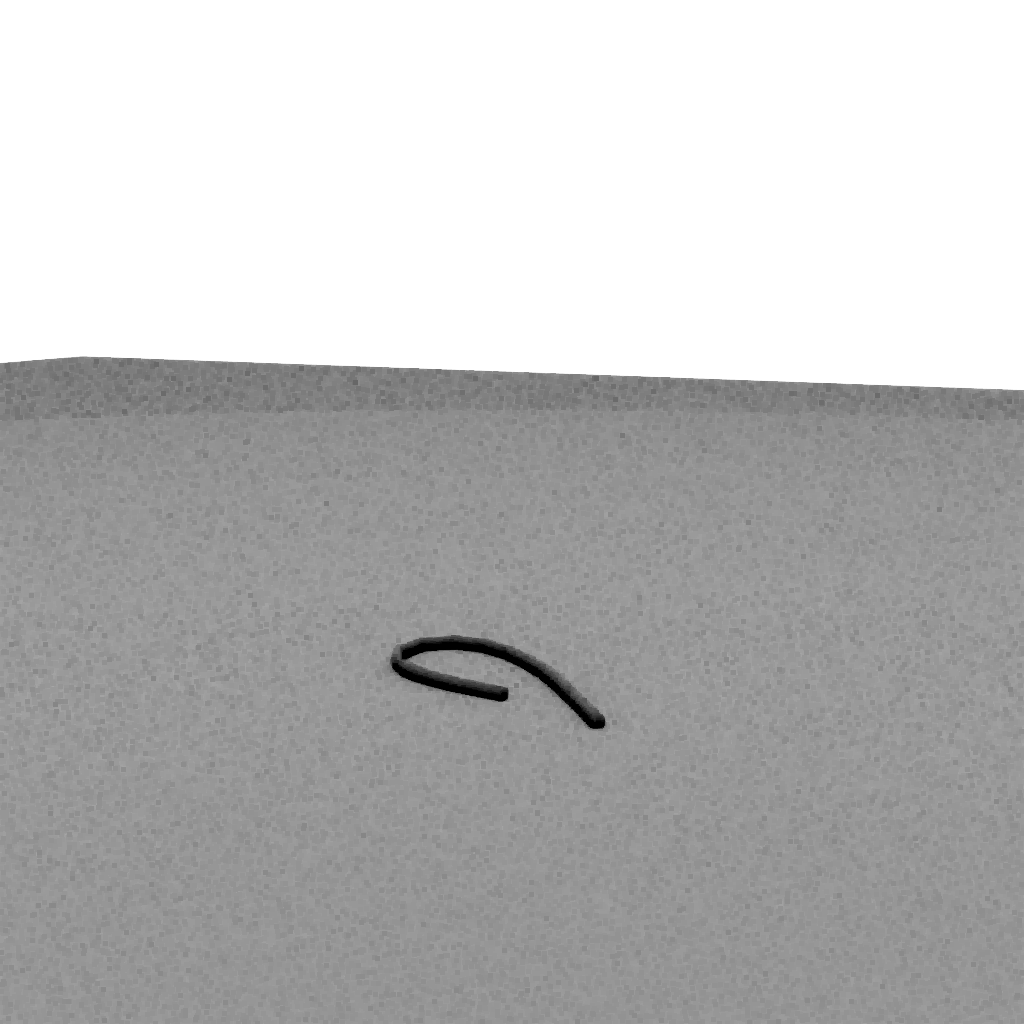}
        \end{tabular}
    \end{minipage}%
    \hspace{\figgap}%
    \begin{minipage}[t]{0.284\textwidth}
        \vspace{2pt}
        \centering
        \begin{overpic}[width=\linewidth, viewport=400 000 1300 900, clip]{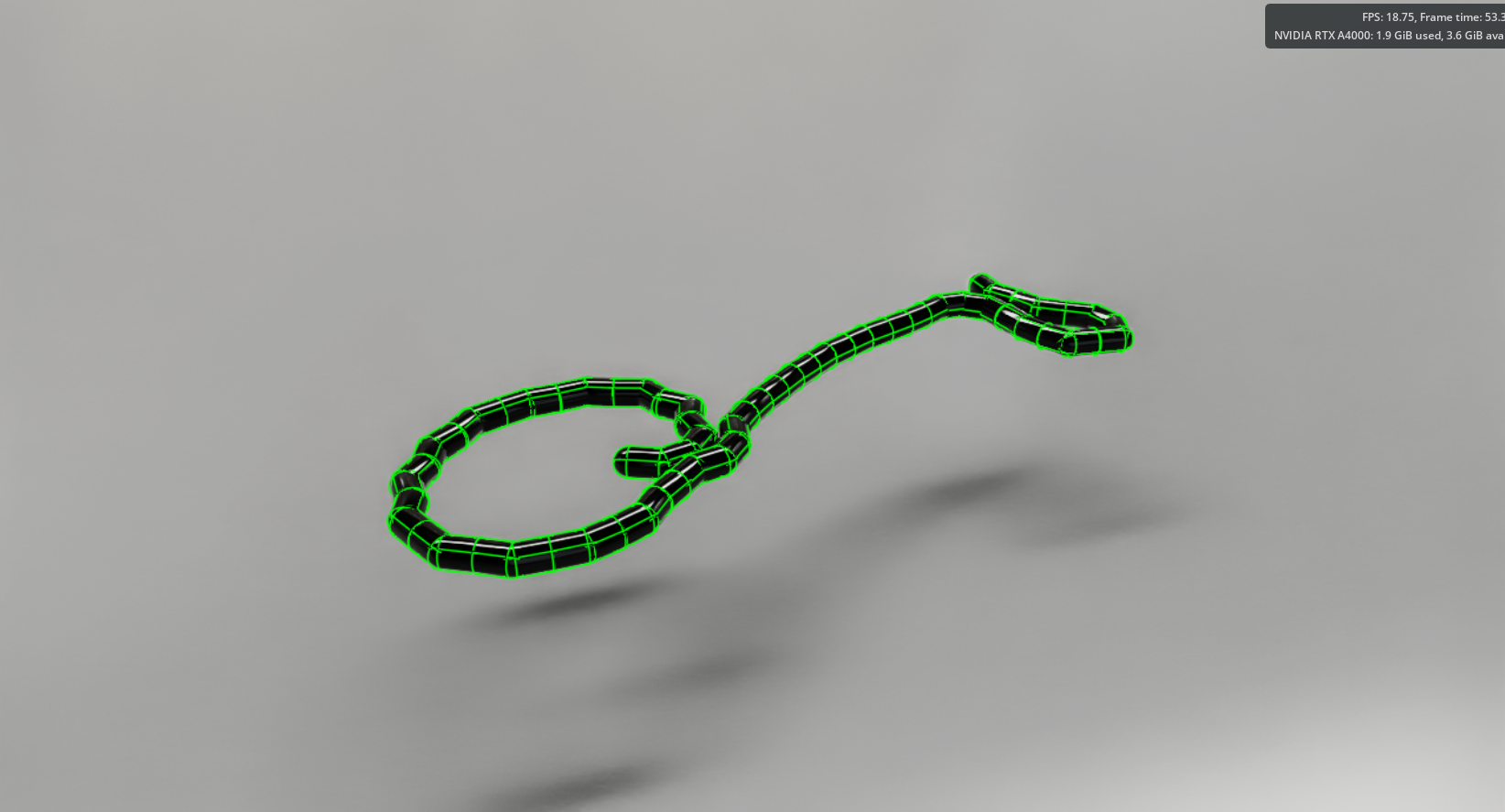}
        \put(4,4){\tiny\textbf{Articulated Rigid-body Chain}}
        \end{overpic}
    \end{minipage}%
    \hspace{\figgap}%
    \begin{minipage}[t]{0.284\textwidth}
        \vspace{2.8pt}
        \centering
        \begin{overpic}[width=\linewidth, viewport=400 100 1100 790, clip]{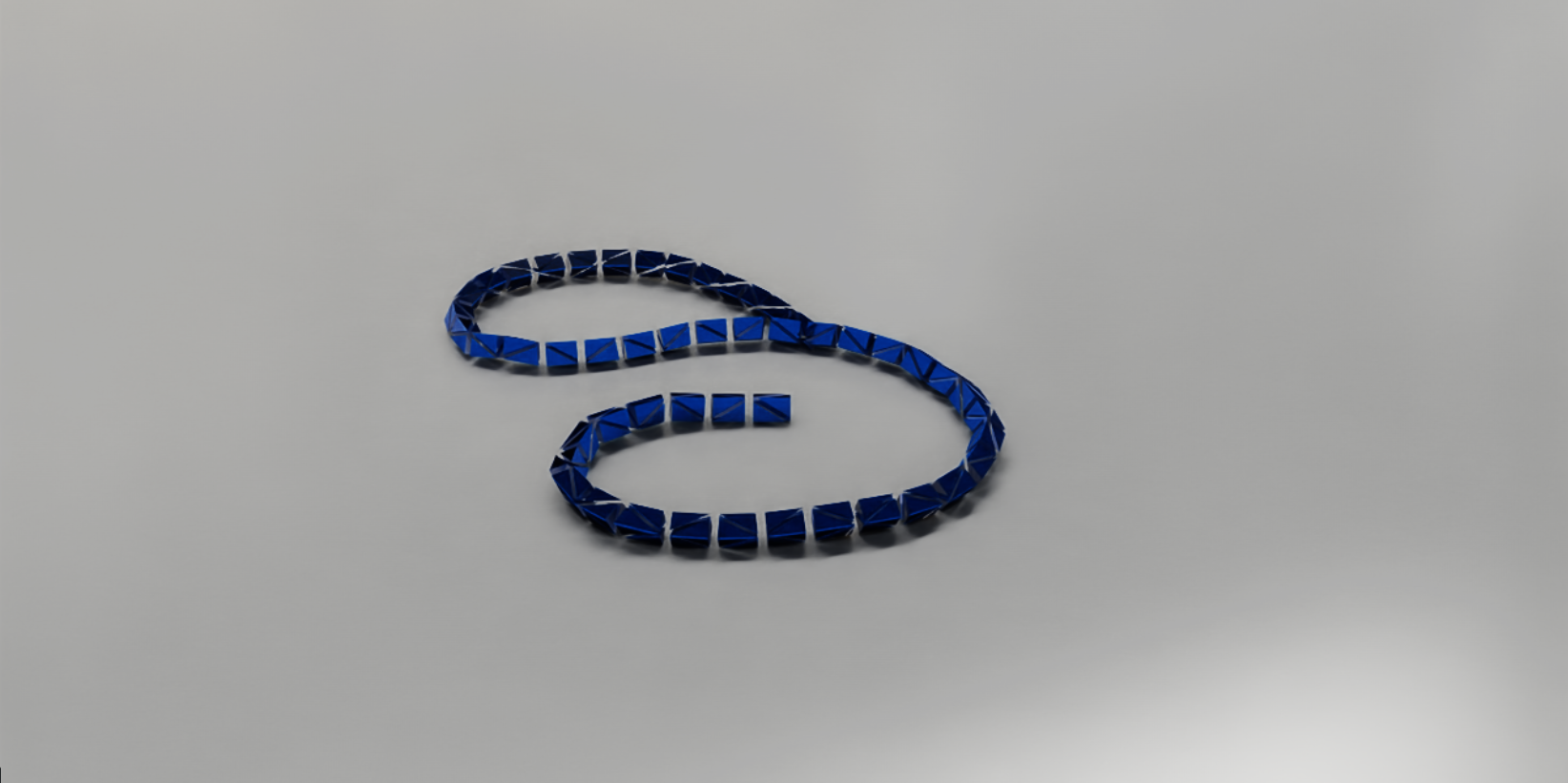}
            \put(4,4){\tiny\textbf{FEM-based Deformable Wire}}
        \end{overpic}
    \end{minipage}%
    }

    \vspace{2pt}

    \makebox[\textwidth][c]{%
    \begin{tabular}{@{}c@{\hspace{\figgap}}c@{\hspace{\figgap}}c@{\hspace{\figgap}}c@{\hspace{\figgap}}c@{\hspace{\figgap}}c@{\hspace{\figgap}}c@{}}
        \begin{overpic}[width=0.139\textwidth, trim=100 100 100 150, clip]{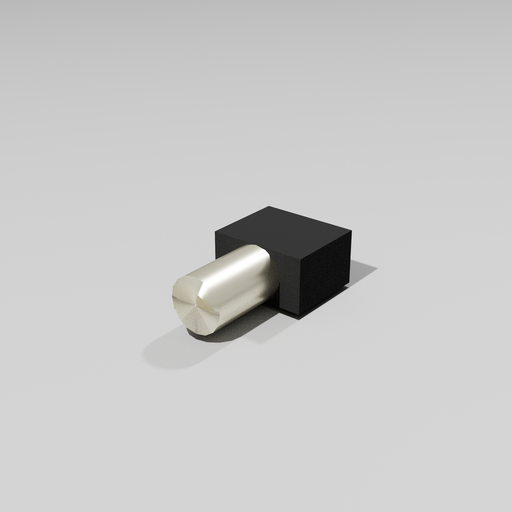}
            \put(4,4){\tiny\textbf{Cylinder}}
        \end{overpic} &
        \begin{overpic}[width=0.139\textwidth, trim=100 100 100 150, clip]{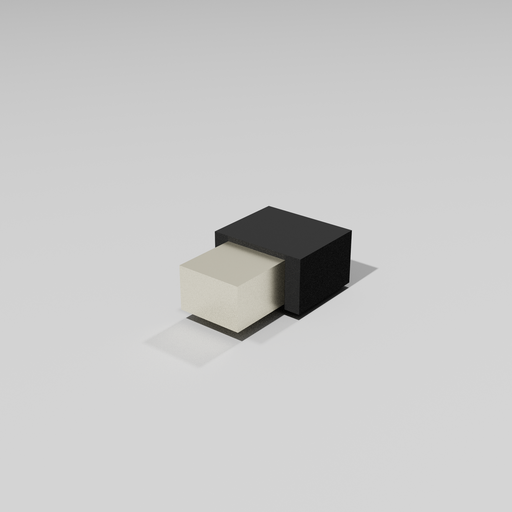}
            \put(4,4){\tiny\textbf{Cuboid}}
        \end{overpic} &
        \begin{overpic}[width=0.139\textwidth, trim=100 100 100 150, clip]{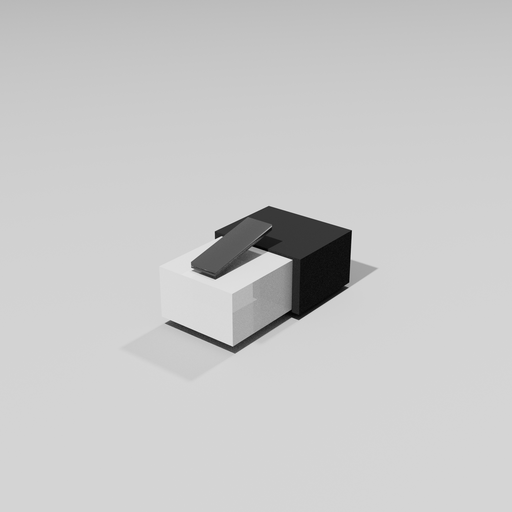}
            \put(4,4){\tiny\textbf{Ethernet-style}}
        \end{overpic} &
        \begin{overpic}[width=0.139\textwidth, trim=100 100 100 150, clip]{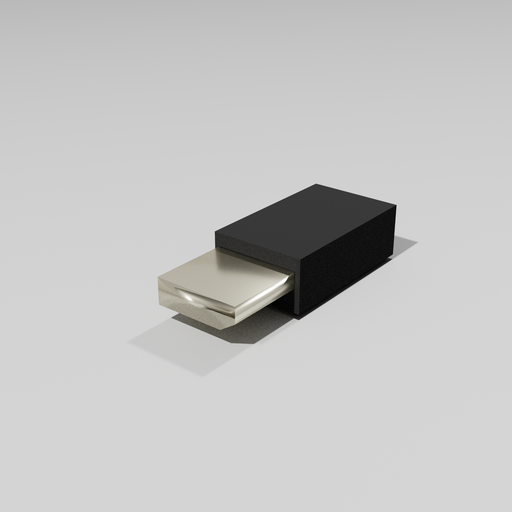}
            \put(4,4){\tiny\textbf{DisplayPort}}
        \end{overpic} &
        \begin{overpic}[width=0.139\textwidth, trim=100 100 100 150, clip]{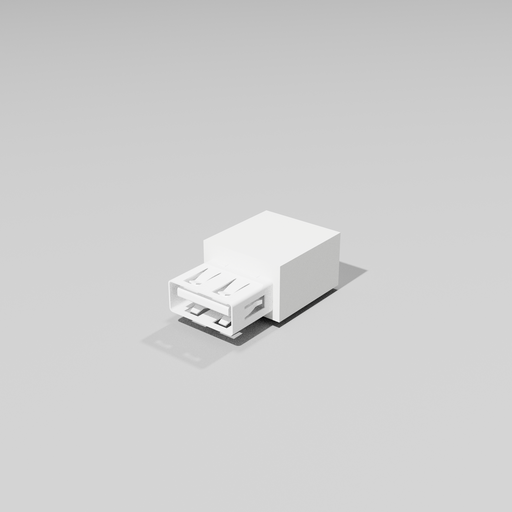}
            \put(4,4){\tiny\textbf{USB-A}}
        \end{overpic} &
        \begin{overpic}[width=0.139\textwidth, trim=150 120 50 130, clip]{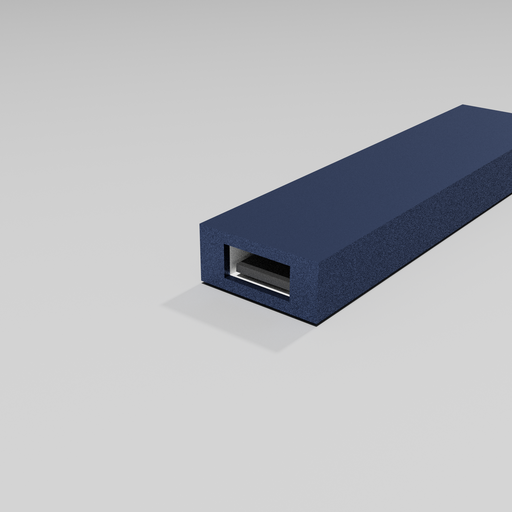}
            \put(4,4){\tiny\textbf{USB-A Socket}}
        \end{overpic} &
        \begin{overpic}[width=0.139\textwidth, trim=100 100 100 150, clip]{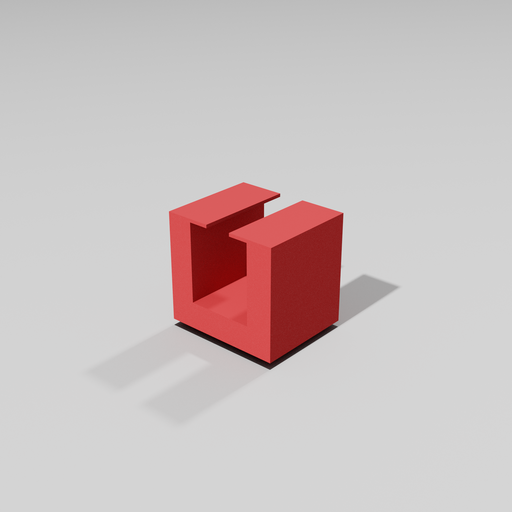}
            \put(4,4){\tiny\textbf{Clip}}
        \end{overpic}
    \end{tabular}%
    }

    \caption{
    \textbf{WireCraft Assets}.
    \textit{Top:} the two configurable wire representations, articulated and FEM-based deformable, shown as rendered DLOs (left) and their construction-level chain/mesh structure (right). \textit{Bottom:} a subset of representative 3D-printable connectors and task fixtures.
    }
    \label{fig:wire_assets}
\end{figure*}

\textbf{DLO Assets.}
The articulated wires are parameterized by length, radius, colors, damping, and stiffness. The deformable wires are simulated via FEM, where Young's modulus, Poisson's ratio, and density govern its elastic response, alongside geometric parameters for length, radius, and mesh resolution. Both share the same connector and task-component interfaces, so all tasks can be instantiated with either wire type.

\textbf{Connector Assets.} 
DLO-mounted connectors are 3D-printable and span two families: reference-based connectors modeled after real standards such as USB-A, and geometry-based connectors built from primitives such as cylinders and cuboids. They differ in rotational symmetry, allowing task difficulty to range from rotation-invariant to orientation-sensitive insertion.

\textbf{Fixture Assets.} 
Task-side fixtures, including clips, channels, and sockets, are likewise 3D-printable. They define the geometric constraints of each task family and remain configurable for different layouts and difficulty levels.

\textbf{Robot Embodiments.}
Three simulated embodiments are supported: the UR5, Franka, and Trossen Stationary AI, each with its own gripper and control interface. Tasks can be instantiated across different arms under a unified observation framework, where robot-specific proprioception scales to each robot's degrees of freedom.

\section{WireCraft Benchmark}
\label{sec:wirecraft_benchmark}

\subsection{WireCraft Tasks}

We benchmark three industrial task families. \textbf{Connector Insertion} performs precision contact-rich insertion with a trailing deformable wire; \textbf{Clip Routing} performs fixture-constrained routing under contact and topological constraints; and \textbf{Channel Seating} performs continuous shape control along extended geometric constraints.

\textbf{Task Curriculum.}
Task difficulty varies through connector geometry, fixture complexity, and path geometry. Connector insertion ranges from a rotation-invariant cylindrical connector to an asymmetric, orientation-sensitive DisplayPort-style connector. Clip routing difficulty scales with the number and placement of clips, while channel seating difficulty is controlled by channel curvature.

\textbf{Randomization.}
All tasks share a randomized free-fall wire initialization, where the wire is spawned with varying position and orientation. Task-specific randomization varies the socket position for insertion, clip position and orientation for routing, and channel geometry for seating; detailed ranges are provided in Appendix~\ref{appendix:task_randomization}.

\subsection{WireCraft Demonstrations}
\label{subsec:wirecraft_dataset}

\textbf{Data Collection.}
WireCraft supports four data collection methods that share a common dataset schema. %
Scripted policies provide scalable, successful demonstrations using privileged simulator state; RL rollouts add closed-loop interaction data and failure-adjacent states; human teleoperation in simulation captures corrective visual strategies for contact-rich manipulation; and real-world UR5 demonstrations provide physical validation. Our data-collection stack builds on an open-source ROS\,2 + LeRobot teleoperation stack for WireCraft, with a shared episode recorder for simulation and real-robot data.

\textbf{Shared Dataset Schema.} All sources are serialized into a LeRobot v3.0 dataset format. For the UR5 robot, they include a 7-D state (six joint positions and gripper), a 7-D action (six joint commands and gripper), synchronized wrist and third-person RGB streams at the task control rate, and the language instruction for each task. This shared schema allows IL and VLA policies to be trained from any mixture of these sources without policy-specific task wrappers.

\section{Experiments}
\label{sec:baseline_experiments}

To establish the benchmark's validity, we evaluate representative RL, IL, and VLA methods on our task suite under common task definitions, randomization, and success criteria. Experiments cover RL, IL, and VLA baselines on deformable-wire connector insertion, and privileged state PPO, ACT, and DiT VLA baselines on clip routing and channel seating.

For connector insertion, \textbf{Reach Success} (\srreach{}) measures whether the gripper-held connector comes within 2~cm of the socket, and \textbf{Insert Success} (\srinsert{}) measures whether the tip seats in the socket. For clip routing, success requires the wire to be seated in the clips. For channel seating, success requires the wire to lie inside the channel over 80\% of its length. Simulation results are reported as mean $\pm$ standard deviation across seeds.

\subsection{Baselines on Connector Insertion}
\label{subsec:il_vla_baselines}

On Ethernet-style deformable-wire insertion with 3 mm clearance, we evaluate PPO~\cite{schulman2017ppo}, SAC from Demonstration (SACfD)~\cite{vecerik2017leveraging}, Vision PPO~\cite{schulman2017ppo, oquab2024dinov2, jaegleperceiver, Pinto2017AsymmetricAC}, ACT \cite{Zhaoetal2023act}, Diffusion Policy \cite{chi2023diffusion}, a Diffusion Transformer VLA policy \cite{cadene2024lerobot}, and \pizero{} \cite{black2025pi05}. IL and VLA policies use the same dataset and evaluation, while state PPO and SACfD use privileged state information. IL/VLA results report the best observed checkpoint among three \srreach{}-selected checkpoints (see Appendix \ref{app:ilvla_details}).

\begin{figure*}[h]
    \centering
    \includegraphics[
        width=0.7\textwidth,
    ]{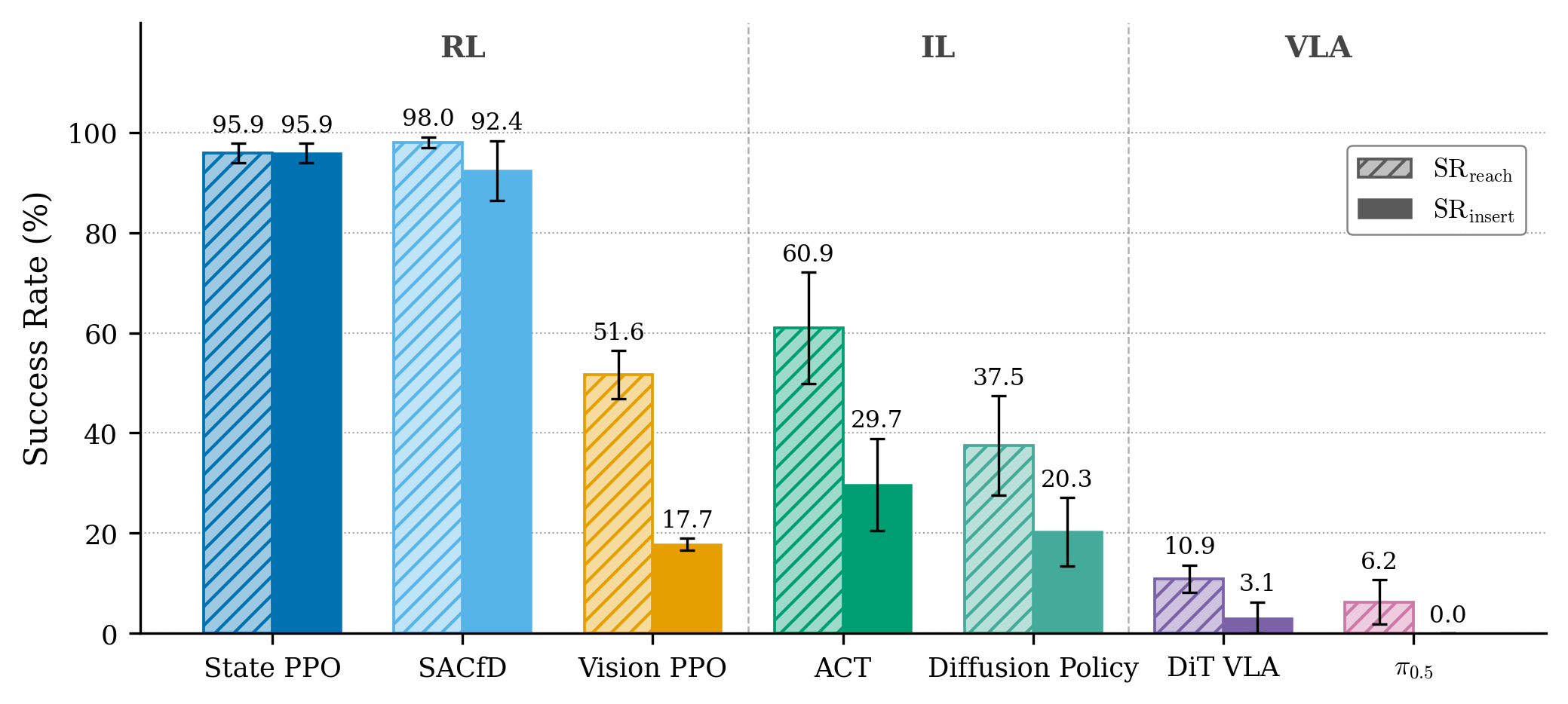}

    \caption{
    \textbf{Ethernet-style Connector Insertion performance.}
    \srreach{} and \srinsert{} on the shared simulated insertion task show privileged RL solves the task reliably, while vision-based baselines exhibit a substantial gap between reaching the socket neighborhood and completing insertion.
    }
    \label{fig:stacked_cuboid_results}
\end{figure*}

Figure~\ref{fig:stacked_cuboid_results} shows three trends. First, privileged state-based RL solves the task reliably ($\srinsert{}\!\geq\!92\%$ for both PPO and SACfD), establishing that the task is well-posed and that contact-rich alignment is tractable when socket and tail pose are directly observable. Second, every non-privileged baseline exhibits a sizeable reach--insert gap, showing that reaching the socket does not ensure successful insertion. Vision PPO, however, collapses at the reach stage ($51.6\%$), indicating the bottleneck for vision RL is both contact dynamics and RGB-only state estimation under occlusion and tight clearance. ACT and Diffusion Policy achieve higher contact-rich insertion phase success rates than the other image-input baselines, but both retain a clear gap between reaching the socket and completing insertion. Third, both VLA baselines underperform IL, with \pizero{} reaching only $6.2\%/0.0\%$; we attribute this primarily to a much shorter fine-tuning budget and a frozen vision tower (Appendix~\ref{app:ilvla_details}), together with a pre-training distribution containing limited fine-grained contact-rich DLO interaction, rather than to a fundamental limitation of the architecture.

\subsection{RL Baselines on Diverse Connector Insertion}
\label{subsec:rl_baselines}

We then extend the RL baselines to the full connector suite under the same 3 mm fixture clearance and task interfaces, evaluating privileged-state PPO, SACfD, and Vision PPO in Table~\ref{tab:rl_port_baselines}.

\begin{table*}[h]
\centering
\small
\caption{
RL baseline success rates (\%) on the diverse connector insertion tasks.
}
\label{tab:rl_port_baselines}
\setlength{\tabcolsep}{6pt}
\begin{tabular}{llccc}
\toprule
Connector & Metric & State PPO & SACfD & Vision PPO \\
\midrule
Cylinder
& \srreach{}  & $95.00 \pm 1.25$ & $97.27 \pm 1.06$ & \bm{$68.07 \pm 6.02$} \\
& \srinsert{} & $93.75 \pm 1.25$ & $85.55 \pm 8.68$ & \bm{$41.49 \pm 1.32$} \\
\midrule
Cuboid
& \srreach{}  & \bm{$97.08 \pm 1.44$} & $96.70 \pm 1.06$ & $41.67 \pm 4.73$ \\
& \srinsert{} & $95.00 \pm 3.75$ & \bm{$93.06 \pm 6.46$} & $ 6.25\pm 3.31$ \\
\midrule
Ethernet-style
& \srreach{}  & $95.86 \pm 1.93$ & \bm{$98.05 \pm 1.05$} & $51.63 \pm 4.82$ \\
& \srinsert{} & \bm{$95.86 \pm 1.93$} & $92.40 \pm 5.91$ & $17.74 \pm 1.21$ \\
\midrule
DisplayPort
& \srreach{}  & $95.00 \pm 1.25$ & $95.62 \pm 10.58$ & $34.01 \pm 5.73$ \\
& \srinsert{} & $92.92 \pm 3.15$ & $77.43 \pm 6.17$ & $3.32 \pm 1.45$ \\
\bottomrule
\end{tabular}
\end{table*}

Three observations follow: First, both privileged baselines, PPO and SACfD, are largely insensitive to connector geometry, solving the alignment-sensitive DisplayPort about as well as the simpler connectors ($\srinsert{}\!\geq\!92\%$ for PPO; $\geq\!77\%$ for SACfD). Second, the ordering is not monotonic in geometric difficulty: the rotation-invariant cylinder, nominally the easiest connector, sits at or below the cuboid and Ethernet-style for both methods. We attribute this tentatively to the shared reward fixing a target socket orientation, which penalizes the cylinder's many equally valid yaw configurations and broadens the set of rewarded behaviors. Third, Vision PPO degrades sharply on every connector, and not monotonically; Ethernet-style is solved more often than the simpler cuboid. One likely explanation is the visual cues each shape provides for yaw estimation: the latch-like feature may give an orientation cue that the smooth cuboid lacks in RGB without depth.

Across connectors, the curriculum degrades vision-based performance while leaving privileged-state performance essentially unaffected, indicating that it stresses the visual observability of orientation rather than control. This supports its use as a difficulty axis (see Appendix~\ref{appendix:parameters}).

\subsection{Clip Routing and Channel Seating}
\label{subsec:ppo_routing_seating}

We further evaluate State PPO, ACT, and DiT VLA on 1-clip routing and straight-channel seating in Table~\ref{tab:routing_seating_stateppo}. ACT and DiT VLA results are averaged over three evaluation seeds. With privileged state observations, PPO achieves 95.32\% and 82.33\% success, respectively. Despite plateauing training losses, image-based ACT and DiT VLA achieve lower success rates, particularly in channel seating, with 18.80\% and 15.62\% success, respectively. Dataset sizes, training configurations, and checkpoint selection are detailed in Appendix~\ref{app:routing_seating_details}.

\begin{table}[h]
\centering
\small
\caption{Evaluation results on 1-clip routing and straight channel seating tasks in simulation.}
\label{tab:routing_seating_stateppo}
\begin{tabular}{llccc}
\toprule
Task & Setting & State PPO & ACT & DiT VLA \\
\midrule
Clip routing & 1-clip & $\bm{95.32 \pm 0.93}$ & $64.06 \pm 11.16$ & $39.06 \pm 16.16$ \\
Channel seating & Straight channel & $\bm{82.33 \pm 0.31}$ & $18.80 \pm 7.65$ & $15.62 \pm 12.88$ \\
\bottomrule
\end{tabular}
\end{table}

\section{Real-World Validation}
\label{sec:real}

For real-world validation, we evaluate all three WireCraft tasks on a physical UR5 using ACT, which achieved the highest insertion success rate among image-input baselines in Section~\ref{sec:baseline_experiments}. We first reproduce the physical setup in simulation, then train ACT on four combinations of real demonstrations (R), scripted simulation demonstrations (S), and simulation RL rollouts (P): R, R+S, R+P, and R+S+P. All configurations train for 80k steps, with checkpoints selected by the lowest L1 loss on held-out real demonstrations. Preliminary simulation-only policies showed poor zero-shot transfer, so we focus the quantitative evaluation on mixtures containing real demonstrations.

\begin{figure}[h]
\centering
\includegraphics[width=\textwidth]{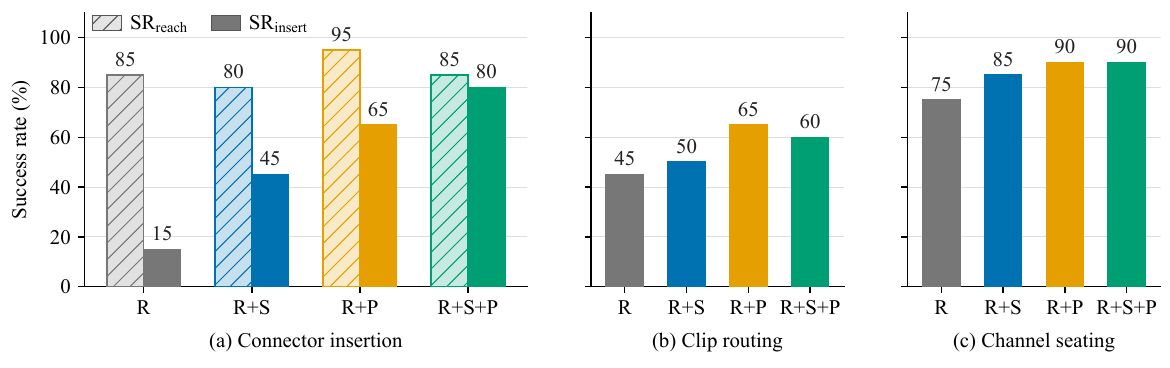}
\caption{
ACT success rates on the physical UR5 for (a) Ethernet-style connector insertion, (b) 1-clip routing, and (c) straight-channel seating. Insertion reports both \srreach{} and \srinsert{}; routing and seating each report task success.
}
\label{fig:real_world_results}
\end{figure}

The hardware evaluation uses a narrower randomization range than simulation. For insertion, the connector is placed within a $3\times3$~cm region with varying tilt, while the socket remains fixed. For 1-clip routing, a single clip is randomly positioned near the wire on both sides. For straight-channel seating, one wire endpoint remains fixed while the other is placed at varying positions outside the channel. We evaluate each task and data-source combination over 20 real-robot trials, and report different effects of simulation data across tasks in Figure~\ref{fig:real_world_results}.

For insertion, reach success remains between 80\% and 95\% across data combinations, while insertion success increases from 15\% with R to 80\% with R+S+P. Higher reach success does not necessarily translate into higher insertion success: R+P reaches more often than R+S+P (95\% vs.\ 85\%), but inserts less often (65\% vs.\ 80\%). Qualitatively, R+S+P more often produces insertion-compatible grasp poses, whereas R+P sometimes reaches the socket with a grasp that leaves the connector poorly aligned for insertion. The benefit of broader variability of data coverage is therefore associated with grasping the connector in a pose that supports subsequent socket alignment. Successful pickup and reach alone are insufficient.

Clip routing also benefits from simulated data, but the improvement is non-monotonic: success increases from 45\% with R to 65\% with R+P, while R+S+P achieves 60\%. Qualitatively, failures with simulated training data more often place the wire on top of the clip opening instead of beside the clip. These near-misses remain unsuccessful under the binary criterion but indicate closer placement alignment. Clip localization during wire grasping relies on the side-camera view alone, leaving limited visual information for preparing the subsequent placement.

For channel seating, simulated data increase success from 75\% with R to 90\% with R+P and R+S+P. Most failures occur during the initial grasp; once the wire is successfully grasped, seating is usually completed. Its remaining bottleneck is therefore grasp acquisition, unlike the final alignment and placement failures observed in insertion and clip routing.

Overall, simulated trajectories improve the observed success rates across all three tasks, with task-dependent effects on grasp selection and fixture engagement. The hardware results support the portability of the three task definitions under restricted initialization ranges. Since source diversity and dataset size increase together, these comparisons show the benefit of supplementing real demonstrations with simulated trajectories, without isolating the effect of diversity alone. Further setup and failure details are provided in Appendix~\ref{appendix:real_world}.

\section{Limitations}
\label{sec:limitations}

WireCraft has several limitations. First, policy transfer between the articulated and FEM-based wire models has not been evaluated, and their dynamic deformation and contact response remain quantitatively unmatched. Neither model has been calibrated against the physical wire, so hardware task validation does not establish wire-model accuracy or reliable zero-shot transfer. Second, full force/torque feedback remains unused: privileged insertion uses only saturated fingertip contact magnitudes, while the UR5's mounted force/torque sensor is unread. Simulated contact forces remain physically unvalidated, and the benefits of force/torque feedback and optional depth observations for task completion are untested. Third, hardware evaluation covers all three task families but only selected settings on one UR5, leaving generalization across robots, fixtures, and wire materials untested. Success metrics also do not capture contact loads or wire damage. Finally, the VLA results reflect constrained fine-tuning and do not establish an upper bound on VLA capability; longer training, improved checkpoint selection, and stronger visual adaptation remain future work.

\section{Conclusion}
\label{sec:conclusion}

We presented WireCraft, a simulation benchmark for industrial DLO manipulation across connector insertion, clip routing, and channel seating. WireCraft provides configurable 3D-printable assets, articulated and FEM-based wire representations, shared policy interfaces, and trajectory data from scripted, RL, and real-world sources, together with teleoperation tools. Our experiments show a clear gap between task feasibility and current vision-based policy performance. Privileged state-based RL solves the evaluated insertion, routing, and seating settings, indicating that the tasks are well posed in simulation. In contrast, vision-based RL, imitation learning, and VLA policies degrade sharply on connector insertion, while ACT and DiT VLA also fall below State PPO on routing and seating. The reach--insert gap shows that reaching the socket neighborhood is not sufficient: fine alignment, occlusion, deformation, and contact-rich seating remain the central bottlenecks. Real-world UR5 validation covers all three task families and shows improved ACT success rates when real demonstrations are supplemented with simulated trajectories.

By releasing WireCraft, we aim to provide a common substrate for studying where industrial DLO policies fail and how they improve, including visual localization, deformation-aware control, contact-rich insertion, fixture-constrained routing, and sim-to-real transfer.

\acknowledgments{This research has been partially funded by the Industrial Technology Innovation Program [P0030285, Autonomous Assembly through AI Agent-Driven Dexterous Manipulation of Flexible and Complex Industrial Components] of the Ministry of Trade, Industry and Energy of the Republic of Korea.}

\bibliography{example}  %

@article{zhu2022challenges, 
  title={Challenges and outlook in robotic manipulation of deformable objects},
  author={Zhu, Jihong and Cherubini, Andrea and Dune, Claire and Navarro-Alarcon, David and Alambeigi, Farshid and Berenson, Dmitry and Ficuciello, Fanny and Harada, Kensuke and Kober, Jens and Li, Xiang and others},
  journal={IEEE Robotics \& Automation Magazine},
  volume={29},
  number={3},
  pages={67--77},
  year={2022},
  publisher={IEEE}
}

@article{gu2026survey, 
  title={A survey on robotic manipulation of deformable objects: Recent advances, open challenges and new frontiers},
  author={Gu, Feida and Wang, Zhipeng and Zhu, Zhongpan and Ma, Jiajun and Zhou, Yanmin and Jiang, Shuo and He, Bin},
  journal={Neurocomputing},
  pages={134058},
  year={2026},
  publisher={Elsevier}
}

@article{luo2024multistage, 
  title={Multistage cable routing through hierarchical imitation learning},
  author={Luo, Jianlan and Xu, Charles and Geng, Xinyang and Feng, Gilbert and Fang, Kuan and Tan, Liam and Schaal, Stefan and Levine, Sergey},
  journal={IEEE Transactions on Robotics},
  volume={40},
  pages={1476--1491},
  year={2024},
  publisher={IEEE}
}

@INPROCEEDINGS{11127451,
  author={Li, Mingen and Yu, Houjian and Choi, Changhyun},
  booktitle={2025 IEEE International Conference on Robotics and Automation (ICRA)}, 
  title={Routing Manipulation of Deformable Linear Object Using Reinforcement Learning and Diffusion Policy}, 
  year={2025},
  volume={},
  number={},
  pages={01-07},
  doi={10.1109/ICRA55743.2025.11127451}}

@misc{li2025hierarchicaldloroutingreinforcement,
      title={Hierarchical DLO Routing with Reinforcement Learning and In-Context Vision-language Models}, 
      author={Mingen Li and Houjian Yu and Yixuan Huang and Youngjin Hong and Changhyun Choi and Hantao Ye},
      year={2025},
      eprint={2510.19268},
      archivePrefix={arXiv},
      primaryClass={cs.RO},
      url={https://arxiv.org/abs/2510.19268}, 
}

@inproceedings{wilson2023cable,
  title={Cable Routing and Assembly using Tactile-driven Motion Primitives},
  author={Wilson, Achu and Jiang, Helen and Lian, Wenzhao and Yuan, Wenzhen},
  booktitle={2023 IEEE International Conference on Robotics and Automation (ICRA)},
  pages={10408--10414},
  year={2023},
  organization={IEEE}
}

@inproceedings{Li_2024,
   title={Learning for Deformable Linear Object Insertion Leveraging Flexibility Estimation from Visual Cues},
   url={http://dx.doi.org/10.1109/ICRA57147.2024.10610419},
   DOI={10.1109/icra57147.2024.10610419},
   booktitle={2024 IEEE International Conference on Robotics and Automation (ICRA)},
   publisher={IEEE},
   author={Li, Mingen and Choi, Changhyun},
   year={2024},
   month=may, pages={5183–5189} }

@article{Cao2026RoboticCF,
  title={Robotic Cable Following and Insertion With Tactile Sensing},
  author={Bin Cao and Xizhe Zang and Shouqiang Li and Xuehe Zhang and Chang-le Li and Jie Zhao},
  journal={IEEE Sensors Journal},
  year={2026},
  volume={26},
  pages={6216-6224},
  url={https://api.semanticscholar.org/CorpusID:284252279}
}

@inproceedings{zhao2022offline,
  title={Offline Meta-Reinforcement Learning for Industrial Insertion},
  author={Zhao, Tony Z and Luo, Jianlan and Sushkov, Oleg and Pevceviciute, Rugile and Heess, Nicolas and Scholz, Jon and Schaal, Stefan and Levine, Sergey},
  booktitle={2022 IEEE International Conference on Robotics and Automation (ICRA)},
  pages={6386--6393},
  year={2022},
  organization={IEEE}
}

@article{yu2025generalizable,
  title={Generalizable whole-body global manipulation of deformable linear objects by dual-arm robot in 3-d constrained environments},
  author={Yu, Mingrui and Lv, Kangchen and Wang, Changhao and Jiang, Yongpeng and Tomizuka, Masayoshi and Li, Xiang},
  journal={The International Journal of Robotics Research},
  volume={44},
  number={4},
  pages={607--639},
  year={2025},
  publisher={Sage Publications Sage UK: London, England}
}

@ARTICLE{9732654,
  author={Jin, Shiyu and Lian, Wenzhao and Wang, Changhao and Tomizuka, Masayoshi and Schaal, Stefan},
  journal={IEEE Robotics and Automation Letters}, 
  title={Robotic Cable Routing with Spatial Representation}, 
  year={2022},
  volume={7},
  number={2},
  pages={5687-5694},
  doi={10.1109/LRA.2022.3158377}}

@inproceedings{10.1145/3776942.3777006,
author = {Li, Wen and Zhu, Zhinan and Li, Dandan and Gong, Zheng and Lv, Naijing and Chen, Hao and Li, Jianguo},
title = {An Automated Cable Laying Simulation Method by Robot for Planar Routing},
year = {2025},
isbn = {9798400715839},
publisher = {Association for Computing Machinery},
address = {New York, NY, USA},
url = {https://doi.org/10.1145/3776942.3777006},
doi = {10.1145/3776942.3777006},
booktitle = {Proceedings of the 2025 11th Annual International Conference on Network and Information Systems for Computers},
pages = {209–214},
numpages = {6},
location = {
},
series = {ICNISC '25}
}

@article{james2020rlbench,
  title={Rlbench: The robot learning benchmark \& learning environment},
  author={James, Stephen and Ma, Zicong and Arrojo, David Rovick and Davison, Andrew J},
  journal={IEEE Robotics and Automation Letters},
  volume={5},
  number={2},
  pages={3019--3026},
  year={2020},
  publisher={IEEE}
}

@INPROCEEDINGS{9561766,
  author={Laezza, Rita and Gieselmann, Robert and Pokorny, Florian T. and Karayiannidis, Yiannis},
  booktitle={2021 IEEE International Conference on Robotics and Automation (ICRA)}, 
  title={ReForm: A Robot Learning Sandbox for Deformable Linear Object Manipulation}, 
  year={2021},
  volume={},
  number={},
  pages={4717-4723},
  doi={10.1109/ICRA48506.2021.9561766}}

@INPROCEEDINGS{9926677,
  author={Chang, Peng and Luo, Rui and Zolotas, Mark and Padır, Taşkın},
  booktitle={2022 IEEE 18th International Conference on Automation Science and Engineering (CASE)}, 
  title={Manipulation of Deformable Linear Objects in Benchmark Task Spaces}, 
  year={2022},
  volume={},
  number={},
  pages={1910-1916},
  doi={10.1109/CASE49997.2022.9926677}}

@article{taomaniskill3,
  title={ManiSkill3: GPU Parallelized Robotics Simulation and Rendering for Generalizable Embodied AI},
  author={Stone Tao and Fanbo Xiang and Arth Shukla and Yuzhe Qin and Xander Hinrichsen and Xiaodi Yuan and Chen Bao and Xinsong Lin and Yulin Liu and Tse-kai Chan and Yuan Gao and Xuanlin Li and Tongzhou Mu and Nan Xiao and Arnav Gurha and Viswesh Nagaswamy Rajesh and Yong Woo Choi and Yen-Ru Chen and Zhiao Huang and Roberto Calandra and Rui Chen and Shan Luo and Hao Su},
  journal = {Robotics: Science and Systems},
  year={2025},
}

@inproceedings{black2025pi05,
  title     = {$\pi_{0.5}$: a Vision-Language-Action Model with Open-World Generalization},
  author    = {Black, Kevin and Brown, Noah and others},
  booktitle = {Proceedings of the 9th Conference on Robot Learning (CoRL)},
  year      = {2025},
  note      = {Oral Presentation}
}

@article{mittal2025isaaclab,
  title   = {Isaac Lab: A GPU-Accelerated Simulation Framework for Multi-Modal Robot Learning},
  author  = {Mittal, Mayank and others},
  journal = {arXiv preprint arXiv:2511.04831},
  year    = {2025}
}

@inproceedings{lin2021softgym,
  title={Softgym: Benchmarking deep reinforcement learning for deformable object manipulation},
  author={Lin, Xingyu and Wang, Yufei and Olkin, Jake and Held, David},
  booktitle={Conference on Robot Learning},
  pages={432--448},
  year={2021},
  organization={PMLR}
}

@article{schulman2017ppo,
  title={Proximal policy optimization algorithms},
  author={Schulman, John and Wolski, Filip and Dhariwal, Prafulla and Radford, Alec and Klimov, Oleg},
  journal={arXiv preprint arXiv:1707.06347},
  year={2017}
}

@article{Zhaoetal2023act,
  author  = {Zhao, Tony Z. and Kumar, Vikash and Levine, Sergey and Finn, Chelsea},
  title   = {Learning Fine-Grained Bimanual Manipulation with Low-Cost Hardware},
  journal = {Robotics: Science and Systems (RSS)},
  year    = {2023},
  doi     = {10.15607/RSS.2023.XIX.016}
}

@article{Peeblesetal2023dit,
  author  = {Peebles, William and Xie, Saining},
  title   = {Scalable Diffusion Models with Transformers},
  journal = {2023 IEEE/CVF International Conference on Computer Vision (ICCV)},
  year    = {2023},
  pages   = {4172--4182},
  doi     = {10.1109/iccv51070.2023.00387}
}

@misc{isaacsim,
  author = {NVIDIA},
  title = {NVIDIA Isaac Sim},
  year = {2026},
  publisher = {GitHub},
  journal = {GitHub repository},
  howpublished = {\url{https://github.com/isaac-sim/IsaacSim}},
  url = {https://docs.isaacsim.omniverse.nvidia.com/latest/index.html}
}

@misc{isaaclab_deformable_object,
  author       = {{Isaac Lab Project Developers}},
  title        = {{Interacting with a Deformable Object}},
  howpublished = {Isaac Lab Documentation},
  year         = {2026},
  url          = {https://isaac-sim.github.io/IsaacLab/main/source/tutorials/01_assets/run_deformable_object.html},
  note         = {Accessed: 2026-05-16}
}

@misc{physx_soft_bodies,
  author       = {{NVIDIA}},
  title        = {{Soft Bodies}},
  howpublished = {PhysX Documentation},
  year         = {2024},
  url          = {https://nvidia-omniverse.github.io/PhysX/physx/5.4.0/docs/SoftBodies.html},
  note         = {Accessed: 2026-05-16}
}

@article{oquab2024dinov2,
  title={DINOv2: Learning Robust Visual Features without Supervision},
  author={Oquab, Maxime and Darcet, Timoth{\'e}e and Moutakanni, Th{\'e}o and Vo, Huy and Szafraniec, Marc and Khalidov, Vasil and Fernandez, Pierre and Haziza, Daniel and Massa, Francisco and El-Nouby, Alaaeldin and others},
  journal={Transactions on Machine Learning Research Journal},
  year={2024}
}

@inproceedings{Pinto2017AsymmetricAC,
  title={Asymmetric actor critic for image-based robot learning},
  author={Lerrel Pinto and Marcin Andrychowicz and Peter Welinder and Wojciech Zaremba and Pieter Abbeel},
  booktitle={Robotics: Science and Systems},
  year={2018}
}

@inproceedings{jaegleperceiver,
  title={Perceiver IO: A General Architecture for Structured Inputs \& Outputs},
  author={Jaegle, Andrew and Borgeaud, Sebastian and Alayrac, Jean-Baptiste and Doersch, Carl and Ionescu, Catalin and Ding, David and Koppula, Skanda and Zoran, Daniel and Brock, Andrew and Shelhamer, Evan and others},
  year={2022},
  booktitle={International Conference on Learning Representations}
}

@article{kimble2022performance,
  title={Performance measures to benchmark the grasping, manipulation, and assembly of deformable objects typical to manufacturing applications},
  author={Kimble, Kenneth and Albrecht, Justin and Zimmerman, Megan and Falco, Joe},
  journal={Frontiers in Robotics and AI},
  volume={9},
  pages={999348},
  year={2022},
  publisher={Frontiers Media SA}
}

@inproceedings{huang2021plasticinelab,
  title={PlasticineLab: A Soft-Body Manipulation Benchmark with Differentiable Physics},
  author={Huang, Zhiao and Hu, Yuanming and Du, Tao and Zhou, Siyuan and Su, Hao and Tenenbaum, Joshua B and Gan, Chuang},
  year={2021},
  booktitle={International Conference on Learning Representations}
}

@inproceedings{chendaxbench,
  title={DaxBench: Benchmarking Deformable Object Manipulation with Differentiable Physics},
  author={Chen, Siwei and Xu, Yiqing and Yu, Cunjun and Li, Linfeng and Ma, Xiao and Xu, Zhongwen and Hsu, David},
  booktitle={The Eleventh International Conference on Learning Representations},
  year={2023}
}

@inproceedings{seita2021learning,
  title={Learning to rearrange deformable cables, fabrics, and bags with goal-conditioned transporter networks},
  author={Seita, Daniel and Florence, Pete and Tompson, Jonathan and Coumans, Erwin and Sindhwani, Vikas and Goldberg, Ken and Zeng, Andy},
  booktitle={2021 IEEE International Conference on Robotics and Automation (ICRA)},
  pages={4568--4575},
  year={2021},
  organization={IEEE}
}

@inproceedings{zhou2020dlocableharness,
  title={A Practical Solution to Deformable Linear Object Manipulation: A Case Study on Cable Harness Connection},
  author={Zhou, Hang and Li, Shunchong and Lu, Qi and Qian, Jinwu},
  booktitle={2020 5th International Conference on Advanced Robotics and Mechatronics (ICARM)},
  pages={329--333},
  year={2020},
  organization={IEEE}
}

@inproceedings{zhang2024harnessing,
  title={Harnessing with twisting: Single-arm deformable linear object manipulation for industrial harnessing task},
  author={Zhang, Xiang and Lin, Hsien-Chung and Zhao, Yu and Tomizuka, Masayoshi},
  booktitle={2024 IEEE/RSJ International Conference on Intelligent Robots and Systems (IROS)},
  pages={4069--4075},
  year={2024},
  organization={IEEE}
}

@inproceedings{yu2022shape,
  title={Shape control of deformable linear objects with offline and online learning of local linear deformation models},
  author={Yu, Mingrui and Zhong, Hanzhong and Li, Xiang},
  booktitle={2022 International Conference on Robotics and Automation (ICRA)},
  pages={1337--1343},
  year={2022},
  organization={IEEE}
}

@inproceedings{zhang2021deformable,
  title={Deformable linear object prediction using locally linear latent dynamics},
  author={Zhang, Wenbo and Schmeckpeper, Karl and Chaudhari, Pratik and Daniilidis, Kostas},
  booktitle={2021 IEEE International Conference on Robotics and Automation (ICRA)},
  pages={13503--13509},
  year={2021},
  organization={IEEE}
}

@article{yan2020self,
  title={Self-supervised learning of state estimation for manipulating deformable linear objects},
  author={Yan, Mengyuan and Zhu, Yilin and Jin, Ning and Bohg, Jeannette},
  journal={IEEE robotics and automation letters},
  volume={5},
  number={2},
  pages={2372--2379},
  year={2020},
  publisher={IEEE}
}

@article{cao2024shape,
  title={Shape control of elastic deformable linear objects for robotic cable assembly},
  author={Cao, Bin and Zang, Xizhe and Zhang, Xuehe and Chen, Zhuo and Li, Shouqiang and Zhao, Jie},
  journal={Advanced Intelligent Systems},
  volume={6},
  number={7},
  pages={2300835},
  year={2024},
  publisher={Wiley Online Library}
}

@inproceedings{tanureza2025industrial,
  title={Industrial Cabling in Constrained Environments: a Practical Approach and Current Challenges},
  author={Tanureza, Jaya and Michalak, Benjamin and Radke, Marcel and Haninger, Kevin},
  booktitle={2025 IEEE/SICE International Symposium on System Integration (SII)},
  pages={1428--1433},
  year={2025},
  organization={IEEE}
}

@inproceedings{kienle2025ai,
  title={AI-based Framework for Robust Model-Based Connector Mating in Robotic Wire Harness Installation},
  author={Kienle, Claudius and Alt, Benjamin and Schneider, Finn and Pertlwieser, Tobias and J{\"a}kel, Rainer and Rayyes, Rania},
  booktitle={2025 IEEE 21st International Conference on Automation Science and Engineering (CASE)},
  pages={2444--2449},
  year={2025},
  organization={IEEE}
}

@inproceedings{azulay2025motorcycle,
  title={MOTORCYCLE 1.0: Automating Bimanual Cable Routing Around Fixtures on the NIST Task Board},
  author={Azulay, Osher and Kondap, Kavish and Drake, Jaimyn and Xie, Shuangyu and Li, Hui and Chitta, Sachin and Goldberg, Ken},
  booktitle={2025 IEEE 21st International Conference on Automation Science and Engineering (CASE)},
  pages={2636--2641},
  year={2025},
  organization={IEEE}
}

@article{zhang2026modesuite,
  title={MoDeSuite: Robot Learning Task Suite for Benchmarking Mobile Manipulation with Deformable Objects},
  author={Zhang, Yuying and Luck, Kevin Sebastian and Verdoja, Francesco and Kyrki, Ville and Pajarinen, Joni},
  journal={IEEE Robotics and Automation Letters},
  year={2026},
  publisher={IEEE}
}

@article{chi2023diffusion,
  title={Diffusion policy: Visuomotor policy learning via action diffusion},
  author={Chi, Cheng and Xu, Zhenjia and Feng, Siyuan and Cousineau, Eric and Du, Yilun and Burchfiel, Benjamin and Tedrake, Russ and Song, Shuran},
  journal={The International Journal of Robotics Research},
  volume={44},
  number={10-11},
  pages={1684--1704},
  year={2025},
  publisher={Sage Publications Sage UK: London, England}
}

@inproceedings{hou2025dita,
  title={Dita: Scaling diffusion transformer for generalist vision-language-action policy},
  author={Hou, Zhi and Zhang, Tianyi and Xiong, Yuwen and Duan, Haonan and Pu, Hengjun and Tong, Ronglei and Zhao, Chengyang and Zhu, Xizhou and Qiao, Yu and Dai, Jifeng and others},
  booktitle={Proceedings of the IEEE/CVF International Conference on Computer Vision},
  pages={7686--7697},
  year={2025}
}

@misc{cadene2024lerobot,
  title={LeRobot: State-of-the-Art Machine Learning for Real-World Robotics in PyTorch},
  author={Cadene, Remi and Alibert, Simon and Soare, Alexander and Gallouedec, Quentin and Zouitine, Adil and Palma, Steven and Kooijmans, Pepijn and Aractingi, Michel and Shukor, Mustafa and Aubakirova, Dana and others},
  year={2024},
  howpublished={\url{https://github.com/huggingface/lerobot}}
}

@article{vecerik2017leveraging,
  title={Leveraging demonstrations for deep reinforcement learning on robotics problems with sparse rewards},
  author={Vecerik, Mel and Hester, Todd and Scholz, Jonathan and Wang, Fumin and Pietquin, Olivier and Piot, Bilal and Heess, Nicolas and Roth{\"o}rl, Thomas and Lampe, Thomas and Riedmiller, Martin},
  journal={arXiv preprint arXiv:1707.08817},
  year={2017}
}

@INPROCEEDINGS{chen2023dermujoco,
  author={Chen, Qi Jing and Bretl, Timothy and Pham, Quang-Cuong},
  booktitle={2025 IEEE/RSJ International Conference on Intelligent Robots and Systems (IROS)}, 
  title={Accurate Simulation and Parameter Identification of Deformable Linear Objects using Discrete Elastic Rods in Generalized Coordinates}, 
  year={2025},
  volume={},
  number={},
  pages={20454-20460},
  doi={10.1109/IROS60139.2025.11247160}}

@inproceedings{sun2024dexdlo,
  title={{DexDLO}: Learning Goal-Conditioned Dexterous Policy for Dynamic Manipulation of Deformable Linear Objects},
  author={Sun, Zhaole and Zhu, Jihong and Fisher, Robert B.},
  booktitle={2024 IEEE International Conference on Robotics and Automation (ICRA)},
  pages={16009--16015},
  year={2024},
  organization={IEEE},
  doi={10.1109/ICRA57147.2024.10610754}
}

@inproceedings{chen2024deform,
  title={Differentiable Discrete Elastic Rods for Real-Time Modeling of Deformable Linear Objects},
  author={Chen, Yizhou and Zhang, Yiting and Brei, Zachary and Zhang, Tiancheng and Chen, Yuzhen and Wu, Julie and Vasudevan, Ram},
  booktitle={Proceedings of The 8th Conference on Robot Learning},
  pages={2996--3014},
  year={2025},
  editor={Agrawal, Pulkit and Kroemer, Oliver and Burgard, Wolfram},
  volume={270},
  series={Proceedings of Machine Learning Research},
  publisher={PMLR},
  url={https://proceedings.mlr.press/v270/chen25d.html}
}

@inproceedings{cao2026dlolab,
  title={{DLO-Lab}: Benchmarking Deformable Linear Object Manipulations with Differentiable Physics},
  author={Cao, Junyi and Wang, Yian and Xiong, Ziyan and Lin, Chunru and Chen, Zhehuan and Gan, Chuang},
  booktitle={Proceedings of the 43rd International Conference on Machine Learning (ICML)},
  year={2026},
  note={Accepted},
  url={https://icml.cc/virtual/2026/poster/63391}
}

@inproceedings{govoni2025dlo,
  title={Performance Analysis of a Mass-Spring-Damper Deformable Linear Object Model in Robotic Simulation Frameworks},
  author={Govoni, Andrea and Zubair, Nadia and Soprani, Simone and Palli, Gianluca},
  booktitle={European Robotics Forum 2025},
  pages={187--192},
  year={2025},
  series={Springer Proceedings in Advanced Robotics},
  volume={36},
  publisher={Springer},
  doi={10.1007/978-3-031-89471-8_29}
}

@article{chen2025deft,
  title={{DEFT}: Differentiable Branched Discrete Elastic Rods for Modeling Furcated {DLOs} in Real-Time},
  author={Chen, Yizhou and Wu, Xiaoyue and Zong, Yeheng and Chen, Yuzhen and Li, Anran and Zhang, Bohao and Vasudevan, Ram},
  journal={arXiv preprint arXiv:2502.15037},
  year={2025},
  doi={10.48550/arXiv.2502.15037},
  url={https://arxiv.org/abs/2502.15037}
}

@incollection{bergou2008discrete,
  title={Discrete elastic rods},
  author={Bergou, Mikl{\'o}s and Wardetzky, Max and Robinson, Stephen and Audoly, Basile and Grinspun, Eitan},
  booktitle={ACM Siggraph 2008 Papers},
  pages={1--12},
  year={2008}
}

@misc{nist_assembly_task_boards,
  author       = {{National Institute of Standards and Technology}},
  title        = {{Assembly Performance Metrics and Test Methods}},
  howpublished = {\url{https://www.nist.gov/el/intelligent-systems-division-73500/robotic-grasping-and-manipulation-assembly/assembly}},
  year         = {2018},
  note         = {Accessed: 2026-05-22}
}

@article{chen2026manipulationnet,
  title={Manipulationnet: An infrastructure for benchmarking real-world robot manipulation with physical skill challenges and embodied multimodal reasoning},
  author={Chen, Yiting and Kimble, Kenneth and Adelson, Edward H and Asfour, Tamim and Chanrungmaneekul, Podshara and Chitta, Sachin and Chitambar, Yash and Chen, Ziyang and Goldberg, Ken and Kragic, Danica and others},
  journal={arXiv preprint arXiv:2603.04363},
  year={2026}
}

@inproceedings{hoang2025geometry,
  title={Geometry-aware RL for manipulation of varying shapes and deformable objects},
  author={Hoang, Tai and Le, Huy and Becker, Philipp and Ngo, Vien A and Neumann, Gerhard},
  booktitle={International Conference on Learning Representations},
  volume={2025},
  pages={47376--47405},
  year={2025}
}

@inproceedings{xing2025stabilizing,
  title={Stabilizing reinforcement learning in differentiable multiphysics simulation},
  author={Xing, Eliot and Luk, Vernon and Oh, Jean},
  booktitle={International Conference on Learning Representations},
  volume={2025},
  pages={91165--91198},
  year={2025}
}

@misc{intrinsic2026aic,
  author       = {{Intrinsic}},
  title        = {{AI for Industry Challenge Toolkit}},
  year         = {2026},
  howpublished = {\url{https://github.com/intrinsic-dev/aic}},
  note         = {Accessed: 2026-10-08}
}

\newpage
\appendix

\section{Related Work}
\label{appendix:related_work}
We expand the related-work discussion from the main paper with a feature-level comparison against prior DLO and deformable-object benchmarks, summarized in Table~\ref{tab:bench-comparison}. For each benchmark we mark which of the three industrial task families it covers, what wire physics it uses, which learning baselines it supports, and whether it offers simulated robot evaluation, physical robot validation, and multi-source demonstrations. Most prior benchmarks address only a subset of these dimensions. WireCraft is the only one that spans all three task families with both articulated and deformable wire models, supports RL, IL, and VLA baselines together, and pairs simulated evaluation with physical UR5 validation.

\afterpage{
\begin{table*}[p]
\centering
\caption{\textbf{Comparison with related DLO and deformable-object benchmarks.}
WireCraft combines industrial DLO task families, articulated and FEM-based deformable
wire simulation, RL/IL/VLA baselines, simulated-robot evaluation, and physical UR5
validation within a unified benchmark framework. \textit{Insertion}, \textit{Routing},
and \textit{Seating} refer to end/part insertion, fixture-guided routing, and placement
into extended grooves or channels, respectively. \textit{Multi-Tier} indicates multiple
task levels, task families, or difficulty settings; \textit{Multi-Source} indicates
demonstrations or trajectories from multiple sources under a shared schema. Deformable
cells report the specified physics representation when available. \cmark: supported.
\xmark: not supported. \nmark: not applicable. Benchmarks are sorted by publication year.}
\label{tab:bench-comparison}
\renewcommand{\arraystretch}{1.15}
\setlength{\tabcolsep}{6pt}
\resizebox{\textwidth}{!}{%
\begin{tabular}{>{\bfseries}l cccc cc c ccccc}
\toprule
Benchmark
& \rot{Insertion}
& \rot{Routing}  
& \rot{Seating}
& \rot{Multi-Tier}
& \rot{Articulated Wire}
& \rot{Deformable Wire}
& \rot{RL Baseline}
& \rot{IL Baseline}
& \rot{VLA Baseline}
& \rot{Simulated Robot}
& \rot{Physical Robot}
& \rot{Multi-Source}
 \\
\midrule
RLBench~\cite{james2020rlbench}~{\scriptsize(2020)}
  & \cmark & \xmark & \xmark & \xmark & \xmark & \xmark
  & \xmark & \xmark & \xmark & \cmark & \xmark & \xmark \\
Zhou et al.~\cite{zhou2020dlocableharness}~{\scriptsize(2020)}
  & \cmark & \xmark & \xmark & \xmark & \nmark & \nmark
  & \xmark & \xmark & \xmark & \xmark & \cmark & \xmark \\
DeformableRavens~\cite{seita2021learning}~{\scriptsize(2021)}
  & \xmark & \xmark & \cmark & \cmark & \cmark & \cmarkm{FEM}
  & \xmark & \cmark & \xmark & \cmark & \cmark & \xmark \\
PlasticineLab~\cite{huang2021plasticinelab}~{\scriptsize(2021)}
  & \xmark & \cmark & \xmark & \cmark & \xmark & \cmarkm{MPM}
  & \cmark & \xmark & \xmark & \cmark & \xmark & \xmark \\
ReForm~\cite{9561766}~{\scriptsize(2021)}
  & \cmark & \cmark & \xmark & \cmark & \cmark & \cmarkm{AGX/LEM}
  & \cmark & \xmark & \xmark & \cmark & \xmark & \xmark \\
SoftGym~\cite{lin2021softgym}~{\scriptsize(2021)}
  & \xmark & \xmark & \xmark & \cmark & \xmark & \cmarkm{PBD}
  & \cmark & \xmark & \xmark & \cmark & \cmark & \xmark \\
NIST ATB~\cite{kimble2022performance}~{\scriptsize(2022)}
  & \cmark & \cmark & \xmark & \cmark & \nmark & \nmark
  & \xmark & \xmark & \xmark & \xmark & \cmark & \xmark \\
Jin et al.~\cite{9732654}~{\scriptsize(2022)}
  & \cmark & \cmark & \xmark & \cmark & \nmark & \nmark
  & \xmark & \cmark & \xmark & \xmark & \cmark & \xmark \\
Zhao et al.~\cite{zhao2022offline}~{\scriptsize(2022)}
  & \cmark & \xmark & \xmark & \cmark & \nmark & \nmark
  & \cmark & \cmark & \xmark & \xmark & \cmark & \cmark \\
Chang et al.~\cite{9926677}~{\scriptsize(2022)}
  & \cmark & \cmark & \xmark & \cmark & \nmark & \nmark
  & \xmark & \xmark & \xmark & \xmark & \cmark & \xmark \\
DaXBench~\cite{chendaxbench}~{\scriptsize(2023)}
  & \xmark & \xmark & \xmark & \cmark & \xmark & \cmarkm{MPM/MS}
  & \cmark & \cmark & \xmark & \cmark & \cmark & \xmark \\
Wilson et al.~\cite{wilson2023cable}~{\scriptsize(2023)}
  & \cmark & \cmark & \xmark & \cmark & \nmark & \nmark
  & \xmark & \xmark & \xmark & \xmark & \cmark & \xmark \\
Luo et al.~\cite{luo2024multistage}~{\scriptsize(2024)}
  & \cmark & \cmark & \xmark & \cmark & \nmark & \nmark
  & \xmark & \cmark & \xmark & \xmark & \cmark & \xmark \\
DexDLO~\cite{sun2024dexdlo}~{\scriptsize(2024)}
  & \xmark & \xmark & \xmark & \cmark & \cmark & \cmarkm{ART}
  & \cmark & \xmark & \xmark & \cmark & \xmark & \xmark \\
Li and Choi~\cite{Li_2024}~{\scriptsize(2024)}
  & \cmark & \xmark & \xmark & \cmark & \xmark & \cmarkm{PBD}
  & \cmark & \xmark & \xmark & \cmark & \cmark & \xmark \\
Zhang et al.~\cite{zhang2024harnessing}~{\scriptsize(2024)}
  & \cmark & \cmark & \xmark & \cmark & \nmark & \nmark
  & \xmark & \xmark & \xmark & \xmark & \cmark & \xmark \\
Chen et al.~\cite{chen2023dermujoco}~{\scriptsize(2025)}
  & \xmark & \xmark & \xmark & \xmark
  & \cmark & \cmarkm{DER}\hybrid
  & \xmark & \xmark & \xmark & \cmark & \cmark & \xmark \\
DEFORM~\cite{chen2024deform}~{\scriptsize(2025)}
  & \xmark & \xmark & \xmark & \xmark & \xmark & \cmarkm{DER}
  & \xmark & \xmark & \xmark & \cmark & \cmark & \xmark \\
Yu et al.~\cite{yu2025generalizable}~{\scriptsize(2025)}
  & \xmark & \xmark & \xmark & \cmark & \xmark & \cmarkm{DER}
  & \xmark & \xmark & \xmark & \cmark & \cmark & \xmark \\
ManiSkill3~\cite{taomaniskill3}~{\scriptsize(2025)}
  & \cmark & \xmark & \xmark & \cmark & \xmark & \xmark
  & \cmark & \cmark & \cmark & \cmark & \cmark & \cmark \\
DEFT~\cite{chen2025deft}~{\scriptsize(2025)}
  & \cmark & \xmark & \xmark & \cmark & \xmark & \cmarkm{B-DER}
  & \xmark & \xmark & \xmark & \xmark & \cmark & \xmark \\
Govoni et al.~\cite{govoni2025dlo}~{\scriptsize(2025)}
  & \xmark & \xmark & \xmark & \xmark & \xmark & \cmarkm{MSD}
  & \xmark & \xmark & \xmark & \cmark & \xmark & \xmark \\
Li et al.~\cite{11127451}~{\scriptsize(2025)}
  & \cmark & \cmark & \xmark & \cmark & \xmark & \cmarkm{PBD}
  & \cmark & \cmark & \xmark & \cmark & \cmark & \xmark \\
Li et al.~\cite{li2025hierarchicaldloroutingreinforcement}~{\scriptsize(2025)}
  & \cmark & \cmark & \xmark & \cmark & \xmark & \cmarkm{Particle}
  & \cmark & \xmark & \xmark & \cmark & \cmark & \xmark \\
Kienle et al.~\cite{kienle2025ai}~{\scriptsize(2025)}
  & \cmark & \xmark & \xmark & \cmark & \nmark & \nmark
  & \xmark & \xmark & \xmark & \xmark & \cmark & \xmark \\
Tanureza et al.~\cite{tanureza2025industrial}~{\scriptsize(2025)}
  & \cmark & \cmark & \xmark & \cmark & \nmark & \nmark
  & \xmark & \xmark & \xmark & \xmark & \cmark & \xmark \\
Li et al.~\cite{10.1145/3776942.3777006}~{\scriptsize(2025)}
  & \xmark & \cmark & \xmark & \xmark & \xmark & \cmarkm{MSD}
  & \xmark & \xmark & \xmark & \cmark & \xmark & \xmark \\
Azulay et al.~\cite{azulay2025motorcycle}~{\scriptsize(2025)}
  & \xmark & \cmark & \xmark & \cmark & \nmark & \nmark
  & \xmark & \xmark & \xmark & \xmark & \cmark & \xmark \\
MoDeSuite~\cite{zhang2026modesuite}~{\scriptsize(2026)}
  & \xmark & \xmark & \xmark & \cmark & \xmark & \cmarkm{FEM/PBD}
  & \cmark & \cmark & \xmark & \cmark & \cmark & \xmark \\
DLO-Lab~\cite{cao2026dlolab}~{\scriptsize(2026)}
  & \cmark & \cmark & \xmark & \cmark & \xmark & \cmarkm{Diff. DER}
  & \cmark & \xmark & \xmark & \cmark & \cmark & \xmark \\
Cao et al.~\cite{Cao2026RoboticCF}~{\scriptsize(2026)}
  & \xmark & \cmark & \xmark & \cmark & \nmark & \nmark
  & \xmark & \xmark & \xmark & \xmark & \cmark & \xmark \\
\midrule
\textbf{WireCraft (Ours)}~{\scriptsize\textbf{(2026)}}
  & \cmark & \cmark & \cmark & \cmark
  & \cmark & \cmarkm{FEM}
  & \cmark
  & \cmark & \cmark & \cmark & \cmark & \cmark \\
  
\bottomrule

\end{tabular}}
\end{table*}
}

\section{DLO Physics and Simulation Details}
\label{appendix:physics}

This section expands on the DLO physics summarized in Section~\ref{subsection:wirecraft engine}. WireCraft is built on Isaac Sim 4.5~\cite{isaacsim} and Isaac Lab v2.2.1~\cite{mittal2025isaaclab}, and is additionally verified on Isaac Sim 5.1. We first describe the simulation backend, then the two wire models and their material and contact parameters, the DLO--connector interaction, and finally the computational scaling behavior of each model.

\subsection{Simulation Backend}
\label{appendix:backend}

WireCraft runs on Isaac Sim for photo-realistic rendering and physics, and on Isaac Lab for GPU-accelerated, multi-instance parallel simulation used in both policy training and trajectory generation. The default RL configuration uses a PhysX physics timestep of $1/120$~s with a control decimation of $2$, giving a $60$~Hz control rate and an $8$~s episode horizon. We verified WireCraft across multiple GPUs, namely the NVIDIA GeForce RTX~4080 SUPER, the RTX PRO 4500 Blackwell, and the RTX A4000 and RTX 5070Ti. 

\subsection{DLO Models}
\label{appendix:dlo_models}

WireCraft provides two complementary wire representations for the same benchmark tasks. They are not intended to be parameter-by-parameter calibrated material matches, since the articulated chain and FEM soft body expose different solver-native parameters.  Instead, they are aligned at the task level through shared wire endpoints, connector interfaces, fixtures, observation-action schemas, and success criteria. The articulated wire is a chain of rigid capsule segments connected by 6-DoF joints (PhysX D6 joints) whose angular degrees of freedom are driven by passive implicit spring--damper actuators, simulated under rigid-body dynamics. This representation supports high-throughput trajectory generation but can become unstable under adjacent-segment self-contact or accumulated contact forces. The deformable wire is a PhysX FEM soft body: a slender triangulated cylindrical rod (decagonal cross-section, 10 radial × 40 axial subdivisions; length 0.4 m, radius 3 mm) that PhysX voxelizes into a hexahedral FEM simulation mesh (resolution 40 along the major axis). This captures local bending, compression, and contact deformation at higher fidelity but greater cost. The parameter names reflect each model's native representation and should not be read as physically equivalent material settings.

\begin{table}[h]
\centering
\small
\caption{\textbf{Representative wire model parameters.} Values reflect each model's native representation and should be interpreted as solver-specific benchmark settings rather than physically equivalent material parameters. Articulated wire values correspond to the committed task-environment
configuration.}
\label{tab:wire_model_params}
\begin{tabularx}{\linewidth}{p{0.22\linewidth} X X}
\toprule
\textbf{Setting} & \textbf{FEM-based Deformable Wire} & \textbf{Articulated Wire} \\ \midrule

Representation
& Isaac Sim deformable body volume (PhysX FEM)
& Rigid cylinder chain with D6 joints \\

Total length
& 0.4 m
& 0.4 m \\

Segments / resolution
& 40 segments; hexahedral resolution 40
& 12 rigid segments; $\sim$0.033 m per segment \\

Radius / density
& 0.003 m; 800.0
& 0.004 m; 2000.0 \\

Friction
& Dynamic 5.0; static does not support deformable body
& Static 1.0; dynamic 1.0 \\

Material / joint parameters
& Young's modulus $5.0\times10^{7}$; Poisson's ratio 0.49; vertex-velocity damping 3.0
& Joint stiffness 5.0; damping 5.0; joint limit 0.5 rad \\

Solver / contact settings
& 32 position iterations; contact offset 0.001; rest offset 0.0
& 32 position iterations; 4 velocity iterations; max depenetration velocity 1.0 \\

Collision / body settings
& Collision simplification enabled; sleep threshold 0.01
& Self-collision enabled; max linear/angular velocity 5.0/60.0 \\

\bottomrule
\end{tabularx}
\end{table}

\textbf{Contact and material parameters.} Beyond the per-wire settings in Table~\ref{tab:wire_model_params}, contact behavior is governed by the connector, gripper, and table materials. The rigid connectors use a mass of $0.01$~kg and isotropic friction $2.0$ (static and dynamic, with a \texttt{max} friction-combine mode); the gripper inner-finger pads use friction $5.0$ to provide a firm grasp on the connector and wire; and the table and scene default material uses friction $1.0$. The articulated wire uses restitution $0.0$, while the deformable wire exposes no restitution term. These are fixed defaults; the connector-insertion data pipeline additionally randomizes the connector planar position and orientation and the socket clearance for demonstration diversity, as detailed in Appendix~\ref{appendix:task_randomization}.

\subsection{DLO--Connector Interaction}
\label{appendix:dlo_connector_interaction}

\begin{figure*}[h]
    \centering
    \setlength{\tabcolsep}{1pt}
    \renewcommand{\arraystretch}{0.85}

    \begin{tabular}{@{}c@{\hspace{2pt}}c@{\hspace{2pt}}c@{\hspace{2pt}}c@{\hspace{2pt}}c@{\hspace{2pt}}c@{}}

    \begin{overpic}[width=0.155\textwidth, trim=250 250 250 250, clip]{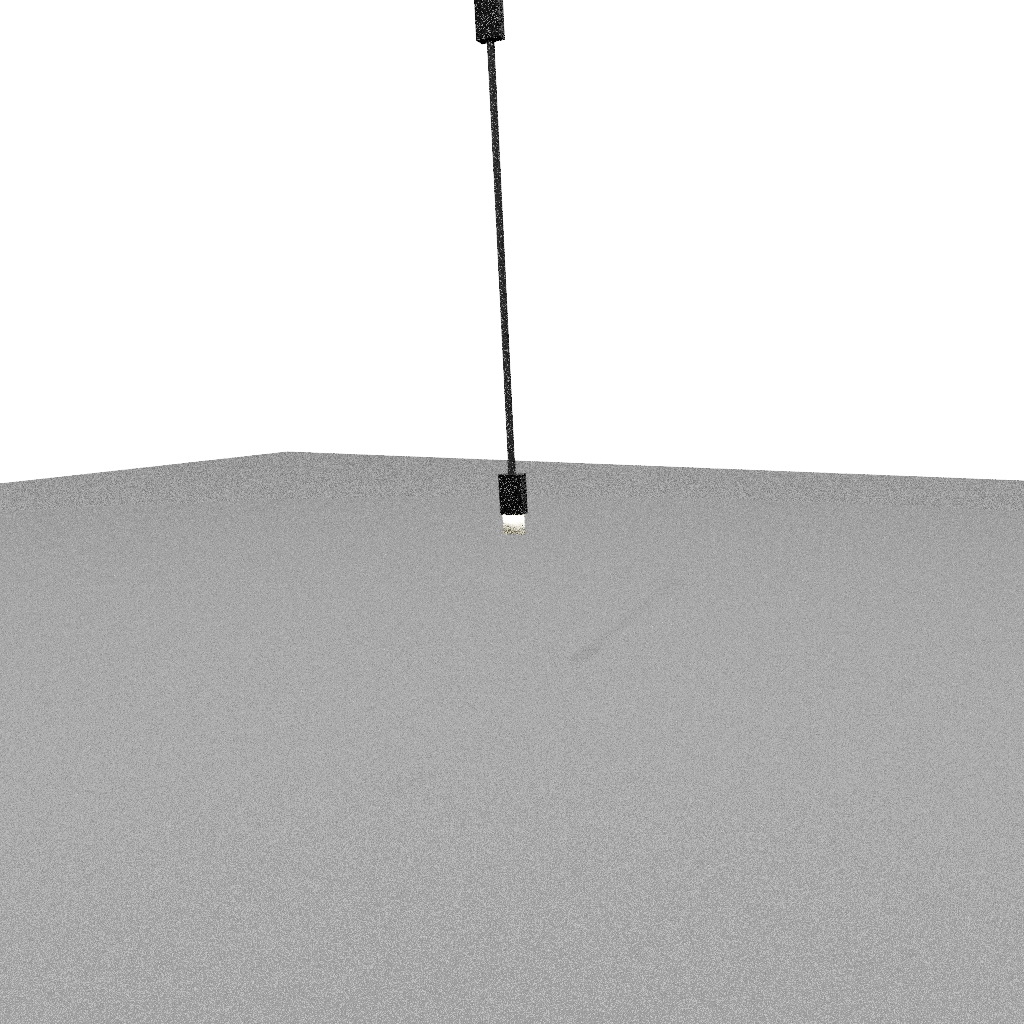}
        \put(4,4){\scriptsize\textbf{Articulation-based}}
    \end{overpic} &
    \includegraphics[width=0.155\textwidth, trim=250 250 250 250, clip]{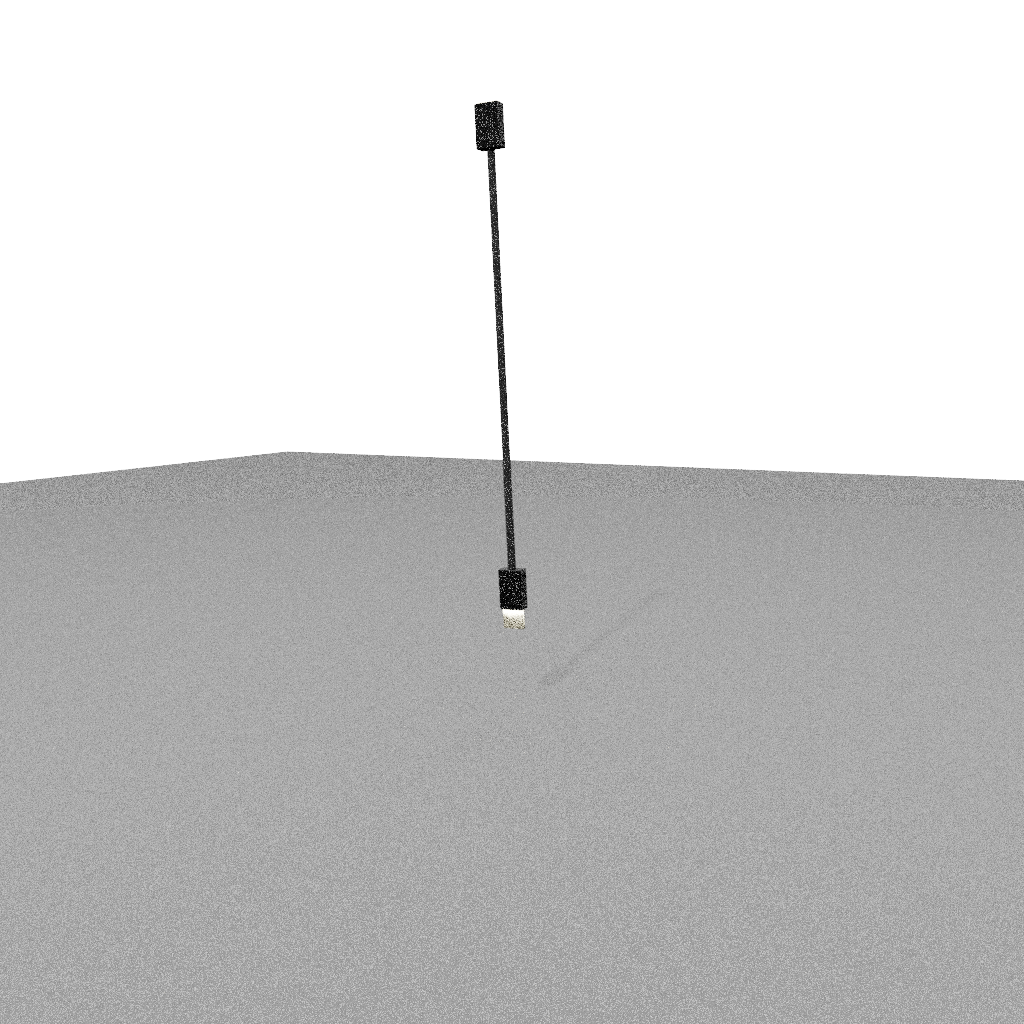} &
    \includegraphics[width=0.155\textwidth, trim=250 250 250 250, clip]{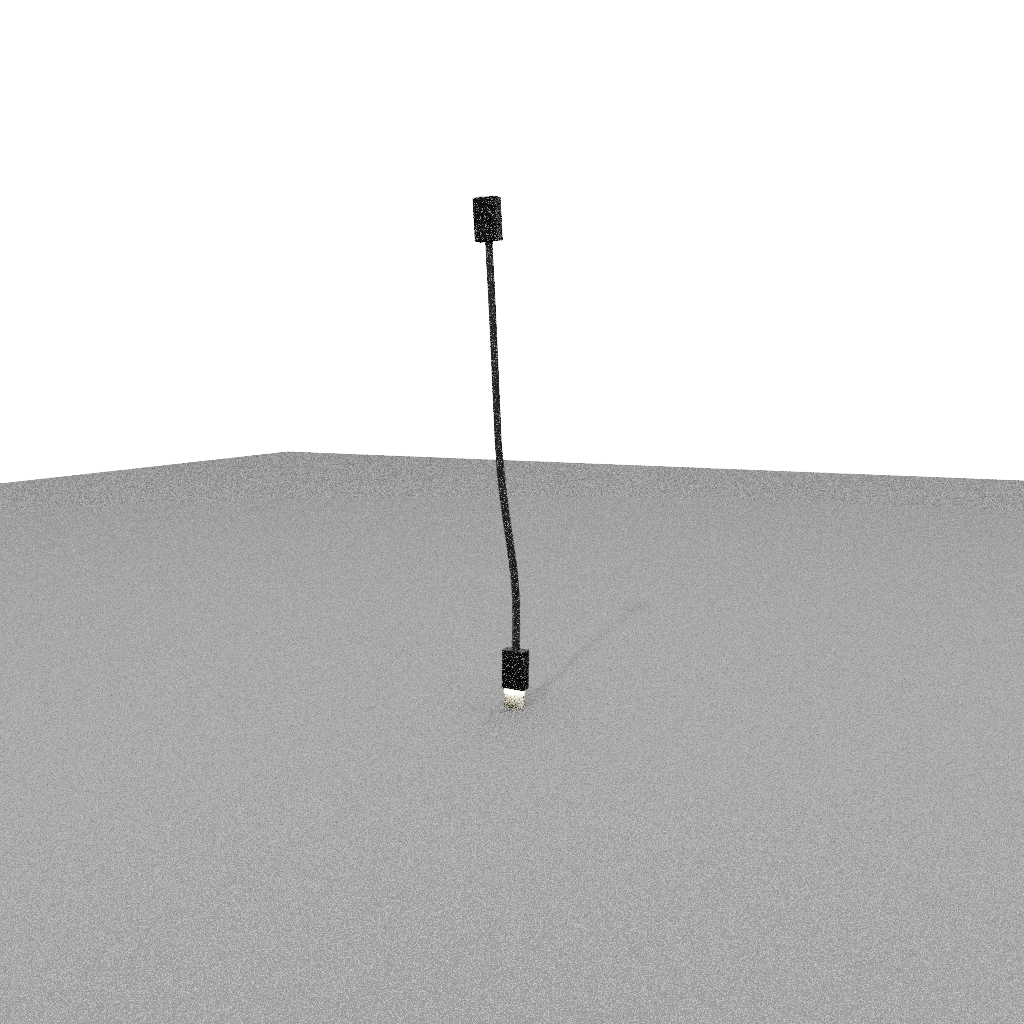} &
    \includegraphics[width=0.155\textwidth, trim=250 250 250 250, clip]{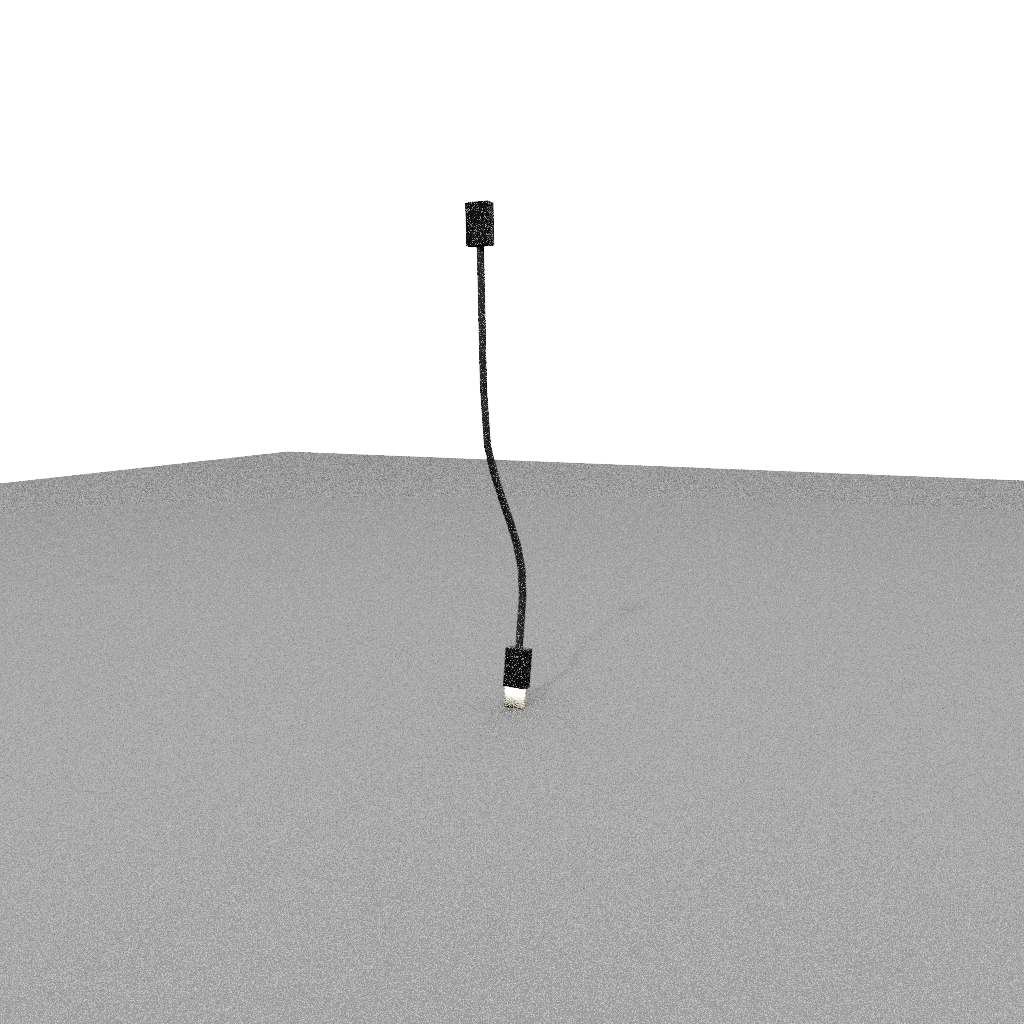} &
    \includegraphics[width=0.155\textwidth, trim=250 250 250 250, clip]{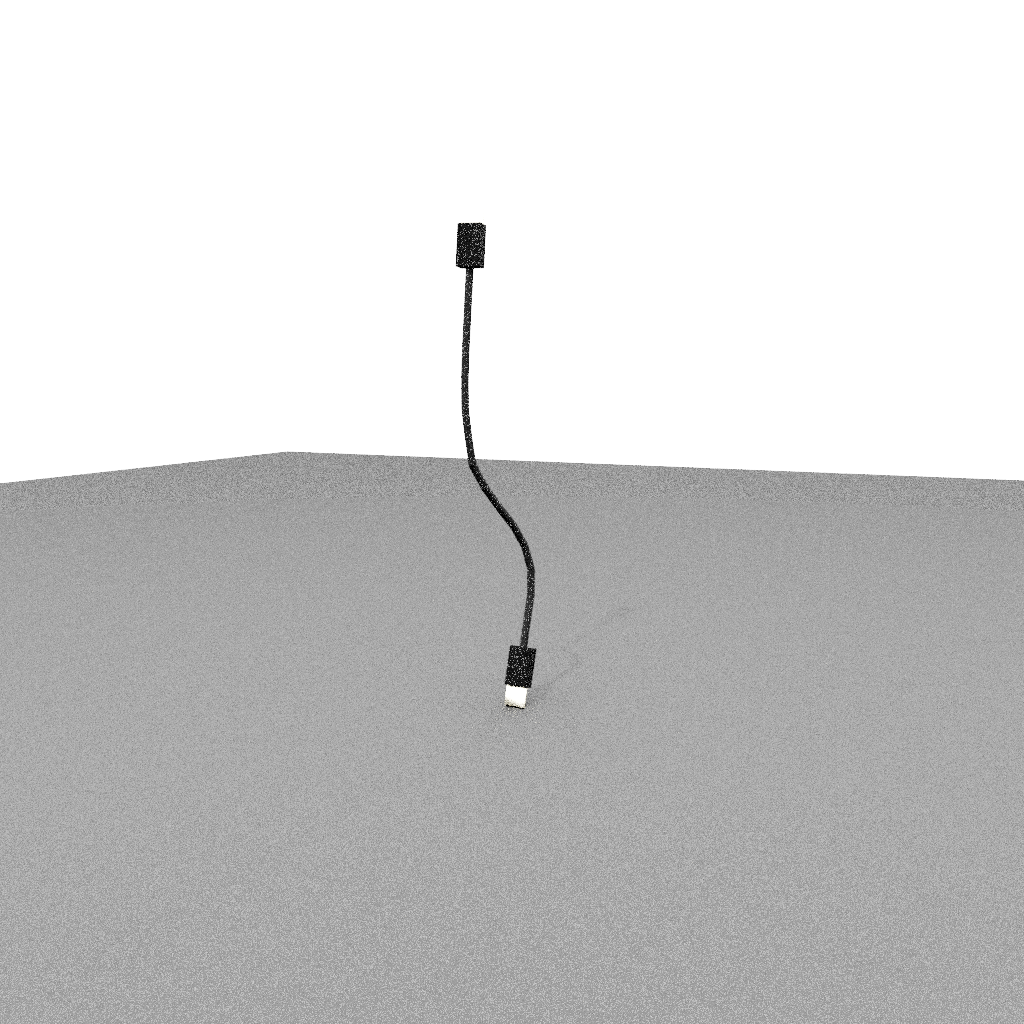} &
    \includegraphics[width=0.155\textwidth, trim=250 250 250 250, clip]{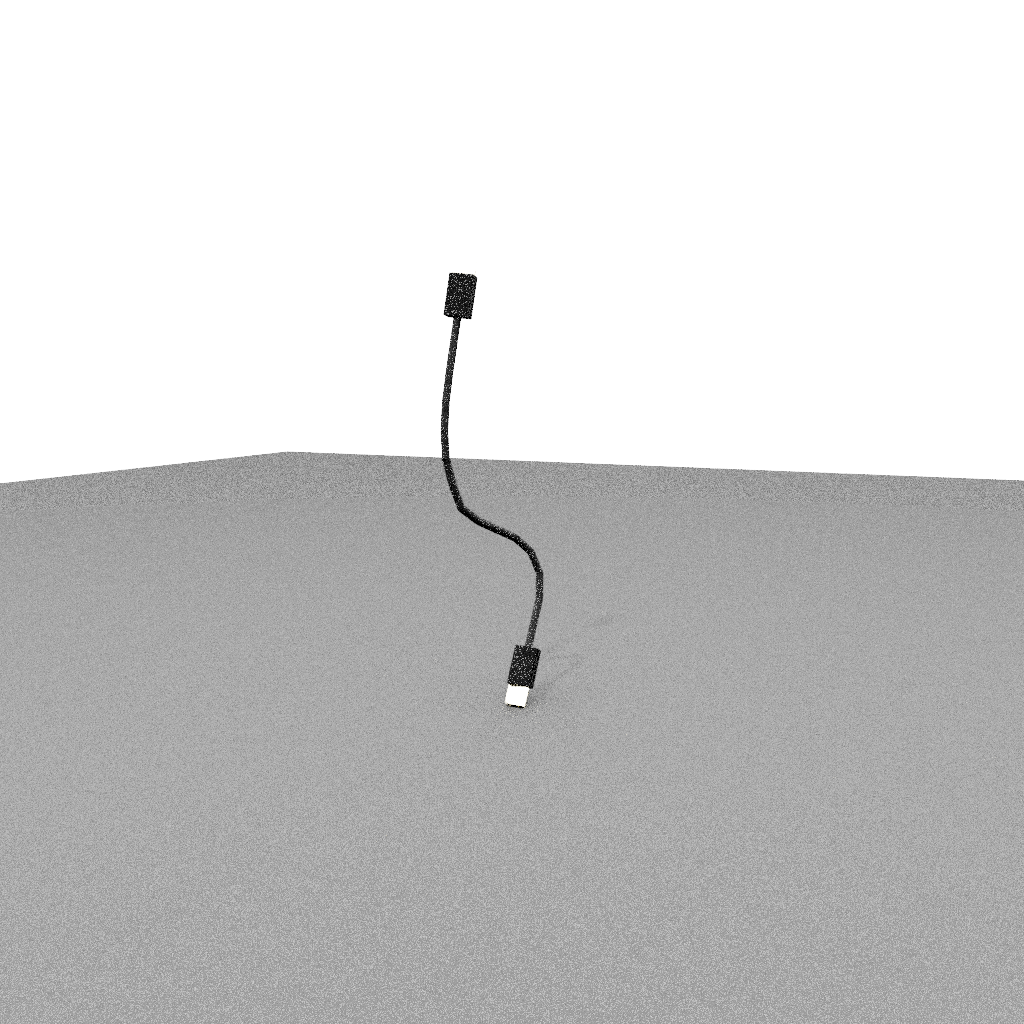}
    \\[-2pt]

    \includegraphics[width=0.155\textwidth, trim=250 250 250 250, clip]{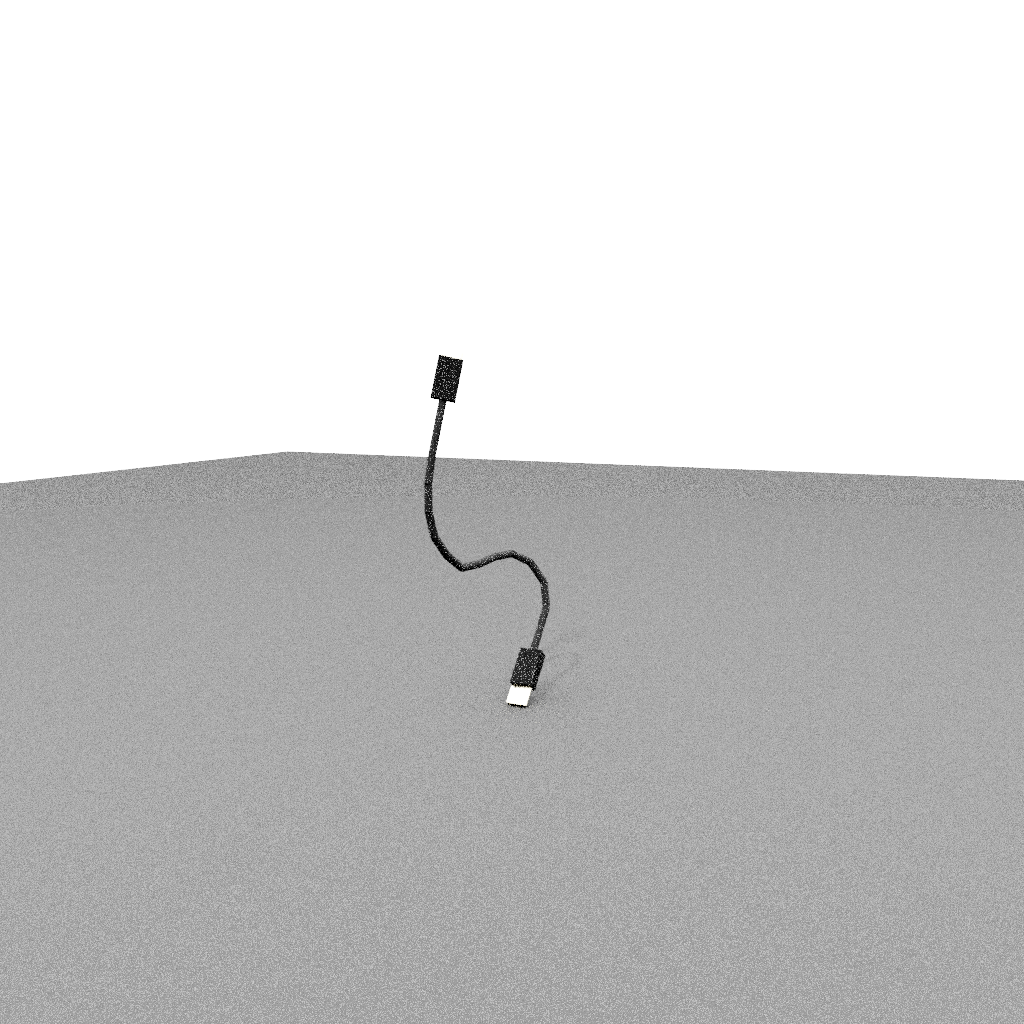} &
    \includegraphics[width=0.155\textwidth, trim=250 250 250 250, clip]{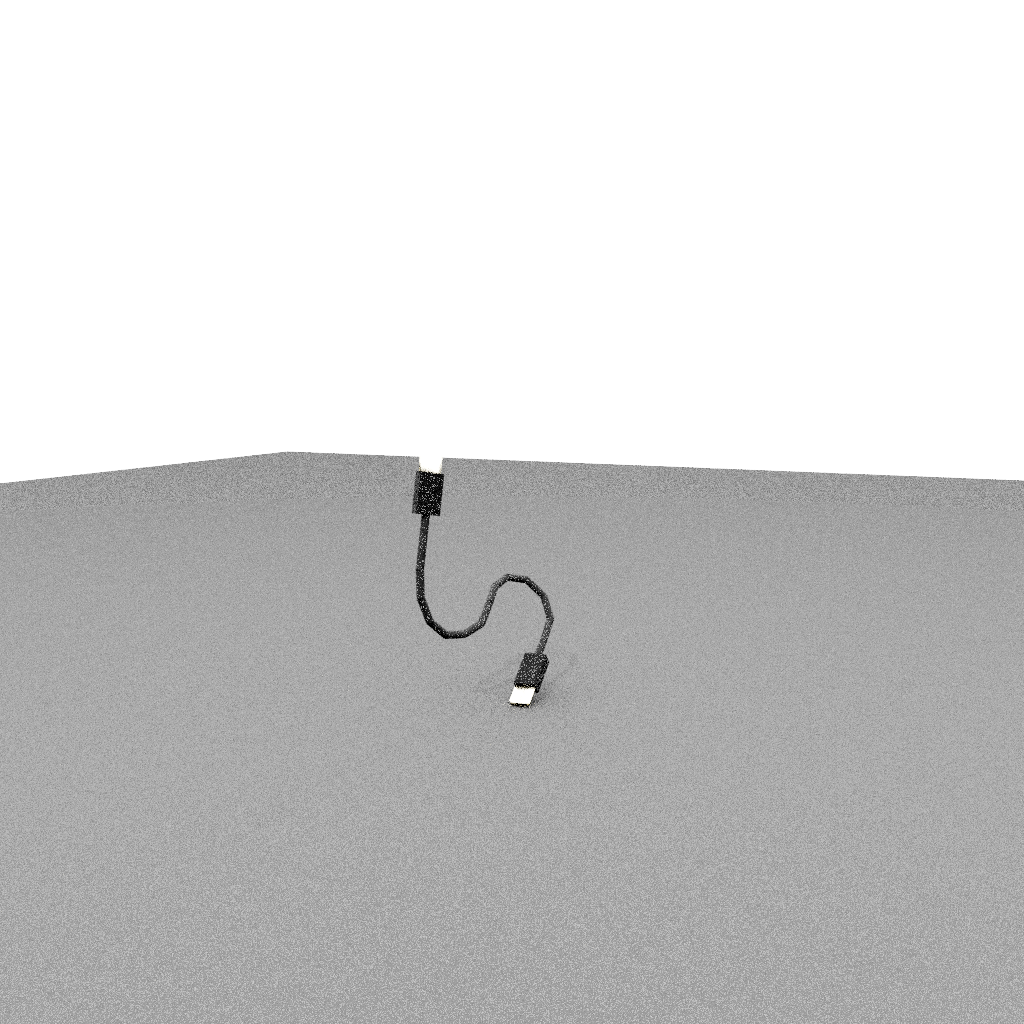} &
    \includegraphics[width=0.155\textwidth, trim=250 250 250 250, clip]{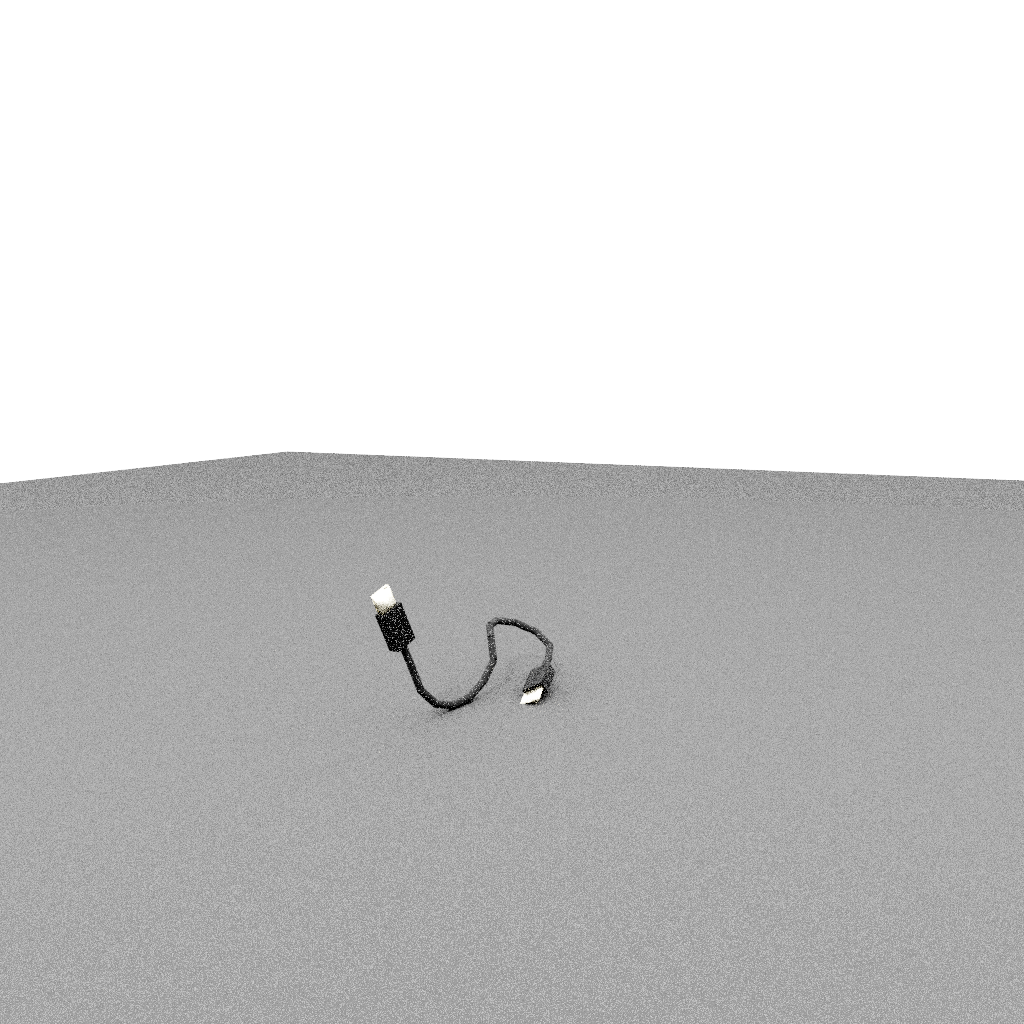} &
    \includegraphics[width=0.155\textwidth, trim=250 250 250 250, clip]{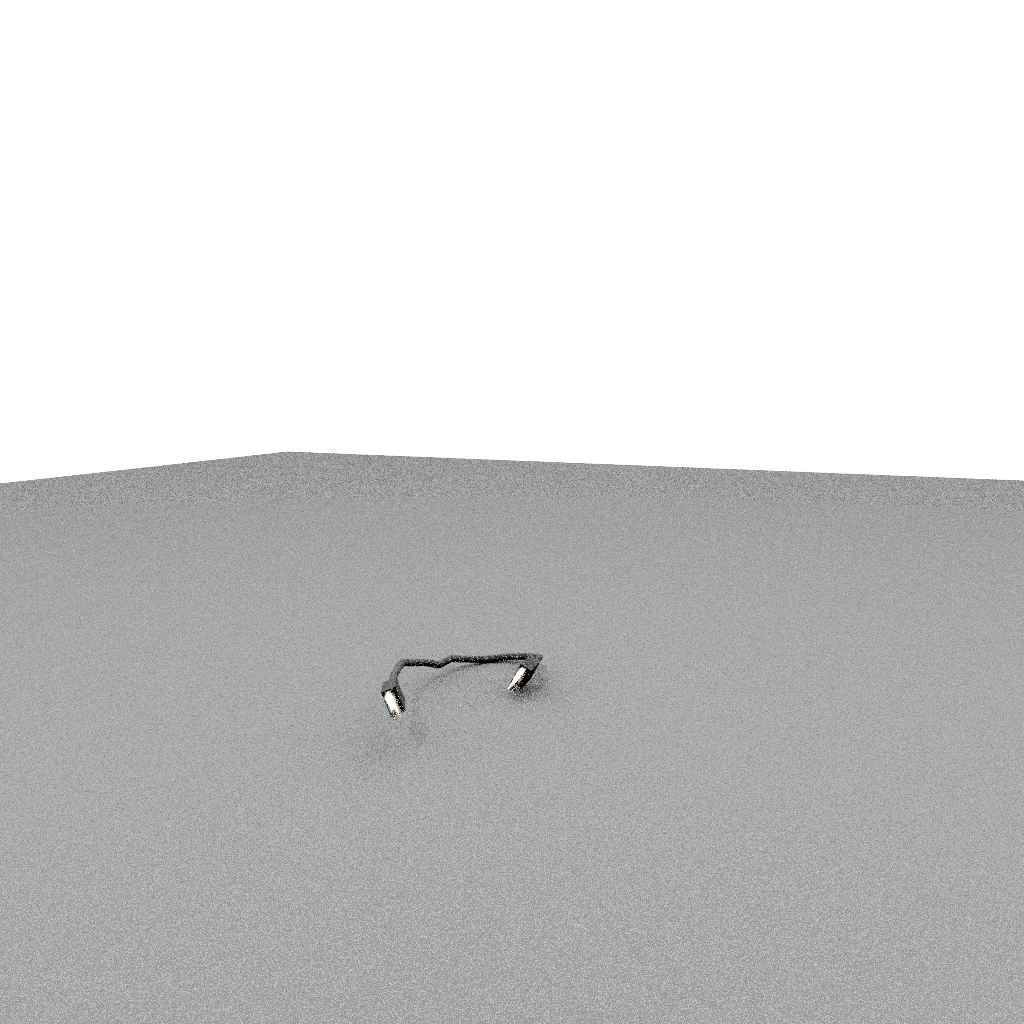} &
    \includegraphics[width=0.155\textwidth, trim=250 250 250 250, clip]{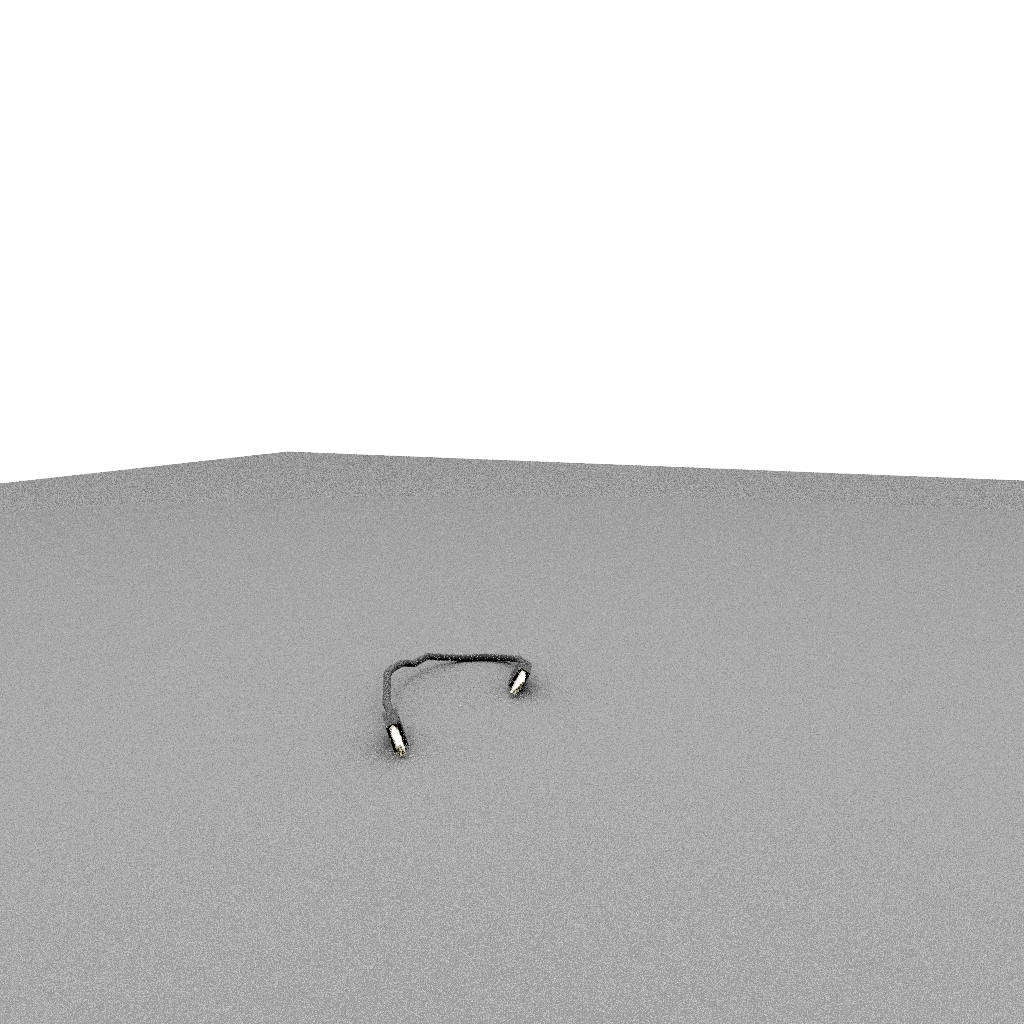} &
    \includegraphics[width=0.155\textwidth, trim=250 250 250 250, clip]{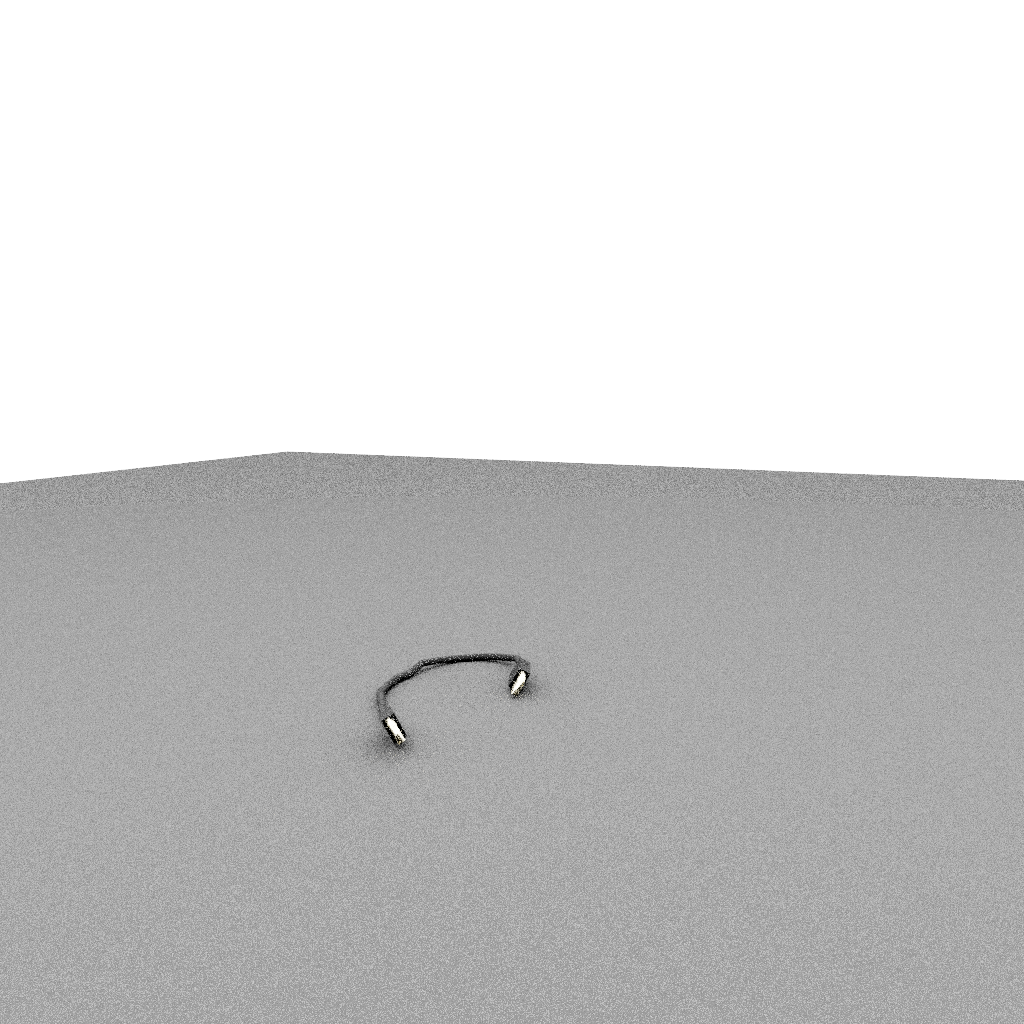}
    \\[-2pt]

    \begin{overpic}[width=0.155\textwidth, trim=250 250 250 250, clip]{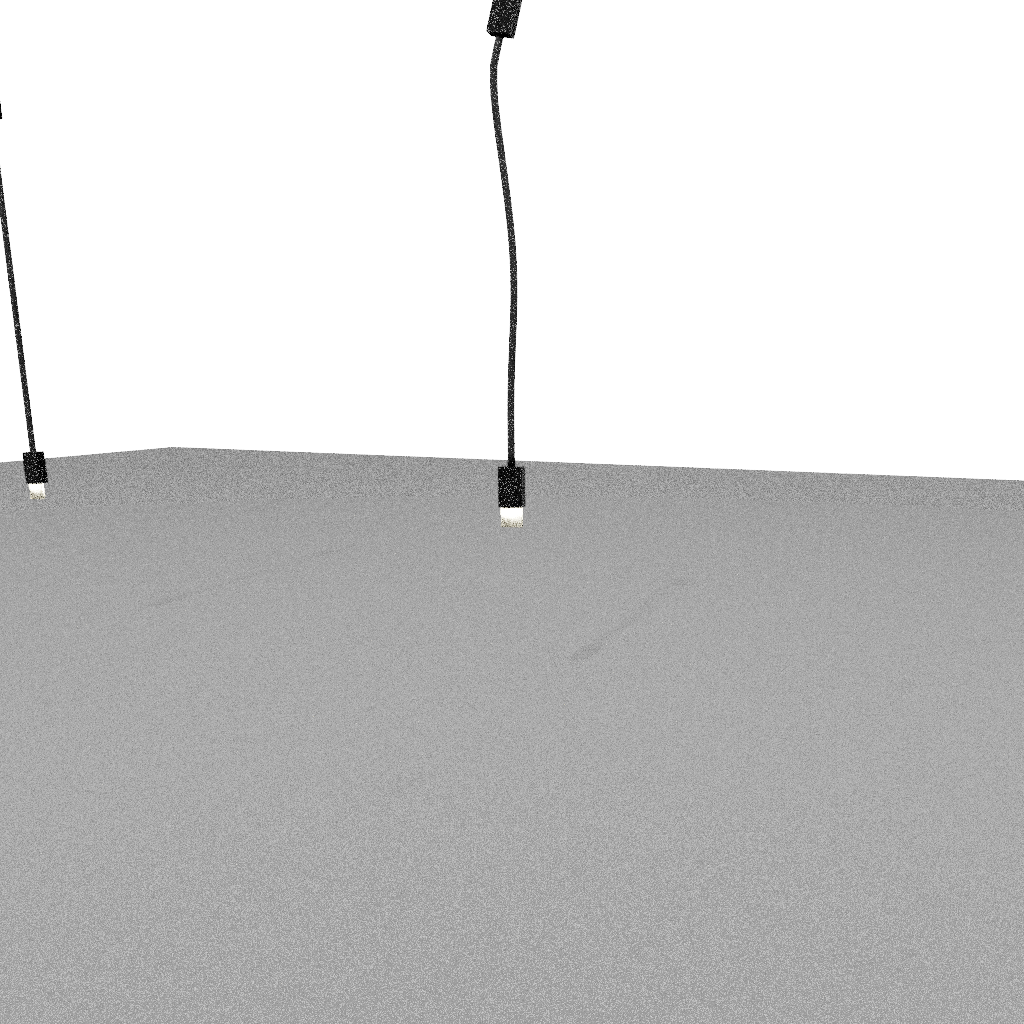}
        \put(4,4){\scriptsize\textbf{Deformable}}
    \end{overpic} &
    \includegraphics[width=0.155\textwidth, trim=250 250 250 250, clip]{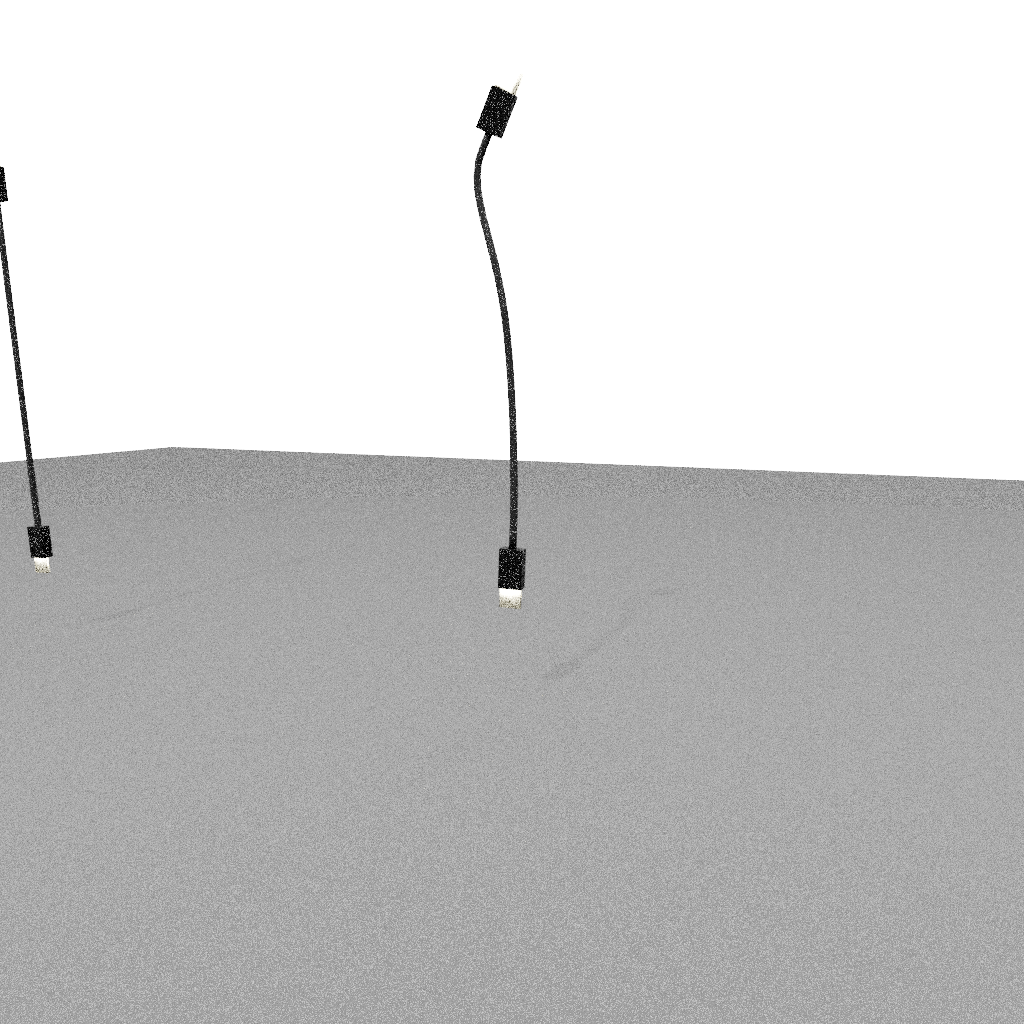} &
    \includegraphics[width=0.155\textwidth, trim=250 250 250 250, clip]{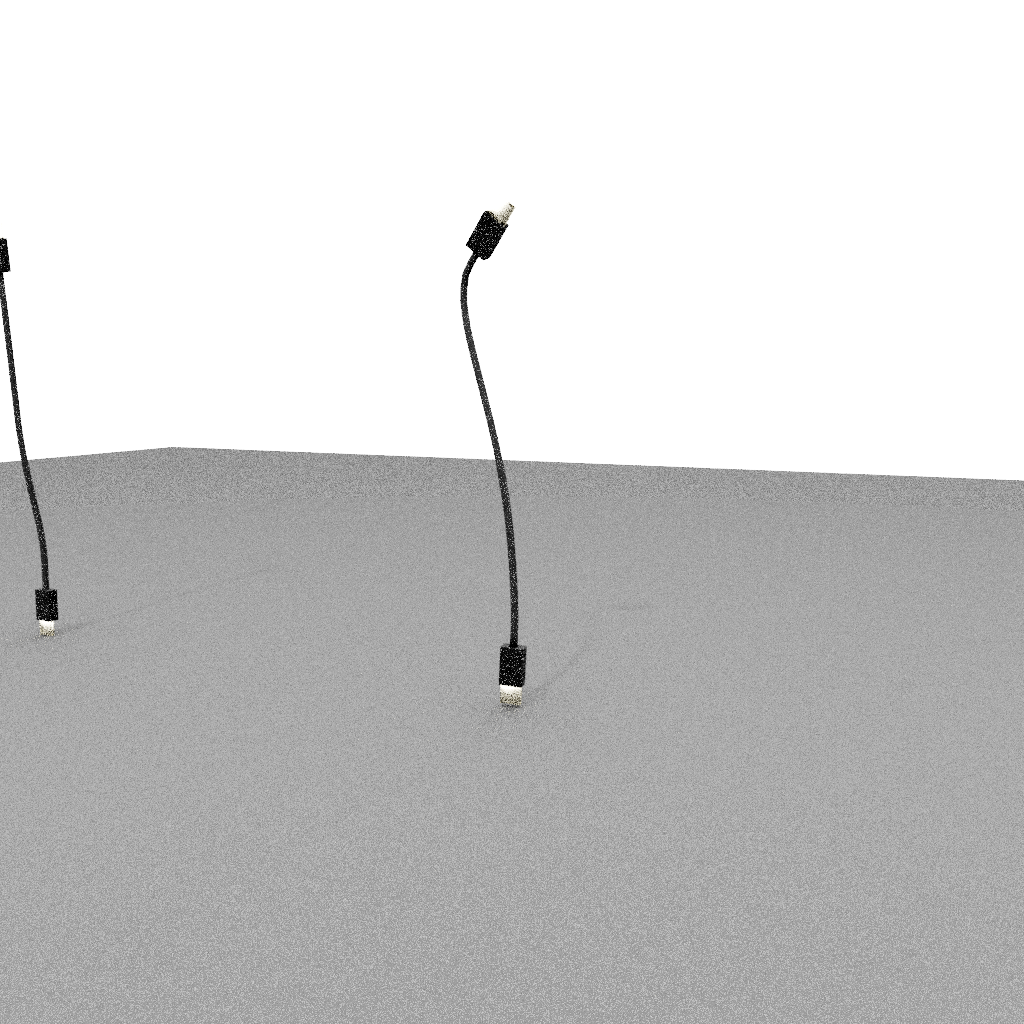} &
    \includegraphics[width=0.155\textwidth, trim=250 250 250 250, clip]{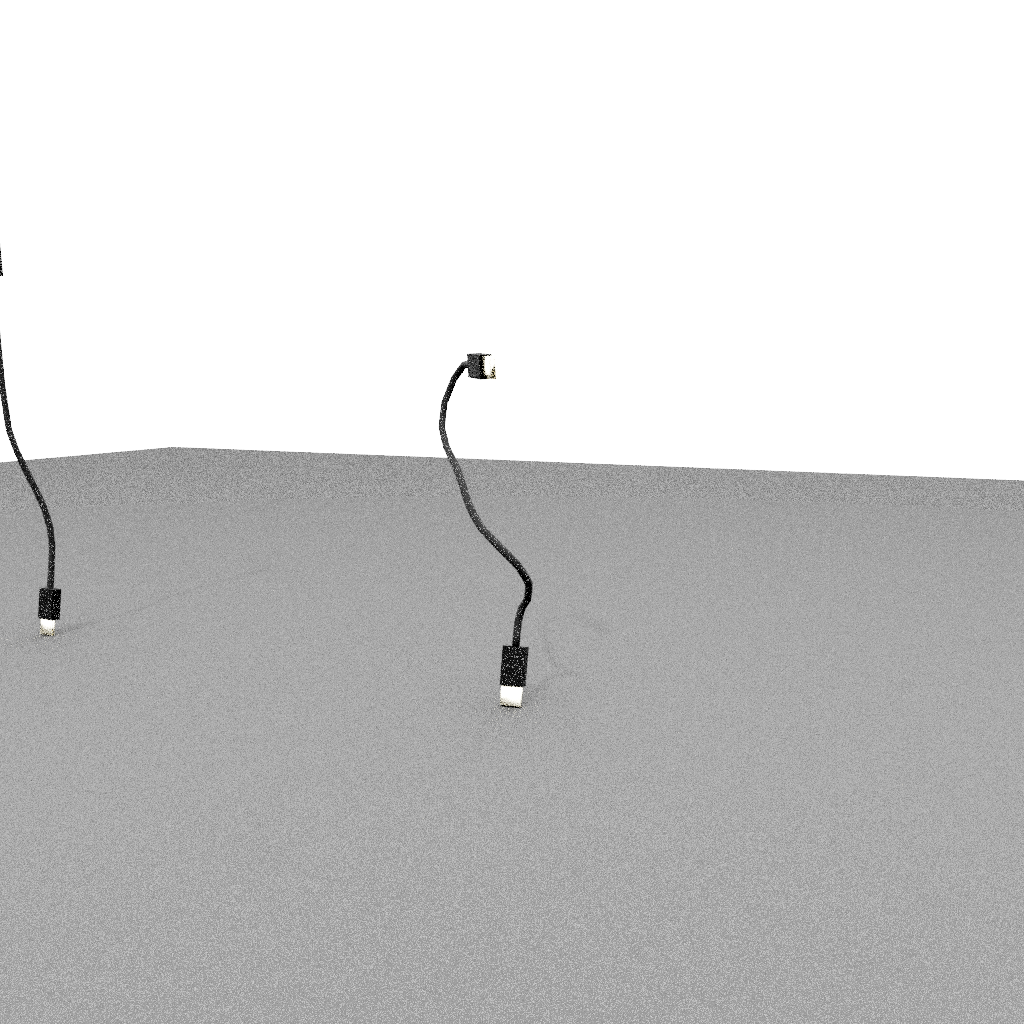} &
    \includegraphics[width=0.155\textwidth, trim=250 250 250 250, clip]{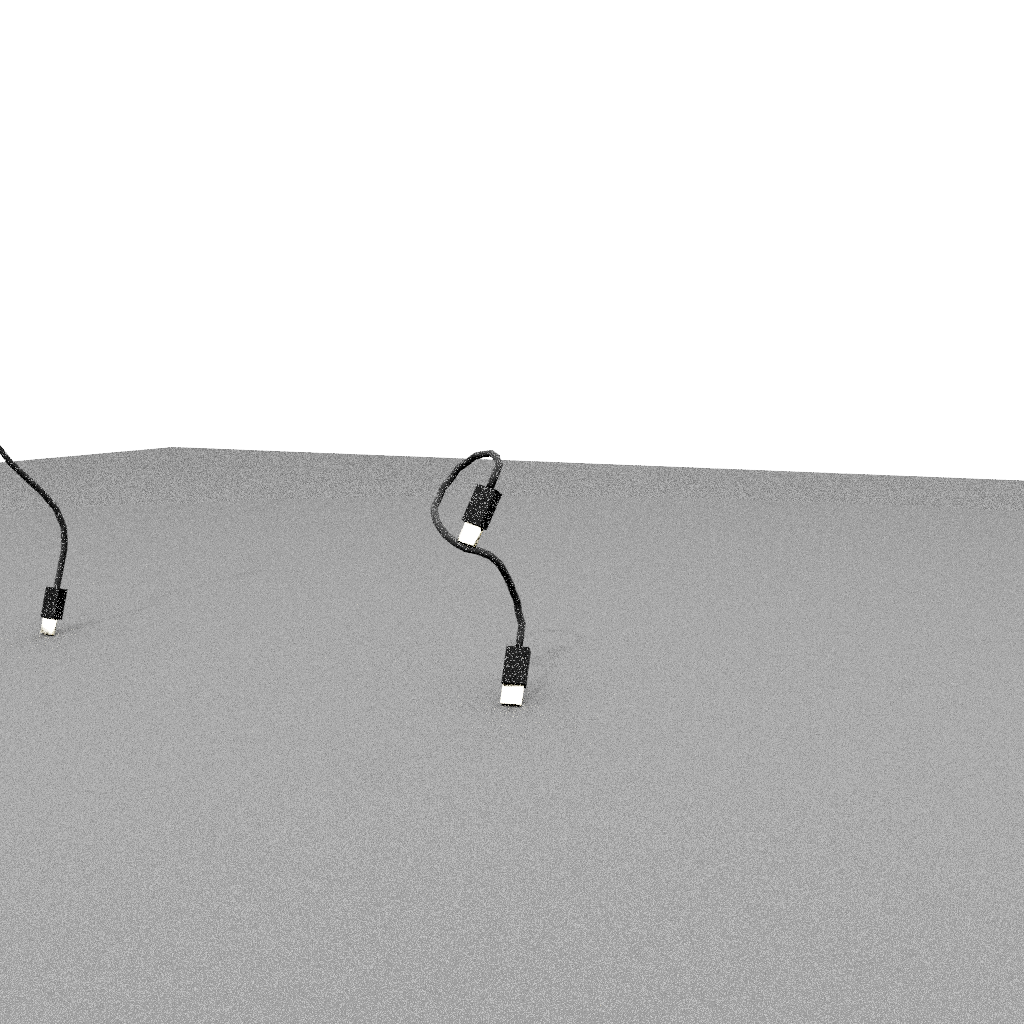} &
    \includegraphics[width=0.155\textwidth, trim=250 250 250 250, clip]{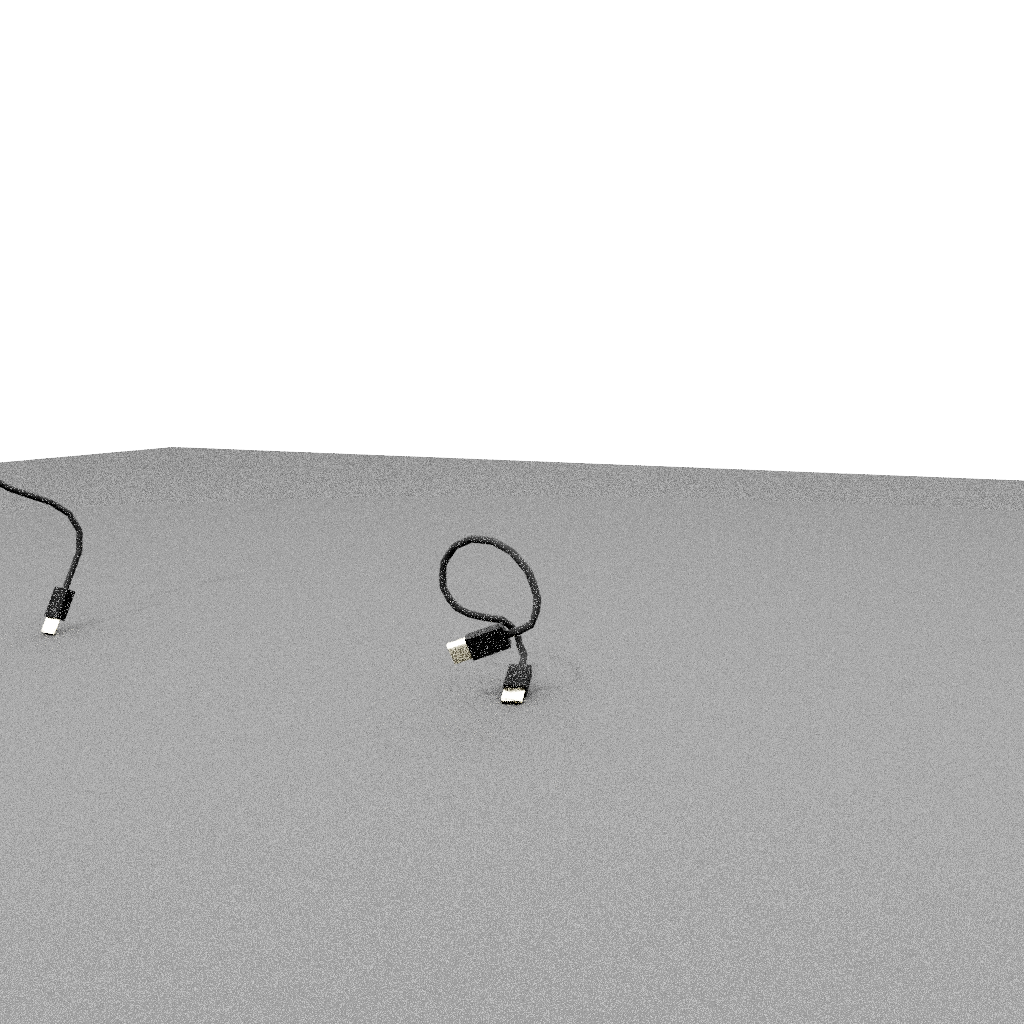}
    \\[-2pt]

    \includegraphics[width=0.155\textwidth, trim=250 250 250 250, clip]{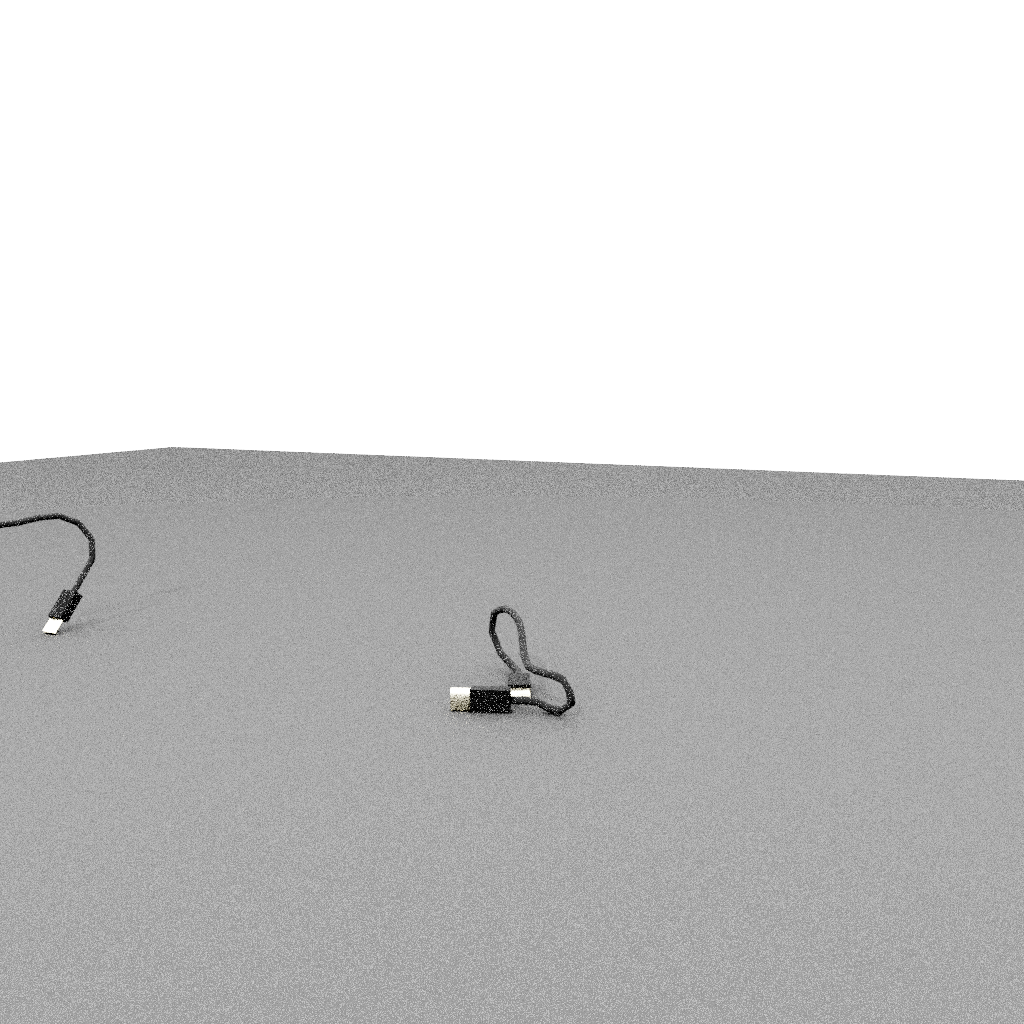} &
    \includegraphics[width=0.155\textwidth, trim=250 250 250 250, clip]{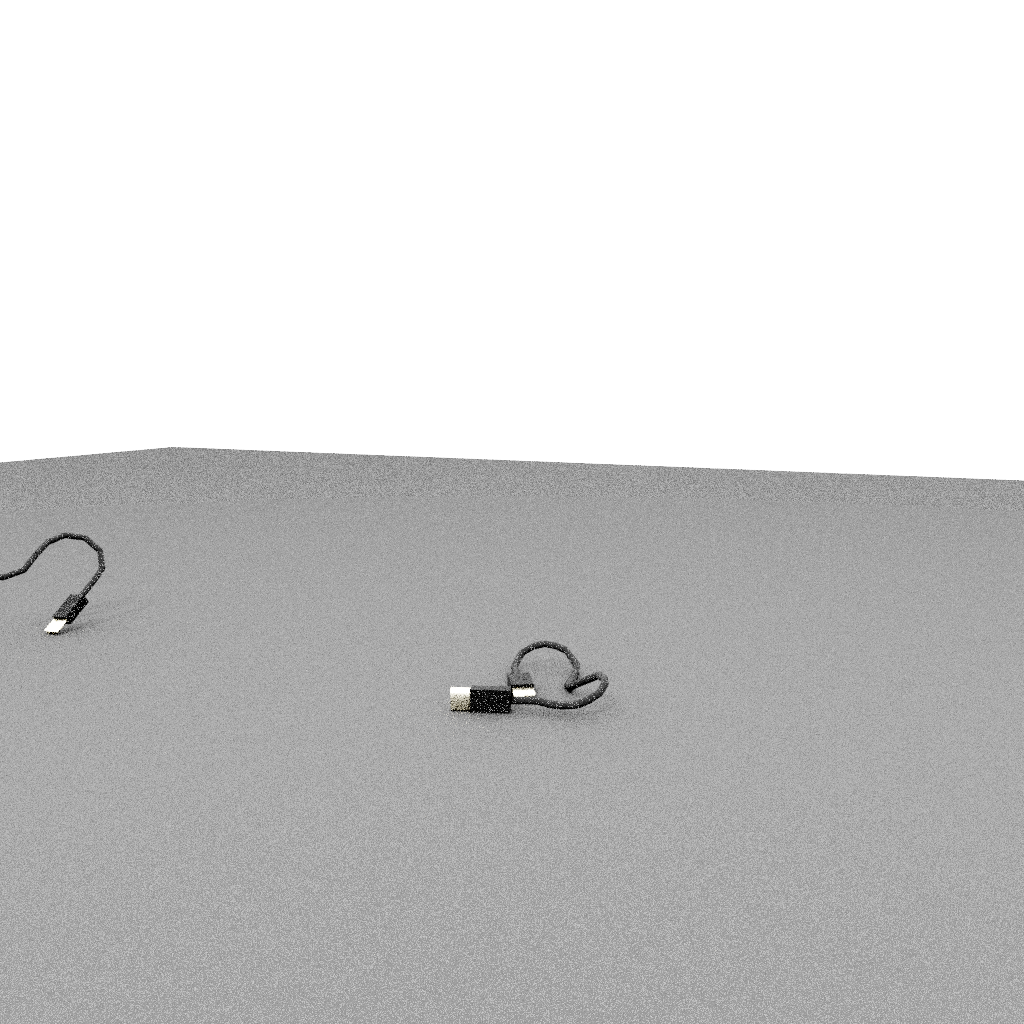} &
    \includegraphics[width=0.155\textwidth, trim=250 250 250 250, clip]{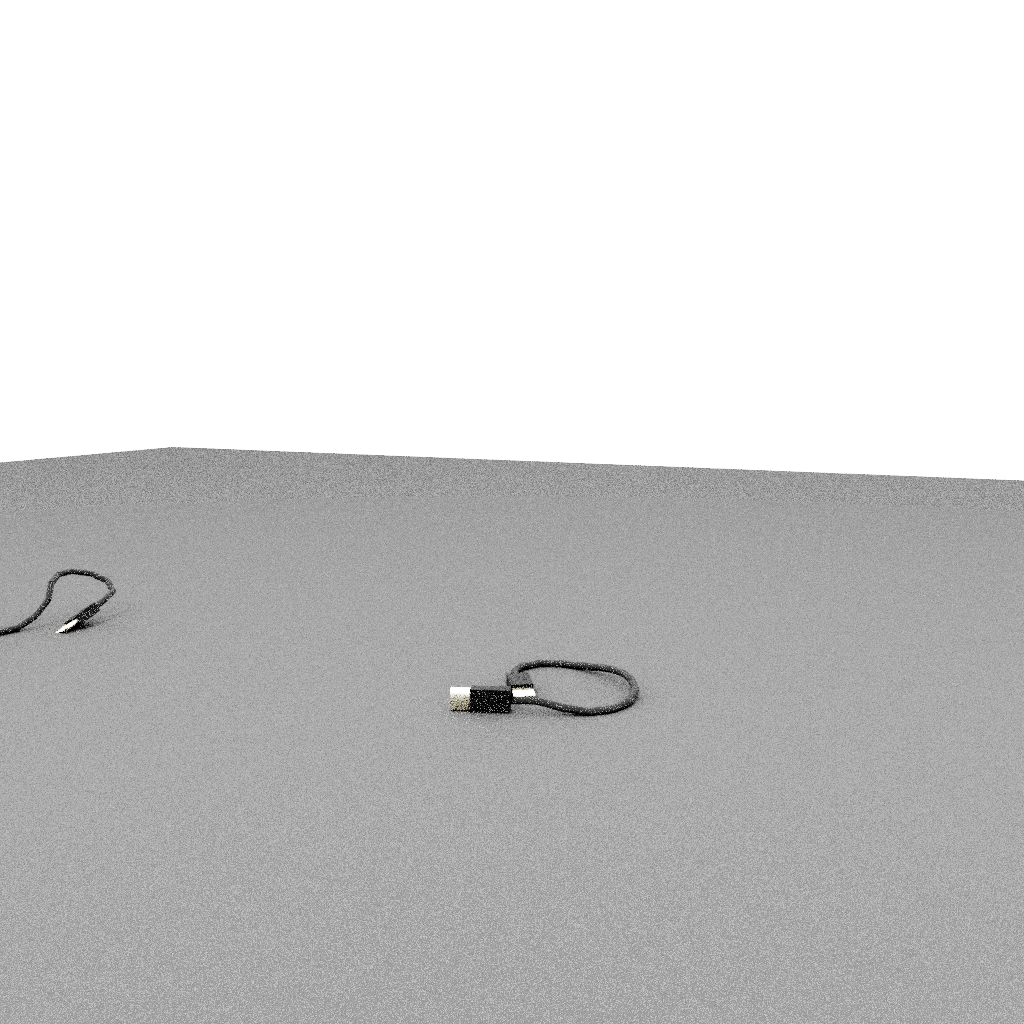} &
    \includegraphics[width=0.155\textwidth, trim=250 250 250 250, clip]{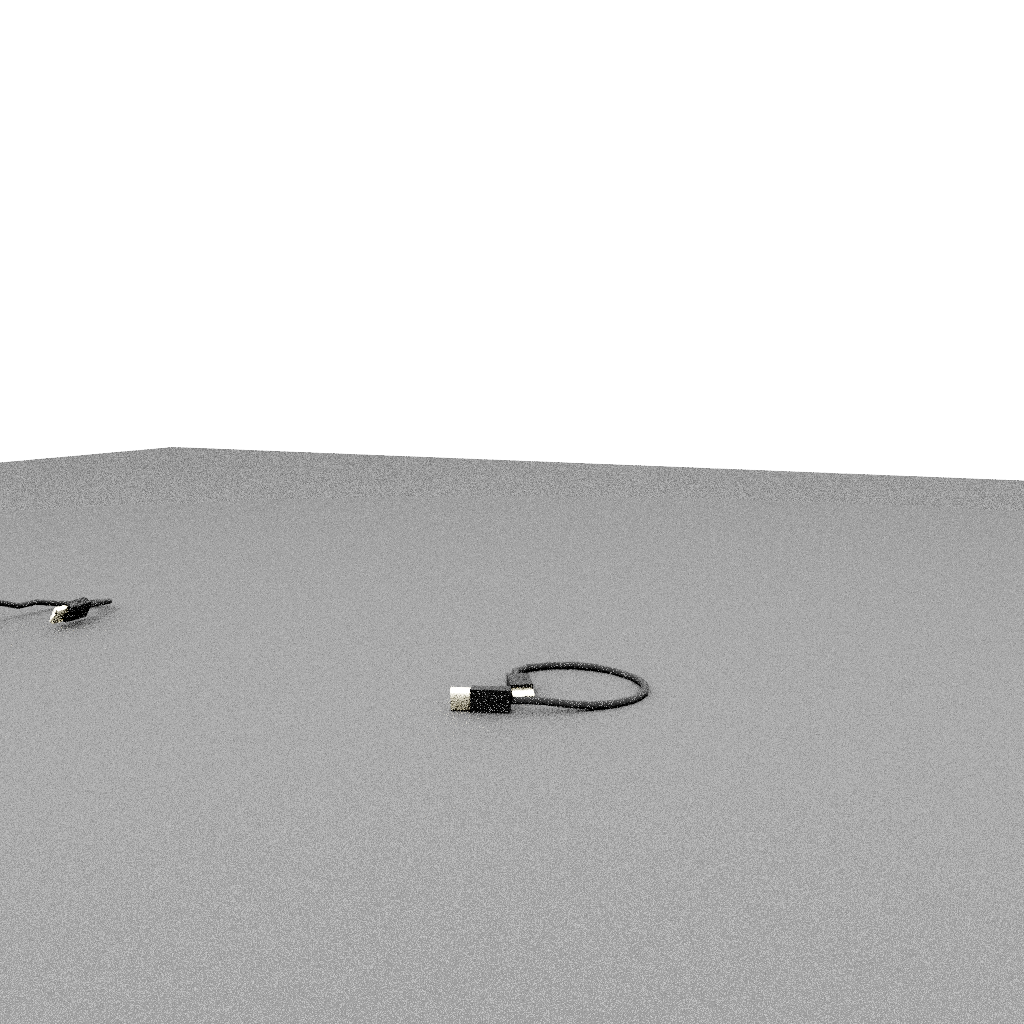} &
    \includegraphics[width=0.155\textwidth, trim=250 250 250 250, clip]{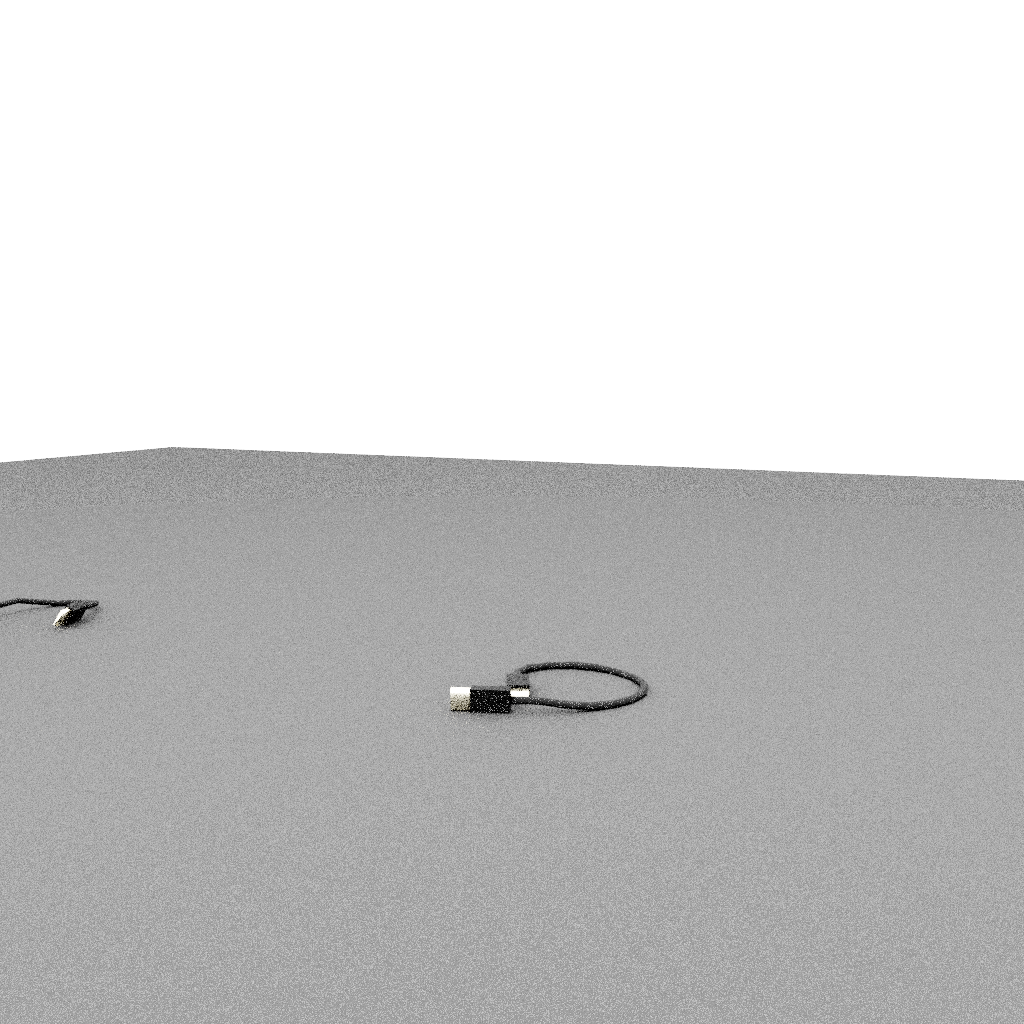} &
    \includegraphics[width=0.155\textwidth, trim=250 250 250 250, clip]{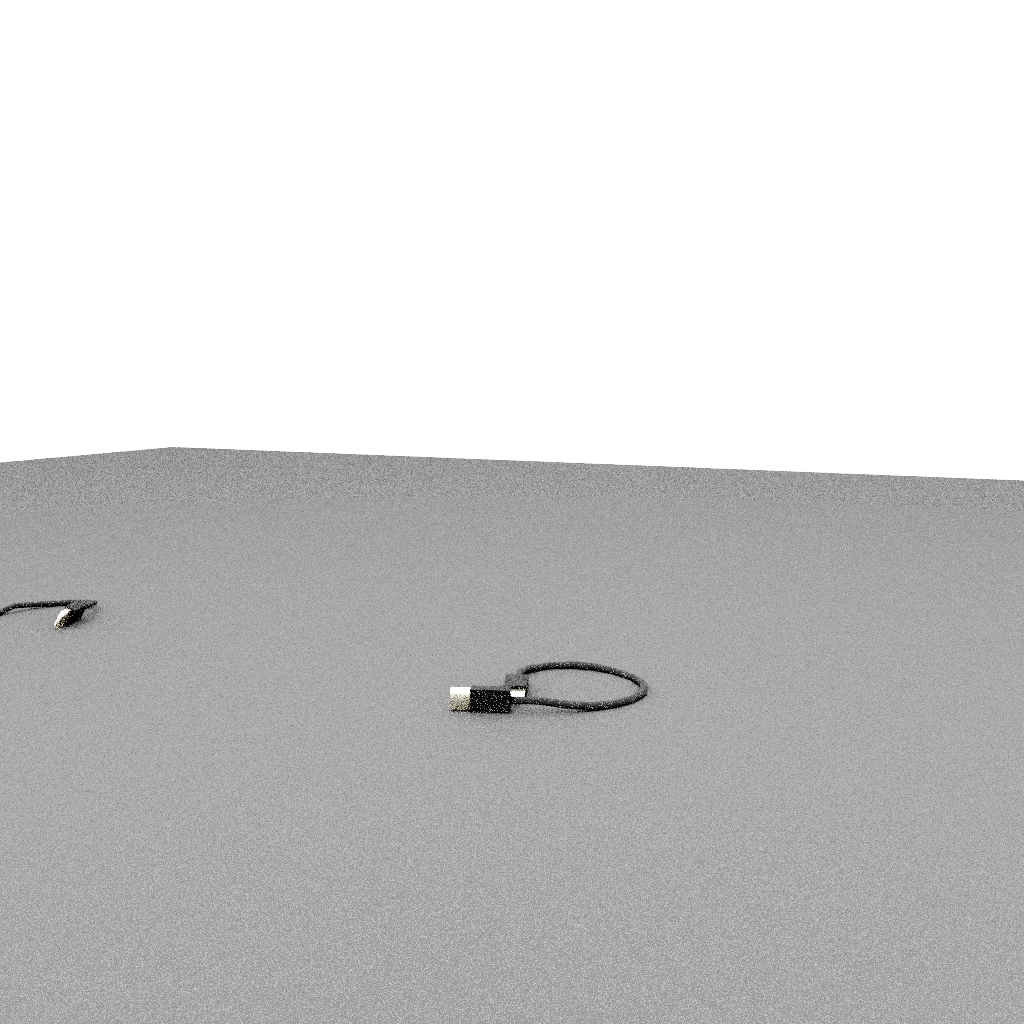}

    \end{tabular}

    \caption{
    \textbf{DLO--Connector Physics Demonstration.} Visualization of articulated
    and deformable wire concatenated with a DisplayPort connector across
    representative simulation frames.
    }
    \label{fig:appendix_wire_with_port}
\end{figure*}

Each connector is coupled to its DLO at the endpoints. For the deformable wire, each rigid connector is bound to the soft body through a PhysX auto-attachment (\texttt{PhysxPhysicsAttachment} with deformable-vertex attachments enabled and a vertex overlap offset of $0.015$~m), applied at both ends; forces transfer through the attached deformable vertices during grasping, routing, and insertion. For the articulated wire, the endpoint connector is rigidly welded to the adjacent segment with a fixed joint. Figure~\ref{fig:appendix_wire_with_port} visualizes the resulting DLO--Connector behavior for both representations across representative simulation frames.

\subsection{Computational Scaling and Throughput}
\label{appendix:wiredesign}

We benchmarked the deformable and articulated wire models in matched DLO-only scenes with ground contact enabled and endpoint connectors disabled. Each run used the $1/120$~s physics timestep without rendering and applied a $5$~s warmup followed by a $60$~s measured rollout per configuration. The sweep covered 64, 128, 256, 512, 1024, 2048, 4096, and 8192 parallel environments, stopping each model after its first failed point; the articulated wire was additionally probed at 16,384, where setup failed. The deformable curve uses the configuration with 32 solver position iterations. Because the two models expose different physical and numerical parameters, we compare their computational scaling under matched external task conditions rather than treating them as material-calibrated equivalents.

\begin{figure}[h]
\centering
\includegraphics[width=0.82\textwidth]{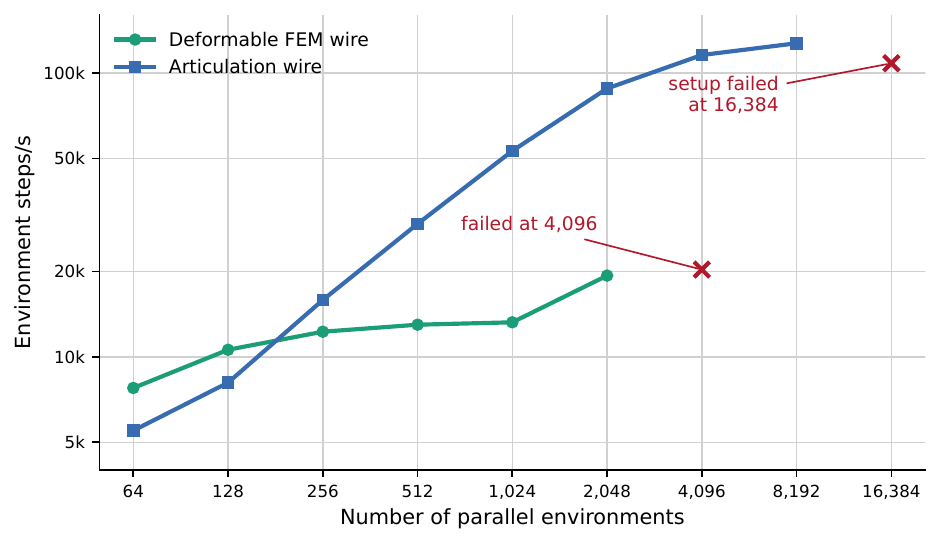}
\caption{\textbf{Scaling behavior of the two WireCraft wire models.} Both models were evaluated in matched headless DLO-only scenes with a $1/120$~s physics timestep, $5$~s warmup, $60$~s measured rollout. Absolute throughput depends on discretization and solver settings; the comparison is intended to show scaling behavior of the selected benchmark configurations. The articulation model scales to 8192 successful environments, while the deformable solver-32 model succeeds through 2048 environments and fails at 4096 owing to PhysX deformable-volume limits.}
\label{fig:wire_model_scaling}
\end{figure}

As shown in Figure~\ref{fig:wire_model_scaling}, which is measured on an NVIDIA RTX A4000 GPU, the articulated model scales successfully to $8192$ parallel environments, while the deformable model succeeds through $2048$ environments and fails at $4096$. The deformable wire is faster at the smallest environment counts, but the articulated wire reaches substantially higher aggregate throughput, whereas the deformable wire shows much weaker throughput scaling before reaching its limit.

The deformable limit is consistent with the per-scene cap of $4095$ deformable-volume actors enforced by the PhysX constant \texttt{PX\_MAX\_NB\_DEFORMABLE\_VOLUME}, which originates from the $12$-bit field used to pack the (deformable-volume id, tetrahedron id) pair into a single $32$-bit narrow-phase index. In practice, this hard ceiling is not the only constraint: GPU memory for deformable-volume contact buffers and the solver heap (\texttt{PxGpuDynamicsMemoryConfig}) is statically pre-allocated, does not grow dynamically, and may bind first depending on mesh resolution, contact density, and available device memory~\cite{physx_soft_bodies}. Because neither constraint is exposed as a runtime knob in Isaac Sim, we treat the deformable model as our higher-fidelity, lower-throughput representation and use the articulated model for high-throughput RL training and bulk trajectory generation; the deformable model is retained for contact-sensitive evaluation and physically detailed trajectory rollouts. Absolute throughput should be read in the context of each model's discretization and solver settings.

\section{Task Design}
\label{appendix:task design}

\subsection{Connector Insertion}
\label{appendix:connector insertion}

Connector insertion requires the robot to grasp the rigid connector at the end of a deformable wire and insert it into a matching socket. The task combines local peg-in-hole precision with global deformation effects: the trailing wire introduces drag, gravity-induced shape changes, and contact disturbances that perturb the connector during final alignment.

The task curriculum is set by connector and socket geometry, which is fixed per task rather than randomized across episodes. WireCraft registers five connector/socket geometries that span a range of rotational symmetries and alignment sensitivities. The cylinder connector requires the held connector's long axis to align with the socket axis but is rotationally symmetric about that axis, making it the easiest variant. The cuboid connector extends the cylinder requirement and additionally requires alignment of the rotation about the long axis. The Ethernet-style connector adds a latch-like geometry and serves as our primary insertion task; the DisplayPort connector has an asymmetric shell and is the most alignment-sensitive; and the USB-A connector provides a reference-standard geometry. Among these, the cylinder, cuboid, Ethernet-style, and DisplayPort connectors are used in the benchmark evaluations (Table 1); the USB-A connector is released as an additional asset but is not included in the reported experiments. A second difficulty axis is the socket clearance, which controls the gap between connector and socket; a smaller clearance yields a tighter, harder insertion.

\subsection{Clip Routing}
\label{appendix:clip routing}

Clip routing requires the robot to route a DLO through one or more structured U-slot clips mounted on the workspace. The task emphasizes fixture interaction and topological constraints: the DLO must be guided into the clip openings while maintaining contact stability and avoiding incomplete seating or slippage.

The task curriculum is set by the number of clips, fixed per task. Difficulty scales from single-clip routing to multi-clip sequences of up to three clips, where additional clips lengthen the multi-step route and require intermediate re-grasps. Each clip's $x$ position is fixed along the routing direction, while its $y$ position and yaw are randomized per episode. A secondary difficulty axis is the clip opening: the top-gap and lip retention can be adjusted, and a sufficiently wide opening removes the lip and reduces the clip to an unconstrained U-slot.

\subsection{Channel Seating}
\label{appendix:channel seating}

Channel seating requires the robot to lay a DLO into a geometric channel so that the DLO body conforms to the target path. Unlike connector insertion, where success depends on a local terminal pose, channel seating evaluates extended shape control over the entire DLO body.

The task curriculum is set by channel geometry, fixed per task. Each channel is constructed from five segments, and difficulty is controlled by the random relative rotation applied at each segment-to-segment joint: the straight channel uses zero inter-segment rotation and tests simple alignment along a linear path, while the curved channel applies a per-joint random bend of up to $\pm 20^\circ$, producing a non-linear path that requires continuous path-following.

\subsection{Randomization}
\label{appendix:task_randomization}

Each task applies per-episode randomization at reset, on top of the fixed per-task curriculum described above. Connector insertion randomizes the wire-attached connector and the socket; clip routing randomizes the clips; and channel seating randomizes the wire-attached connector and, optionally, the channel shape. The full set of ranges is listed in Table~\ref{tab:task_randomization}. The reported insertion ranges correspond to the RL training and evaluation configuration; the scripted and teleoperation data-collection pipelines use a different connector planar range, which we treat as dataset-generation settings rather than the benchmark task randomization.

\begin{table}[h]
\centering
\small
\caption{
Per-episode randomization for the WireCraft task families, grouped by the
randomized component. The connector is attached to the DLO terminal and
randomized before a short free-fall settling drop. Ranges are reported for the
RL training and evaluation configuration.
}
\label{tab:task_randomization}
\setlength{\tabcolsep}{5pt}
\renewcommand{\arraystretch}{1.15}
\begin{tabular}{llp{0.42\linewidth}}
\toprule
Component & Parameter & Default range / description \\
\midrule
\multirow{5}{*}{Connector}
& Spawn height   & $6$--$11$~cm above the table, then free-fall settling \\
& Roll / pitch   & Uniform $[-18^\circ, 18^\circ]$ \\
& Yaw            & Uniform $[-36^\circ, 36^\circ]$ \\
& $X$ offset     & Uniform $\pm 5$~cm \\
& $Y$ offset     & Uniform $\pm 15$~cm \\
\midrule
\multirow{3}{*}{Socket}
& $X$ position   & Fixed along the insertion axis \\
& $Y$ position   & Randomized within a $28$~cm range \\
& $Z$ position   & Randomized within a $15$~cm range \\
\midrule
\multirow{3}{*}{Clips}
& $X$ position   & Fixed along the routing line \\
& $Y$ position   & Uniform within $\pm 6$~cm of the nominal lane \\
& Yaw            & Uniform $[-15^\circ, 15^\circ]$ \\
\midrule
\multirow{2}{*}{Channels}
& Position       & Fixed \\
& Shape & $5$ segments; each adds a signed bend of up to $\pm 20^\circ$ to the running heading \\
\bottomrule
\end{tabular}
\end{table}

\subsection{Control Modes}
\label{appendix:control_modes}

WireCraft exposes three end-effector control modes, selectable per task, so that users can match the action space to the task's requirements. \emph{Joint control} commands joint-position targets directly. \emph{Cartesian control} commands a $6$-DoF end-effector delta pose, resolved to joint targets by a differential-IK controller. \emph{Cartesian control with fixed orientation} commands only the $3$-DoF translation while holding the end-effector orientation fixed, which suits tasks where a constant tool pose is desired. Connector insertion and clip routing use full Cartesian control, since both require reorienting the held object; channel seating uses the fixed-orientation mode to keep the connector level while laying the DLO into the channel. All three modes share the same gripper command and downstream joint-target interface, so a task can switch modes without changing the policy's action head.

\section{Scripted Policy}
\label{app:scripted_policy}

All three task families share a common scripted-policy framework: a per-environment vectorized finite-state machine that, within each phase, reads the current end-effector and fixture poses, expresses a target end-effector pose in the robot base frame, and emits a Cartesian delta-pose action obtained by clamping the pose error toward that target. Each policy selects the control mode best suited to its task (see Appendix~\ref{appendix:control_modes}). A phase advances once the end-effector reaches its target within a position and orientation tolerance and dwells for a minimum number of control steps. At the grasp phase the policy snapshots the relevant poses, so that every post-grasp phase servos toward a stable reference that is unaffected by the deformation introduced once the wire is held. The three policies differ in what they grasp, in their task-specific sub-skills, and in the perturbations they inject to diversify demonstrations, as summarized in Table~\ref{tab:scripted_policy_comparison}.

\begin{table}[h]
\centering
\small
\caption{
Comparison of the three scripted policies. All share the vectorized
finite-state-machine framework and a Cartesian delta-pose action; they differ
in their grasp target, control dimensionality, core sub-skill, and the
perturbations injected for demonstration diversity.
}
\label{tab:scripted_policy_comparison}
\setlength{\tabcolsep}{5pt}
\renewcommand{\arraystretch}{1.2}
\begin{tabularx}{\linewidth}{p{0.18\linewidth} X X X}
\toprule
& \textbf{Insertion} & \textbf{Clip routing} & \textbf{Channel seating} \\
\midrule
Grasp target
& Connector tail
& DLO body node near the clip
& Connector \\

Action
& 7-DoF (pose + gripper)
& 7-DoF (pose + gripper)
& 7-DoF with fixed end-effector orientation  \\

Core sub-skill
& Align and drive into the socket
& Over-stretch, then push the DLO into the slot
& Lift and slowly place into the channel \\

Reference frame
& Socket pose
& Clip pose with a frozen anchor-to-clip direction
& Channel center and length \\

Success
& Connector seated in the socket
& DLO node inside the clip slot
& DLO body laid within the channel over 80\% \\

Injected perturbation
& Pick-and-drop ($30\%$) and failure retry
& Pre-stretch with a frozen target direction
& Delayed grasp and slowed placement \\
\bottomrule
\end{tabularx}
\end{table}

\subsection{Insertion}
\label{app:scripted_insertion}

The insertion policy grasps the connector at its tail and seats it into the socket. After grasping and lifting, it moves to a pre-insertion pose, solves for the gripper orientation that aligns the held connector with the socket under a tightened orientation tolerance, drives the connector into the socket, and releases. To produce demonstrations of failure and recovery, with probability $0.3$ the policy injects a pick-and-drop perturbation, lifting and releasing the connector mid-air before re-approaching and regrasping, and a failure-triggered retry returns to the approach phase if the insertion has not succeeded after the hold phase, for up to two attempts. Per-episode randomization of the approach pitch and the tail grasp offset, together with phase-local slowdowns around the pick and insertion, further diversify the demonstrations.

\subsection{Clip Routing}
\label{app:scripted_clip_routing}

The clip-routing policy grasps the deformable wire itself at the node nearest the clip, rather than the connector, since the connector is wider than the clip mouth and cannot enter the slot. The grasp node is located by proximity and the grasp orientation is set to the local wire tangent. After descending to a fixed table-relative height and closing on the wire, the policy lifts and rotates the held wire to the clip's slot axis. It then over-stretches the wire by translating past the placement target along the anchor-to-clip direction, pulls back to the clip front to release excess tension, and pushes the held wire down past the clip lip so that it seats into the slot under a controlled descent. Success requires a wire node to lie inside the clip slot. The anchor-to-clip direction is frozen at grasp time because the wire's far endpoint is a dynamic body that drifts under tension, which would otherwise cause the placement target to move every step.

\subsection{Channel Seating}
\label{app:scripted_channel_seating}
The channel-seating policy grasps the connector and lays the wire into the channel so that its body follows the target path. Unlike the other two policies, which use full Cartesian control, it uses the fixed-orientation Cartesian mode with a much finer action scale to keep the connector level and produce smooth, non-snagging motion. It grasps the connector gently, holding the gripper open for a settling delay before closing, then lifts vertically and settles. It then descends along the channel path at a strongly reduced speed, seating the wire at the channel-center height before releasing. Following the path rather than descending at a single waypoint lets the same policy handle both straight and curved channels.

\subsection{Scripted Demonstrations}

\begin{figure*}[htbp]
    \centering
    \setlength{\tabcolsep}{2pt}
    \renewcommand{\arraystretch}{0.85}
    \begin{tabular}{@{}c@{\hspace{2pt}}c@{\hspace{2pt}}c@{\hspace{2pt}}c@{}}
    \begin{overpic}[width=0.24\textwidth]{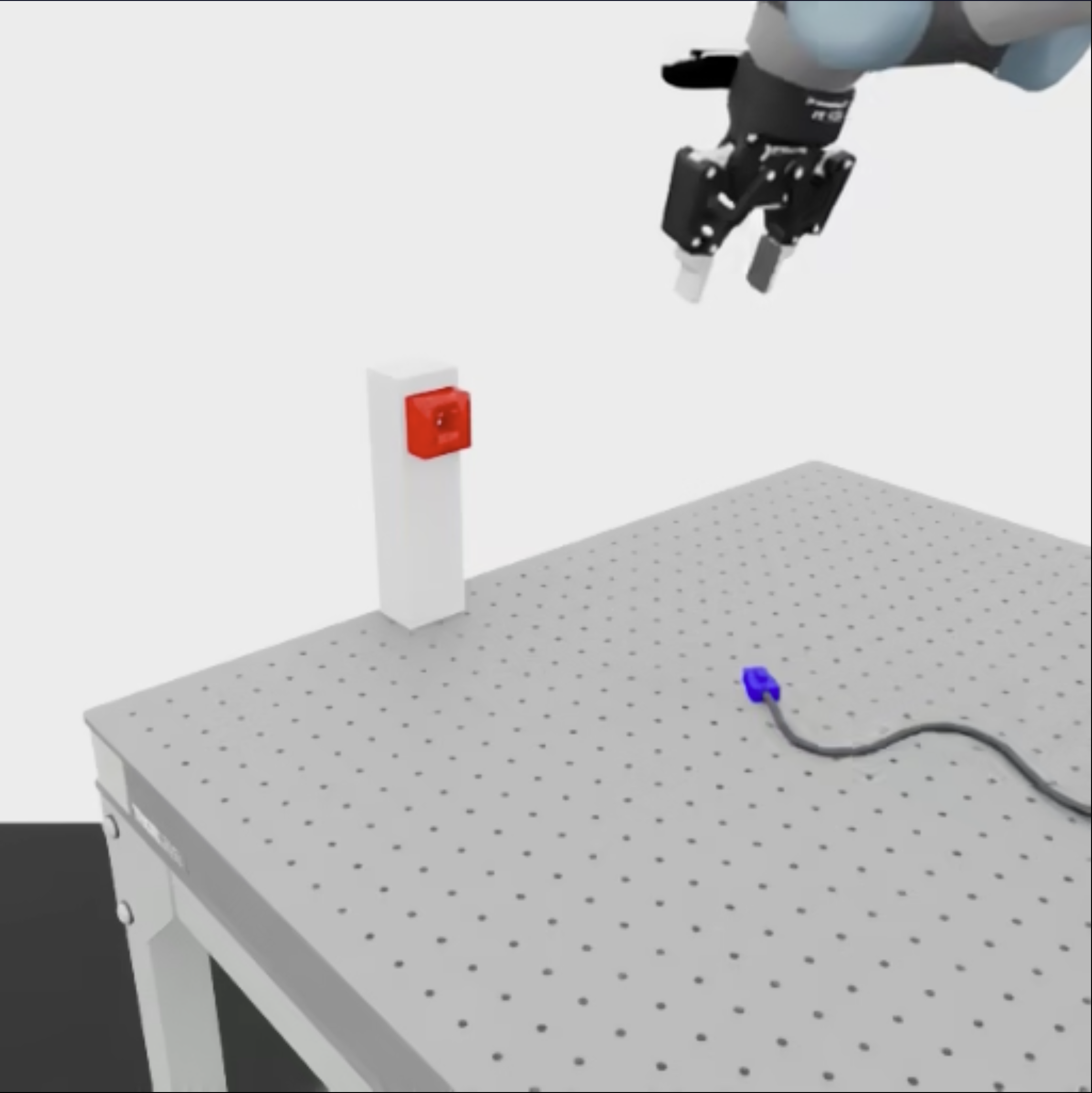}
        \put(3,3){\large\textbf{\textcolor{white}{Insertion}}}
    \end{overpic} &
    \includegraphics[width=0.24\textwidth]{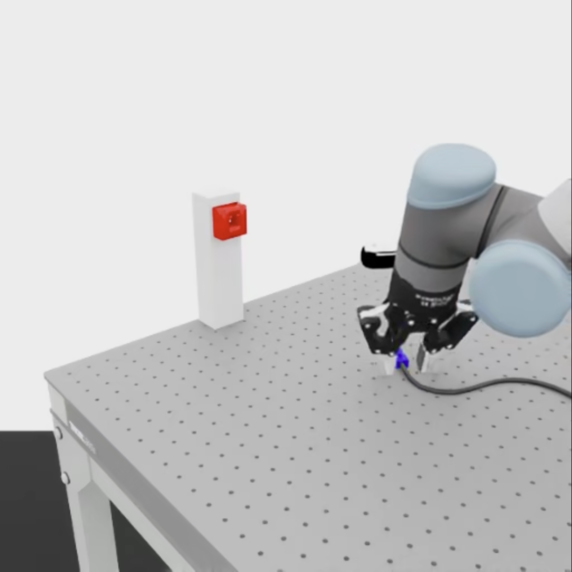} &
    \includegraphics[width=0.24\textwidth]{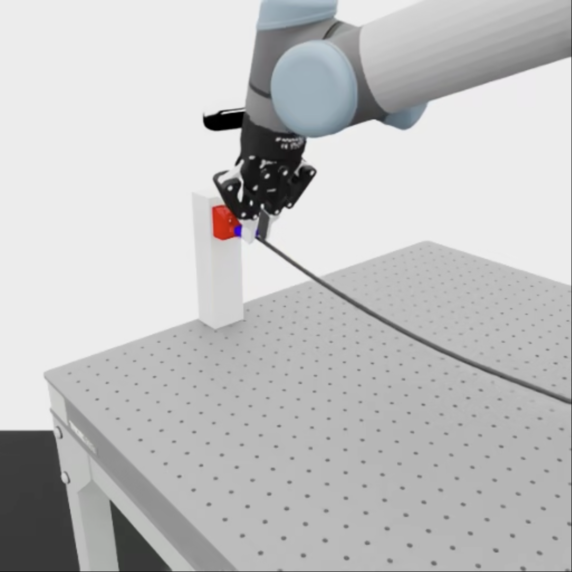} &
    \includegraphics[width=0.24\textwidth]{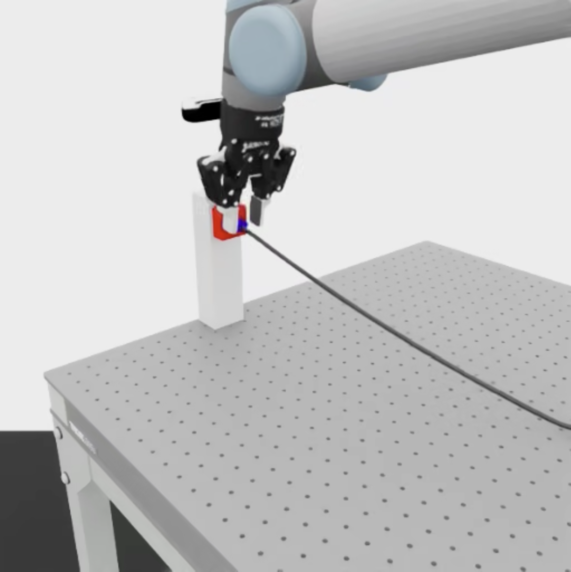}
    \\[2pt]
    \begin{overpic}[width=0.24\textwidth]{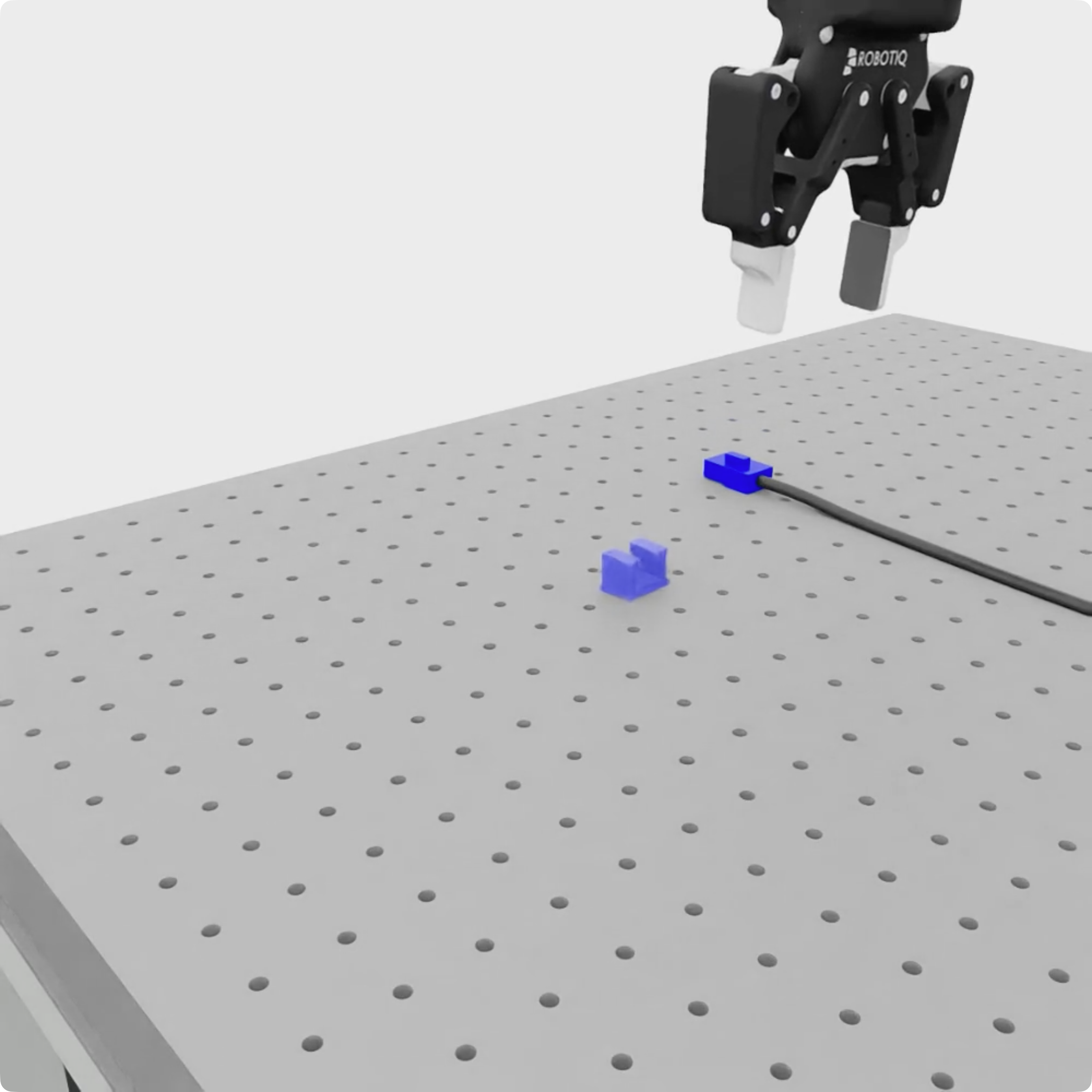}
        \put(3,3){\large\textbf{\textcolor{white}{Clip routing}}}
    \end{overpic} &
    \includegraphics[width=0.24\textwidth]{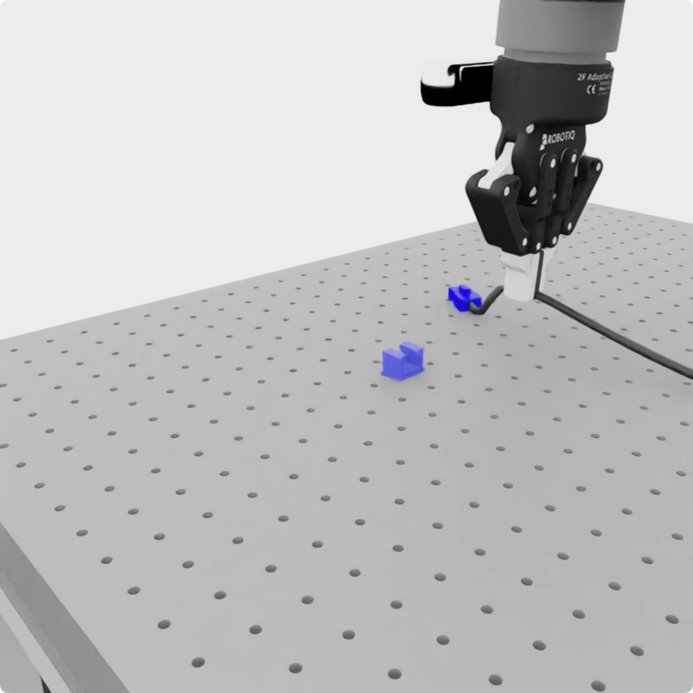} &
    \includegraphics[width=0.24\textwidth]{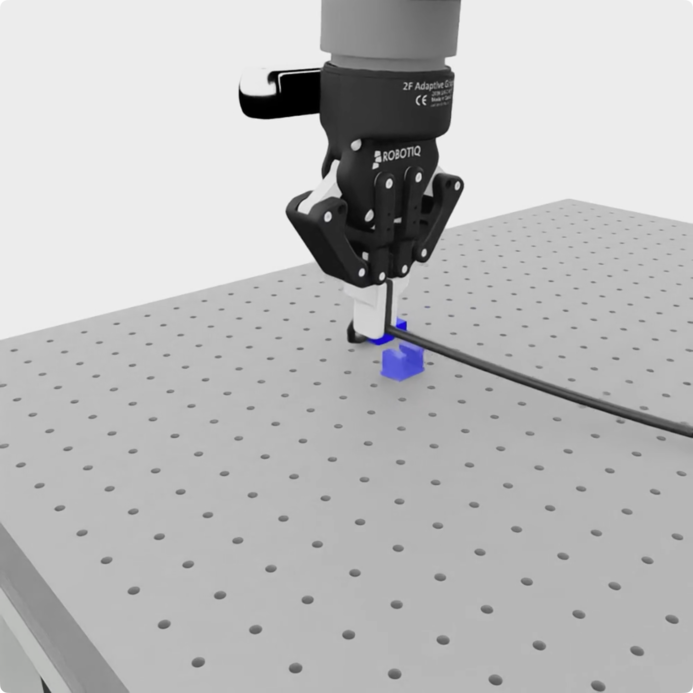} &
    \includegraphics[width=0.24\textwidth]{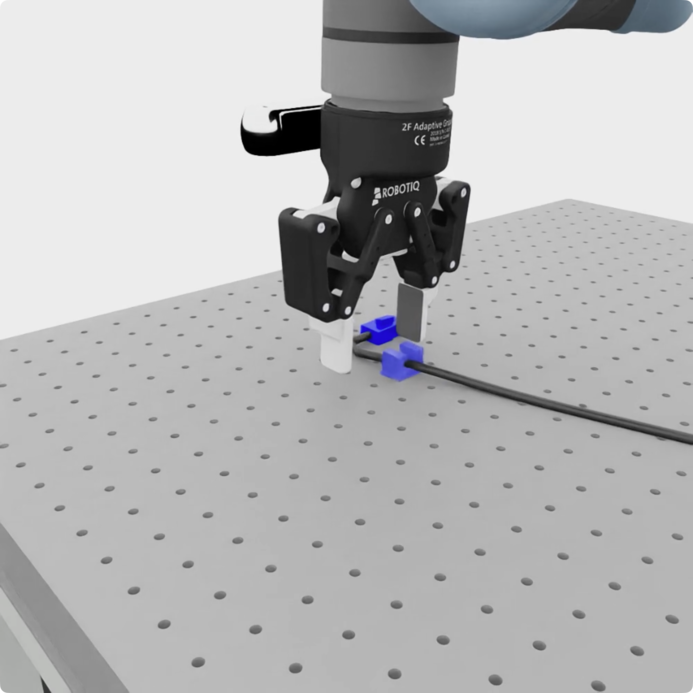}
    \\[2pt]
    \begin{overpic}[width=0.24\textwidth]{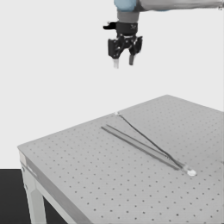}
        \put(3,3){\large\textbf{\textcolor{white}{Channel seating}}}
    \end{overpic} &
    \includegraphics[width=0.24\textwidth]{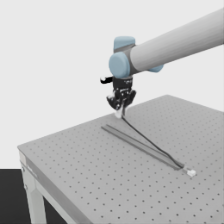} &
    \includegraphics[width=0.24\textwidth]{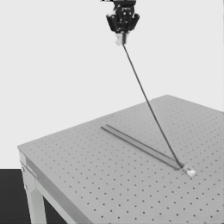} &
    \includegraphics[width=0.24\textwidth]{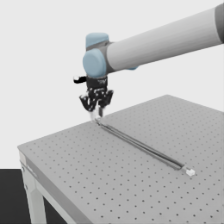}
    \end{tabular}
    \caption{\textbf{Scripted demonstration rollouts.} Each row shows a scripted policy executing one task family across representative phases, progressing from the initial state through grasp and manipulation to task completion.}
    \label{fig:scr_rollout_sequence}
\end{figure*}

See Figure~\ref{fig:scr_rollout_sequence}.

\section{RL Experiment Detail}
\label{appendix:parameters}

\subsection{State-based RL}
\label{app:state_rl}

We detail the privileged RL configuration on connector insertion, our primary RL setting. On insertion, the two privileged baselines, PPO and SACfD, share the same observation, action, and reward, and differ only in the learning algorithm. Clip routing and channel seating reuse the same PPO algorithm, network, and hyperparameters without per-task tuning, under their own task-specific observations and rewards, and we do not detail them further here. All RL results in Section~\ref{sec:baseline_experiments} are reported as mean $\pm$ standard deviation across $3$ seeds.

\paragraph{Observation.}
The insertion observation is a $42$-dimensional privileged state vector (\texttt{observation\_space=42}) : arm joint positions ($6$) and gripper position ($1$); arm joint velocities ($6$) and gripper velocity ($1$); end-effector position ($3$) and quaternion ($4$); left/right fingertip contact ($2$); the connector-to-seated-target vector ($3$); connector position ($3$) and quaternion ($4$); socket position ($3$); seated-target position ($3$); and the insertion-gate and axial/lateral seating-error scalars ($3$).

\paragraph{Action and control.}
The insertion action is a $6$-DoF Cartesian end-effector delta resolved to joint targets by a differential-IK controller, plus one gripper command. Actions are clipped to $[-1,1]$ and scaled by $0.01$ in RL setting. Simulation runs at $\Delta t = 1/120$~s with decimation $2$, giving a $60$~Hz policy rate; episodes last $8$~s.

\paragraph{Reward.}
The insertion reward is a shaped sum of reach ($15$), grasp ($25$), bring ($40$), two-point YZ alignment ($20$), a roll-gated insertion-depth high-water ratchet ($6000$), socket/roll/EE-orientation terms ($10$ each), and a sparse seated bonus ($1000$), against action and table/gripper penalties. PPO and SACfD use this identical reward; PPO applies empirical observation normalization, while SACfD normalizes the state with a running standard scaler. Insertion results across connectors are in Table~\ref{tab:rl_port_baselines}.

\paragraph{Networks.}
PPO~\cite{schulman2017ppo} uses an actor--critic MLP with hidden sizes $[256,128,64]$ and ELU activations for both actor and critic. SACfD~\cite{vecerik2017leveraging} uses a tanh-squashed Gaussian actor ($[256,256]$, ELU) with twin $Q$-critics ($[256,256]$, ELU). PPO hyperparameters are listed in Table~\ref{tab:state_ppo_hp} and SACfD in Table~\ref{tab:sac_hp}.

\begin{table}[h]
\centering\small
\caption{State-based PPO hyperparameters. Vision PPO
(Sec.~\ref{app:vision_ppo}) inherits these except where noted.
\texttt{num\_envs} is set at launch with no committed default.}
\label{tab:state_ppo_hp}
\begin{tabular}{ll}
\toprule
Parameter & Value \\
\midrule
Actor / critic MLP           & $[256,128,64]$, ELU \\
Clip range $\epsilon$        & $0.2$ \\
Entropy coef.                & $0.01$ \\
Value-loss coef.             & $1.0$ \\
GAE $\lambda$                & $0.95$ \\
Discount $\gamma$            & $0.99$ \\
Learning rate                & $5\times10^{-4}$, adaptive ($\mathrm{KL}^{*}{=}0.01$) \\
Steps per env (rollout)      & $128$ \\
Minibatches / epochs         & $8$ / $5$ \\
Max grad norm                & $1.0$ \\
Max iterations               & $20000$ \\
Obs.\ normalization          & empirical (on) \\
\bottomrule
\end{tabular}
\end{table}

\begin{table}[h]
\centering\small
\caption{SACfD hyperparameters on connector insertion tasks.}
\label{tab:sac_hp}
\begin{tabular}{ll}
\toprule
Parameter & Value \\
\midrule
Actor / critic MLP           & $[256,256]$, ELU \\
Discount $\gamma$             & $0.99$ \\
Target smoothing $\tau$       & $0.005$ \\
Actor / critic LR             & $3\times10^{-4}$ / $3\times10^{-4}$ \\
Batch size                    & $256$ \\
Gradient steps / env step     & $1$ \\
Replay capacity               & $8000\times N_\text{env}$ \\
Warmup (learning starts)      & $5000$ \\
Entropy temp.\ $\alpha$       & learned, target $-7$, $\alpha_0{=}0.1$ \\
Reward shaping scale          & $0.01$ \\
State preproc.                & RunningStandardScaler \\
Total timesteps               & $10^6$ \\
\midrule
\multicolumn{2}{l}{\emph{Demonstration augmentation (SACfD)}}\\
Demo set                      & $\sim$150k transitions (scripted) \\
Demo replay ratio             & $0.5$ \\
BC weight (anneal)            & $2.5\!\to\!0.5$ over $[80\text{k},250\text{k}]$ steps \\
\bottomrule
\end{tabular}
\end{table}

\subsection{Vision PPO}
\label{app:vision_ppo}

\paragraph{Design rationale.}
Training a visual policy from scratch with online RL is sample-inefficient: the
encoder must learn useful features purely from the RL reward, which is slow and
wasteful in interaction. We therefore decouple representation learning from
control. The visual encoder is first supervised by behavior cloning on the
corresponding state policy's rollouts, and only then refined with online PPO,
which substantially speeds convergence. For the encoder we use a pretrained
DINOv2 ViT backbone~\cite{oquab2024dinov2, jaegleperceiver} with a learned attention-pooling head rather than a CNN
trained from scratch, since a frozen self-supervised backbone provides
transferable features and keeps the online phase cheap; the attention pool lets
the policy aggregate task-relevant regions across the camera views.

\paragraph{Architecture.}
Vision PPO replaces the privileged actor input with two $224\times224$ RGB streams, a wrist camera and a fixed third-person side camera, plus a $14$-d proprioceptive vector of arm joint positions and velocities together with the gripper position and velocity. It is an \emph{asymmetric} actor--critic~\cite{Pinto2017AsymmetricAC}: the actor sees images and proprioception, while the critic retains the full $42$-d privileged state. The visual actor is a DINOv2 ViT-S/14 backbone with a cross-attention pooling head ($8$ learned queries per camera, $16$ total; query dimension $64$, hidden $512$) that produces the $7$-d action mean. The encoder is behavior-cloned from the state policy's rollouts before PPO; during PPO the backbone and attention pool are frozen, and only the final projection head, the action $\log\sigma$, and the critic are trained. The critic is warm-started from the converged State-PPO critic and remains trainable.

\paragraph{Hyperparameters.}
Vision PPO inherits all PPO settings in Table~\ref{tab:state_ppo_hp} except for a tighter trust region (clip $0.1$, target-KL $0.005$), no entropy bonus, a down-weighted value loss ($0.3$), a lower learning rate ($1\times10^{-4}$), a small initial action noise ($\sigma=0.05$), $64$ parallel environments, and $600$ fine-tuning iterations from the behavior-cloned initialization, with observation normalization disabled. Training was run on an NVIDIA RTX PRO 4500 Blackwell GPU. Per-connector results are in Table~\ref{tab:rl_port_baselines}. 

\subsection{RL Rollout Demonstrations}

\begin{figure*}[htbp]
    \centering
    \setlength{\tabcolsep}{2pt}
    \renewcommand{\arraystretch}{0.85}
    \begin{tabular}{@{}c@{\hspace{2pt}}c@{\hspace{2pt}}c@{\hspace{2pt}}c@{}}
    \begin{overpic}[width=0.24\textwidth]{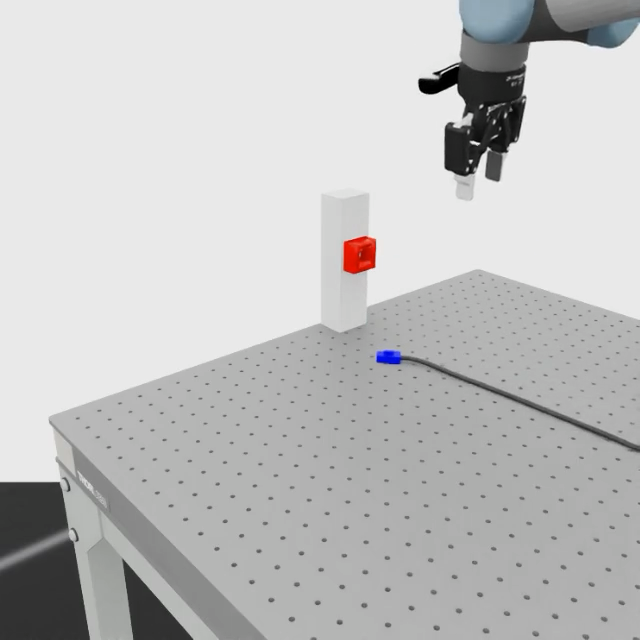}
        \put(3,3){\large\textbf{\textcolor{white}{Insertion}}}
    \end{overpic} &
    \includegraphics[width=0.24\textwidth]{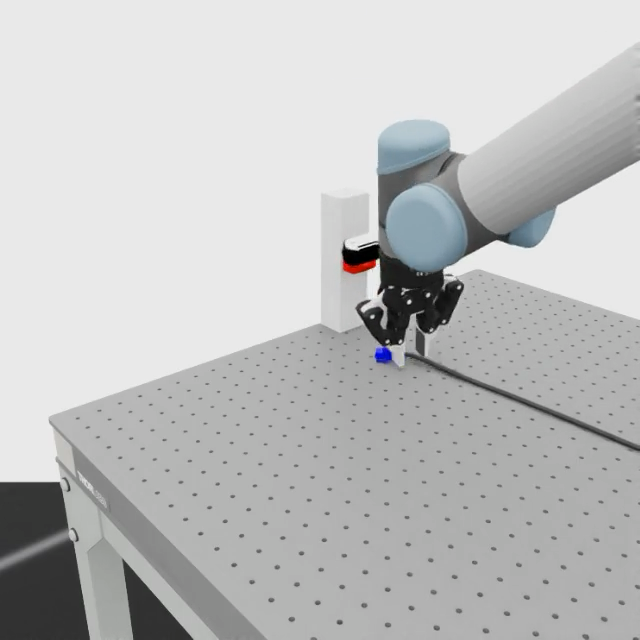} &
    \includegraphics[width=0.24\textwidth]{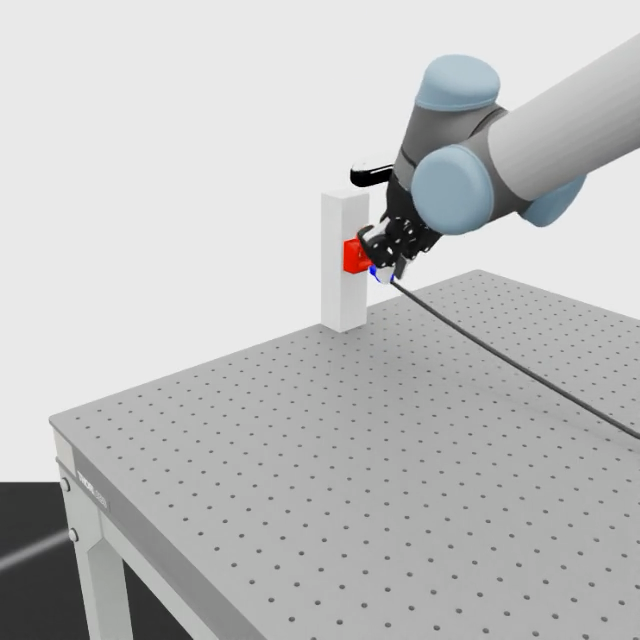} &
    \includegraphics[width=0.24\textwidth]{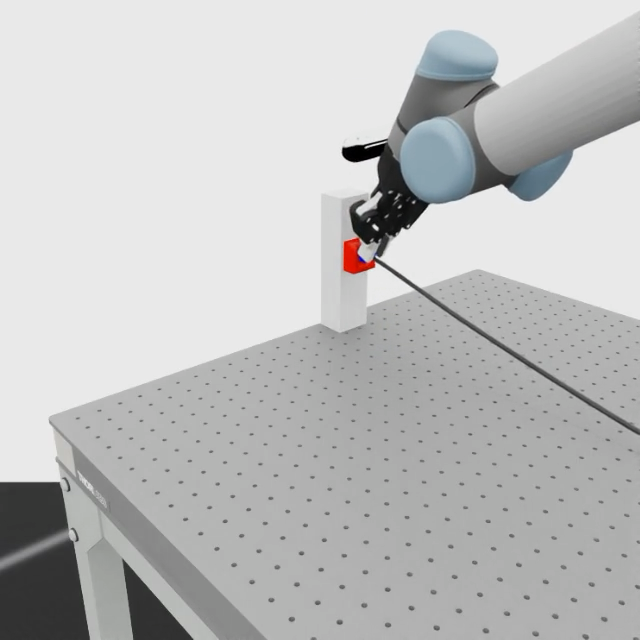}
    \\[2pt]
    \begin{overpic}[width=0.24\textwidth]{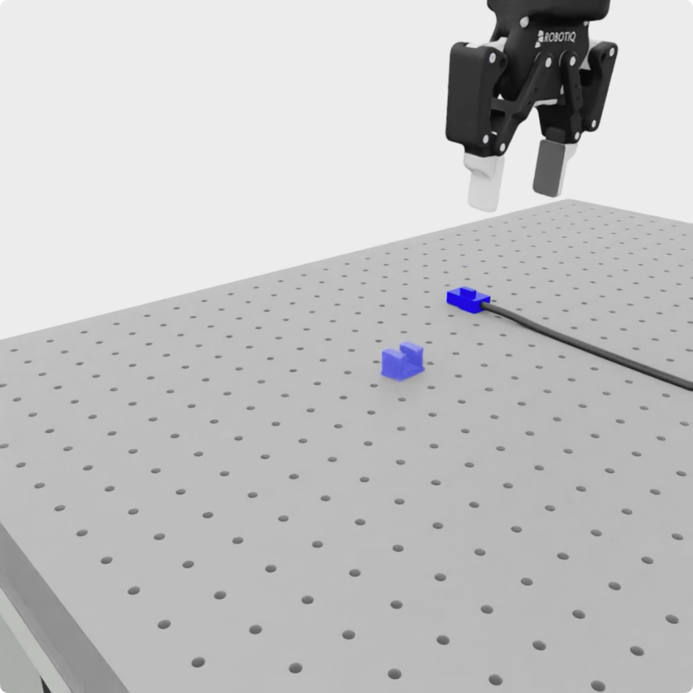}
        \put(3,3){\large\textbf{\textcolor{white}{Clip routing}}}
    \end{overpic} &
    \includegraphics[width=0.24\textwidth]{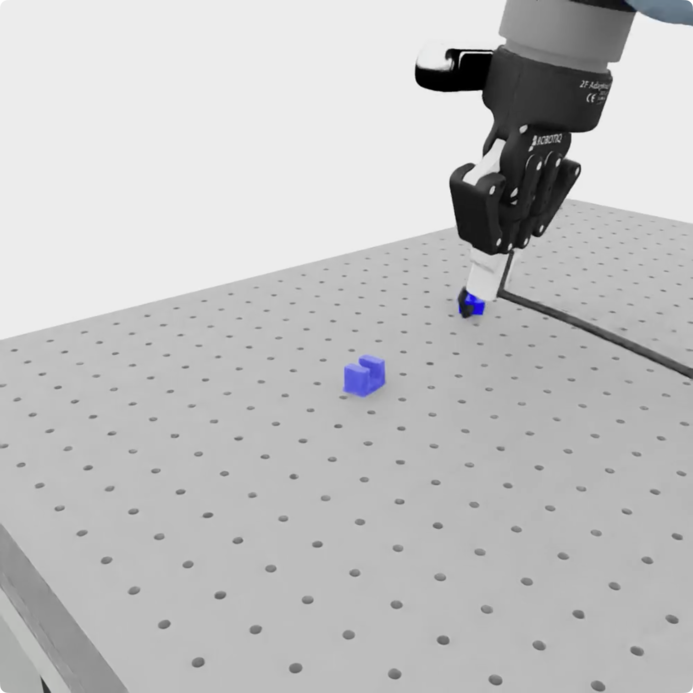} &
    \includegraphics[width=0.24\textwidth]{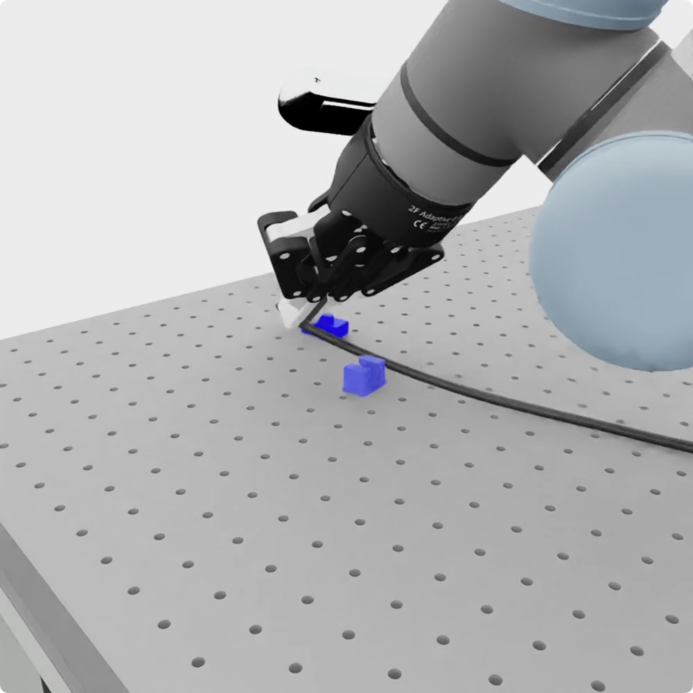} &
    \includegraphics[width=0.24\textwidth]{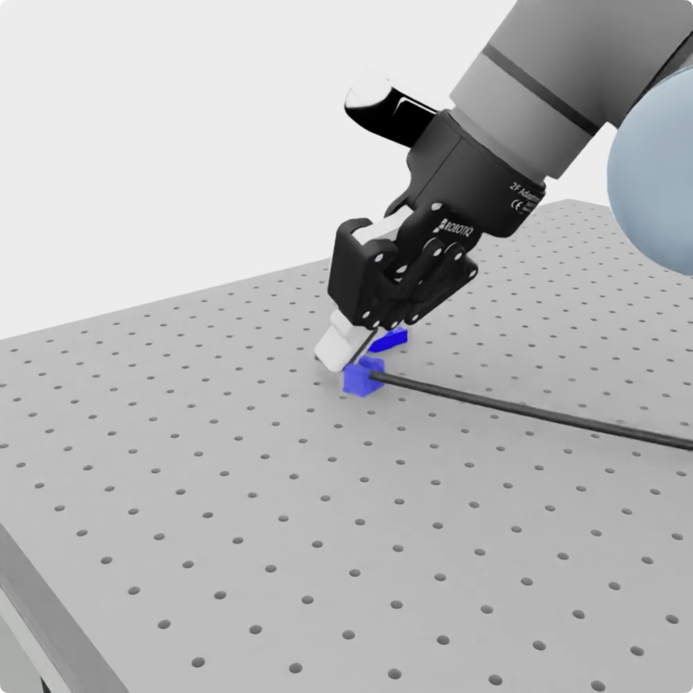}
    \\[2pt]
    \begin{overpic}[width=0.24\textwidth]{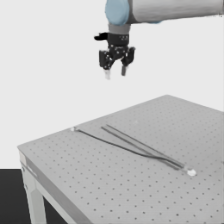}
        \put(3,3){\large\textbf{\textcolor{white}{Channel seating}}}
    \end{overpic} &
    \includegraphics[width=0.24\textwidth]{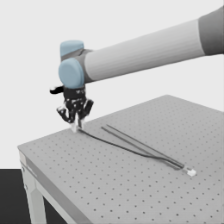} &
    \includegraphics[width=0.24\textwidth]{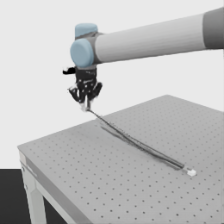} &
    \includegraphics[width=0.24\textwidth]{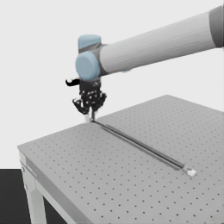}
    \\
    \end{tabular}
    \caption{\textbf{RL rollout demonstrations.} Each row shows a trained State
    PPO policy rolled out on one task family across representative phases. The
    clip-routing row additionally illustrates an error and subsequent recovery.}
    \label{fig:full_sequence_demonstration}
\end{figure*}

See Figure~\ref{fig:full_sequence_demonstration}.

\section{Connector-Insertion IL/VLA Experimental Details}
\label{app:ilvla_details}

\subsection{Dataset}
All four IL/VLA policies are trained on a shared dataset of 1,000 scripted connector-insertion trajectories collected in the same Isaac Lab task used for evaluation. Each trajectory contains two RGB image streams, one from a third-person workspace camera and one from a wrist-mounted camera, together with UR5 joint positions, gripper state, and joint-space actions. The action representation contains six arm commands and one gripper command. The shared dataset and action/state encoding allow all policies to be evaluated under the same observation and control interface.

\subsection{Policy Training Configurations}
See Table~\ref{tab:policy_training_configs}.

\begin{table}[h]
\centering
\scriptsize
\caption{Training configurations for the IL/VLA connector-insertion baselines. All policies use the same dataset and action/state encoding.}
\label{tab:policy_training_configs}
\begin{tabular}{p{0.16\linewidth}p{0.30\linewidth}p{0.20\linewidth}p{0.24\linewidth}}
\toprule
Policy & Model / conditioning & Action prediction & Training \\
\midrule
ACT & ResNet-18 visual encoder with transformer action-chunking policy & Chunk size 100; 50 executed actions per inference call & 100k gradient steps; batch size 64; learning rate $10^{-5}$ \\
Diffusion Policy & ResNet-18 visual encoders with a temporal U-Net action denoiser & Horizon 16; 2 observation steps; 8 action steps & 200k gradient steps; batch size 64; learning rate $10^{-4}$ \\
Diffusion Transformer VLA & Frozen CLIP ViT-B/16 image and text conditioning with a DiT-style action denoiser & Horizon 32; 24 action steps & 300k gradient steps; batch size 32; learning rate $2 \times 10^{-5}$ \\
\pizero{} & Initialized from \texttt{lerobot/pi05\_base}; vision tower frozen; action expert and language adapter trainable & 25 action steps & 50k gradient steps; batch size 32; learning rate $10^{-5}$ \\
\bottomrule
\end{tabular}
\end{table}

\subsection{Checkpoint Selection}
Each policy was checkpointed periodically during training and evaluated using the original reach-success criterion. We retained the top three checkpoints for each policy according to this reach success, rather than selecting checkpoints using insert success after the final evaluation. This avoids choosing checkpoints based on the stricter test metric while still allowing us to report how the best training-time checkpoints behave under both reach and insert success criteria. Because checkpoints are selected by reach success on the evaluation rollouts themselves rather than on a held-out validation set, the reported figures are an optimistic, best-observed estimate under our training budget and should not be read as validation-based model selection; insert success is taken from the same checkpoint and is therefore not separately optimized.

\subsection{Evaluation Protocol}
For each retained checkpoint (See Table~\ref{tab:checkpoint_selection}), we evaluate four random seeds with 16 rollouts per seed, for a total of 64 rollouts per checkpoint (see Table~\ref{tab:all_selected_checkpoint_results}). Each rollout records both reach and insert success from the same trajectory, avoiding separate evaluation runs for the two criteria. Episodes use a fixed horizon of 800 simulator steps. We report mean and standard deviation across seeds.

\begin{table}[h]
\centering
\small
\caption{Available and retained checkpoints for the connector-insertion IL/VLA baselines. Retained checkpoints are the top three checkpoints per policy according to online reach success during training.}
\label{tab:checkpoint_selection}
\begin{tabular}{lcc}
\toprule
Policy & Number of available checkpoints & Retained checkpoints \\
\midrule
ACT & 20 & 25k, 40k, 100k \\
Diffusion Policy & 20 & 140k, 190k, 200k \\
Diffusion Transformer VLA & 23 & 110k, 180k, 200k \\
\pizero{} & 5 & 10k, 20k, 30k \\
\bottomrule
\end{tabular}
\end{table}

\begin{table}[h]
\centering
\scriptsize
\caption{Connector-insertion results for the retained checkpoints selected by reach success. Each entry reports mean $\pm$ standard deviation over four seeds, with 16 rollouts per seed.}
\label{tab:all_selected_checkpoint_results}
\begin{tabular}{lccc}
\toprule
Policy & Checkpoint & \srreach{} (\%) & \srinsert{} (\%) \\
\midrule
ACT & 25k & $43.75 \pm 11.69$ & $3.13 \pm 3.13$ \\
ACT & 40k & $54.69 \pm 17.88$ & $9.38 \pm 6.99$ \\
ACT & 100k & $\mathbf{60.94 \pm 11.16}$ & $\mathbf{29.69 \pm 9.24}$ \\
\midrule
Diffusion Policy & 140k & $29.69 \pm 18.42$ & $17.19 \pm 13.53$ \\
Diffusion Policy & 190k & $35.94 \pm 9.24$ & $17.19 \pm 6.81$ \\
Diffusion Policy & 200k & $\mathbf{37.50 \pm 9.88}$ & $\mathbf{20.31 \pm 6.81}$ \\
\midrule
Diffusion Transformer VLA & 110k & $9.38 \pm 6.99$ & $3.13 \pm 3.13$ \\
Diffusion Transformer VLA & 180k & $4.69 \pm 5.18$ & $1.56 \pm 2.71$ \\
Diffusion Transformer VLA & 200k & $\mathbf{10.94 \pm 2.71}$ & $\mathbf{3.13 \pm 3.13}$ \\
\midrule
\pizero{} & 10k & $0.00 \pm 0.00$ & $0.00 \pm 0.00$ \\
\pizero{} & 20k & $\mathbf{6.25 \pm 4.42}$ & $\mathbf{0.00 \pm 0.00}$ \\
\pizero{} & 30k & $1.56 \pm 2.71$ & $0.00 \pm 0.00$ \\
\bottomrule
\end{tabular}
\end{table}

\section{Clip Routing and Channel Seating Experimental Details}
\label{app:routing_seating_details}

\paragraph{Datasets.}
ACT and DiT VLA use the same scripted demonstration dataset for each task: 1,600 trajectories for 1-clip routing and 1,000 trajectories for straight-channel seating.

\paragraph{Training.}
For both tasks, ACT is trained for 100k gradient steps with batch size 64 and learning rate $10^{-5}$; DiT VLA is trained for 300k gradient steps with batch size 32 and learning rate $2\times10^{-5}$. All four training runs use seed 1000. ACT uses a ResNet-18 visual encoder, predicts 100-action chunks, and executes 50 actions per inference call. DiT VLA uses CLIP ViT-B/16 image and text conditioning, two observation steps, a 32-step prediction horizon, and 24 executed actions per inference call.

\paragraph{Checkpoint selection and evaluation.}
We evaluate the final training checkpoint for each task--policy pair, at 100k steps for ACT and 300k steps for DiT VLA. For ACT and DiT VLA, results are reported as mean $\pm$ standard deviation across three evaluation seeds.

\section{Real-World Validation}
\label{appendix:real_world}

We evaluate ACT on a physical UR5 across connector insertion, 1-clip routing, and straight-channel seating. For each task, we compare four training-data combinations: real demonstrations (R), scripted simulation demonstrations (S), and simulation RL rollouts (P), combined as R, R+S, R+P, and R+S+P. Preliminary trials showed poor zero-shot performance for simulation-only policies, so we restrict the full hardware evaluation to data combinations containing real demonstrations. The comparison examines whether simulated trajectories supplement real demonstrations under shared task definitions and policy interfaces.

\subsection{Setup}
\label{appendix:real_setup}

\paragraph{Physical setup.}
The platform uses a UR5 arm with a Robotiq 2F-85 gripper and two RealSense D405 cameras providing side and wrist RGB views. The fixtures include a 3D-printed Ethernet-style connector and socket, a single clip, and a straight channel. Connector insertion and clip routing use a 6~mm-diameter HDMI cable, while channel seating uses a 10~mm-diameter green cord with red and yellow flecks. A force--torque sensor is mounted between the arm flange and gripper, but its readings are not used by ACT. We reproduce its end-effector offset in simulation and match the task-board appearance and camera poses to the physical setup. ACT observes RGB images and robot proprioception, but not depth and force/torque inputs, and executes action commands at 30~Hz.

\paragraph{Task initialization.}
Hardware trials use narrower initialization ranges than simulation. For insertion, the wire-side connector is placed within a $3\times3$~cm region with varying tilt, while the socket remains fixed. For routing, a single clip is placed on either side of the wire, $2$--$5$~cm away. For seating, one wire endpoint is fixed, while the other is placed $2$--$5$~cm from the channel in a direction perpendicular to its longitudinal axis.

\begin{figure*}[t]
    \centering
    \setlength{\tabcolsep}{2pt}
    \renewcommand{\arraystretch}{0.9}
    \normalfont\normalsize
    \begin{tabular}{@{}cccc@{}}
        Real (side) & Real (wrist) & Sim + noise (side) & Sim + noise (wrist) \\[3pt]
        \includegraphics[width=0.24\textwidth]{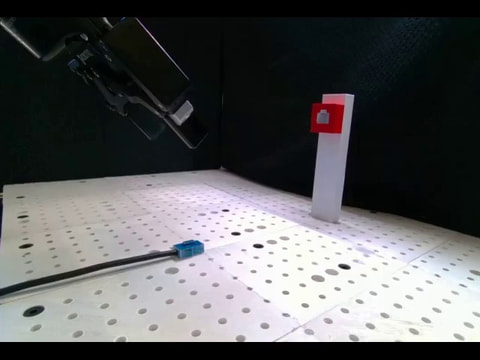} &
        \includegraphics[width=0.24\textwidth]{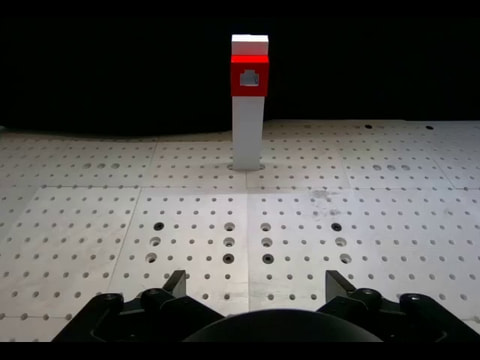} &
        \includegraphics[width=0.24\textwidth]{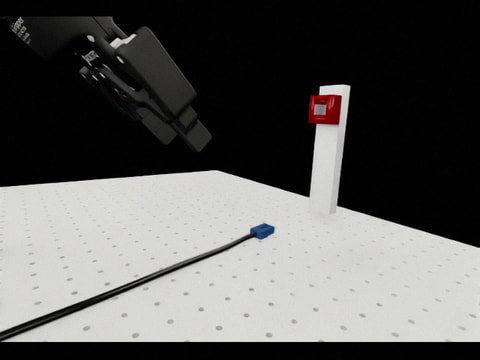} &
        \includegraphics[width=0.24\textwidth]{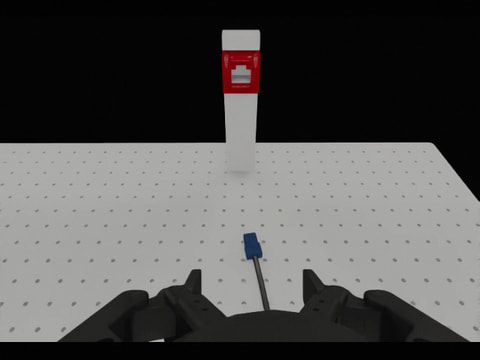} \\[2pt]
        \multicolumn{4}{c}{Connector insertion} \\[5pt]
        \includegraphics[width=0.24\textwidth]{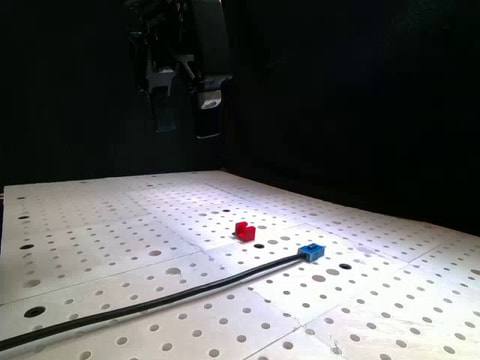} &
        \includegraphics[width=0.24\textwidth]{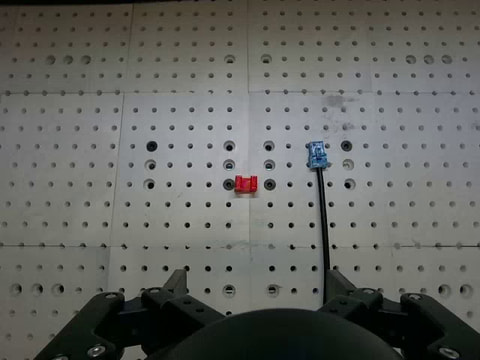} &
        \includegraphics[width=0.24\textwidth]{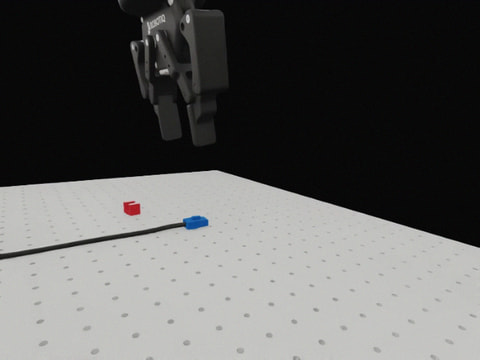} &
        \includegraphics[width=0.24\textwidth]{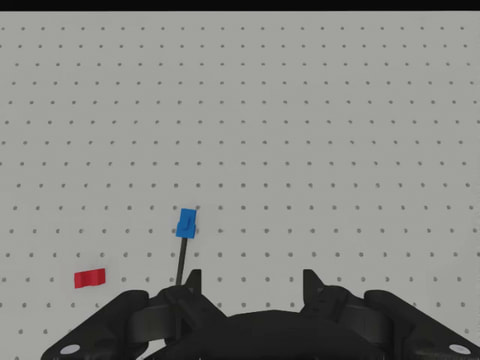} \\[2pt]
        \multicolumn{4}{c}{Clip routing} \\[5pt]
        \includegraphics[width=0.24\textwidth]{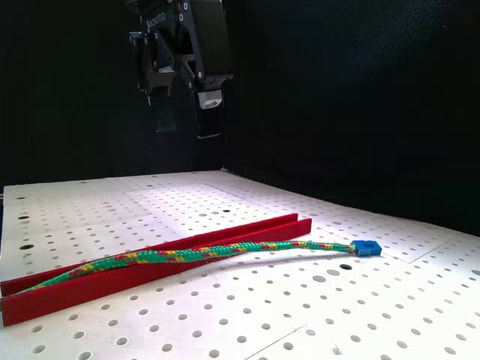} &
        \includegraphics[width=0.24\textwidth]{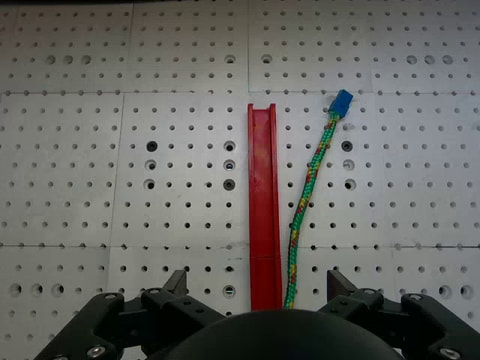} &
        \includegraphics[width=0.24\textwidth]{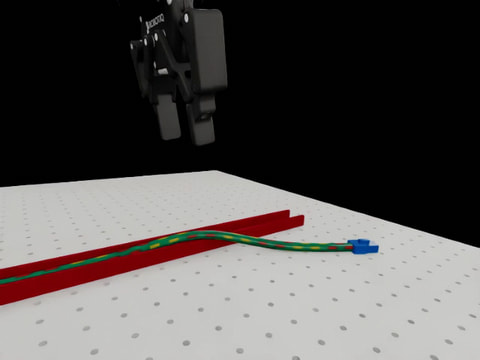} &
        \includegraphics[width=0.24\textwidth]{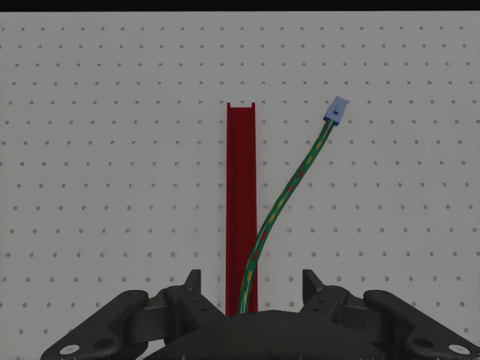} \\[2pt]
        \multicolumn{4}{c}{Channel seating} \\
    \end{tabular}
    \caption{\textbf{Real vs.\ noise-augmented simulation observations across three tasks.}
    Rows show connector insertion, clip routing, and channel seating; columns show the side and wrist views from real demonstrations and scripted simulation trajectories.
    Brightness, contrast, and Gaussian-noise augmentation are applied only to the simulated frames using the training preset.}
    \label{fig:real_sim_noise_comparison}
\end{figure*}

\paragraph{Bridging the sim-to-real visual gap.}
We match the simulated side and wrist views to the physical cameras and augment simulation images during training. Brightness and contrast factors are sampled independently from $[0.85,1.15]$, and Gaussian-noise standard deviation is sampled from $[0,0.015]$ for pixels in $[0,1]$. These augmentations apply only to simulation episodes; real images and evaluation inputs are not augmented. Figure~\ref{fig:real_sim_noise_comparison} shows observations from both domains across the three tasks.

\subsection{Protocol}
\label{appendix:real_protocol}

\paragraph{Data sources.}
Real demonstrations are collected on the physical UR5, while scripted demonstrations and RL rollouts are generated in the matched simulation environments. All sources share the same RGB, joint-state, and action schema. Table~\ref{tab:real_data_sources} reports the dataset sizes and training/validation episode counts. We hold out every tenth episode within each source, giving approximately 10\% validation episodes. The remaining episodes are used for training.

\begin{table}[h]
\centering
\small
\caption{Trajectory data for the three hardware tasks by source. Episodes and frames report total dataset sizes; train and validation columns report episode counts. All sources are recorded at $30$~FPS under the shared schema.}
\label{tab:real_data_sources}
\setlength{\tabcolsep}{6pt}
\renewcommand{\arraystretch}{1.1}
\begin{tabular}{llrrrr}
\toprule
Task & Source & Episodes & Frames & Train & Validation \\
\midrule
\multirow{3}{*}{Insertion}
& Real-world (R) & $50$ & $30{,}018$ & $45$ & $5$ \\
& Scripted (S) & $200$ & $121{,}890$ & $180$ & $20$ \\
& RL rollout (P) & $200$ & $27{,}394$ & $180$ & $20$ \\
\midrule
\multirow{3}{*}{1-clip routing}
& Real-world (R) & $51$ & $30{,}634$ & $46$ & $5$ \\
& Scripted (S) & $200$ & $125{,}304$ & $180$ & $20$ \\
& RL rollout (P) & $200$ & $53{,}226$ & $180$ & $20$ \\
\midrule
\multirow{3}{*}{Straight-channel seating}
& Real-world (R) & $50$ & $30{,}033$ & $45$ & $5$ \\
& Scripted (S) & $200$ & $114{,}954$ & $180$ & $20$ \\
& RL rollout (P) & $200$ & $56{,}511$ & $180$ & $20$ \\
\bottomrule
\end{tabular}
\end{table}

\paragraph{Training.}
All configurations train ACT for 80k gradient steps with batch size 16, learning rate $10^{-5}$, and training seed 1000. ACT uses an ImageNet-initialized ResNet-18 visual backbone, predicts 100-action chunks, and executes 50 actions per inference call. Inputs comprise two RGB views and a seven-dimensional joint/gripper state; outputs are six arm-joint targets and one gripper target.

\paragraph{Checkpoint selection.}
Checkpoints are saved every 5k steps and selected by the lowest masked L1 action-prediction loss over the predicted chunks in normalized action space. Selection uses only the five held-out real episodes for each task, including for mixed-source policies. In the order R, R+S, R+P, and R+S+P, selected checkpoints are $40/80/55/80$k for insertion, $55/80/80/70$k for routing, and $80/80/75/80$k for seating. Thus, configurations share the training budget and selection rule, but not the selected training step. Normalization statistics for the real-only datasets include the held-out episodes.

\subsection{Results and Analysis}
\label{appendix:real_results}

Representative failure modes are shown in Figure~\ref{fig:real_failure_cases}.

\paragraph{Connector insertion.}
Observed failures include unsuccessful pickup and reaching without seating. The gripper sometimes closes beside the connector or fails to secure it during lifting. After successful pickup, the connector may reach the socket but remain unseated because of its position or orientation within the grasp. A grasp sufficient for transport may therefore be unsuitable for insertion.

\paragraph{Clip routing.}
The wire may be released beside the clip opening or rest on the clip rim without entering the slot. In these cases, the gripper reaches the fixture, but the wire remains incorrectly positioned relative to the opening.

\paragraph{Channel seating.}
Missed grasps leave the free endpoint on the board. After successful pickup, moving the endpoint near the channel can still leave part of the wire outside it; successful seating requires placing the wire body within the channel.

The characteristics of the three trajectory sources are summarized in Table~\ref{tab:real_world_qualitative}.

\begin{table}[h]
\centering
\small
\caption{Characteristics of the three trajectory sources used for hardware policy training.}
\label{tab:real_world_qualitative}
\setlength{\tabcolsep}{6pt}
\renewcommand{\arraystretch}{1.35}
\begin{tabularx}{\linewidth}{l X}
\toprule
\textbf{Source} & \textbf{Trajectory characteristics} \\
\midrule
R & Human demonstrations contain pauses and repeated fine adjustments during alignment. Their limited diversity can restrict generalization. \\
S & Scripted trajectories use action noise during generation to broaden motion coverage. Execution is relatively slow and regular, providing varied examples of controlled task completion. \\
P & Task-optimized RL rollouts produce fast, aggressive motions. For this dataset collection, execution speed was reduced to 50\% of the original policy speed. \\
\bottomrule
\end{tabularx}
\end{table}

\begin{figure*}[t]
    \centering
    \setlength{\tabcolsep}{2pt}
    \renewcommand{\arraystretch}{1.0}
    \normalfont\normalsize
    \begin{tabular}{@{}ccc@{}}
    Connector insertion & Clip routing & Channel seating \\[3pt]
    \includegraphics[width=0.324\textwidth]{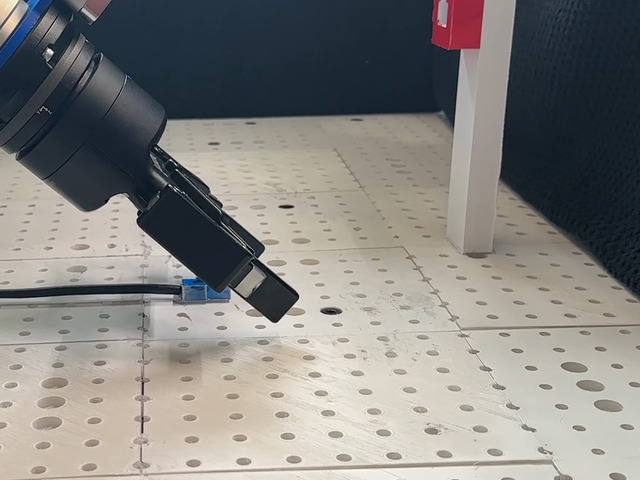} &
    \includegraphics[width=0.324\textwidth]{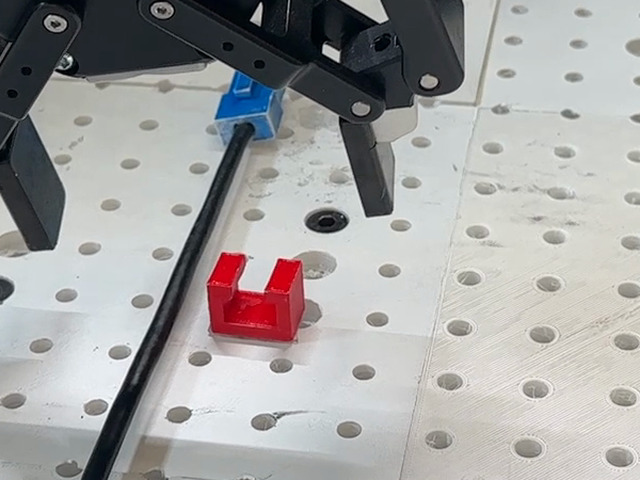} &
    \includegraphics[width=0.324\textwidth]{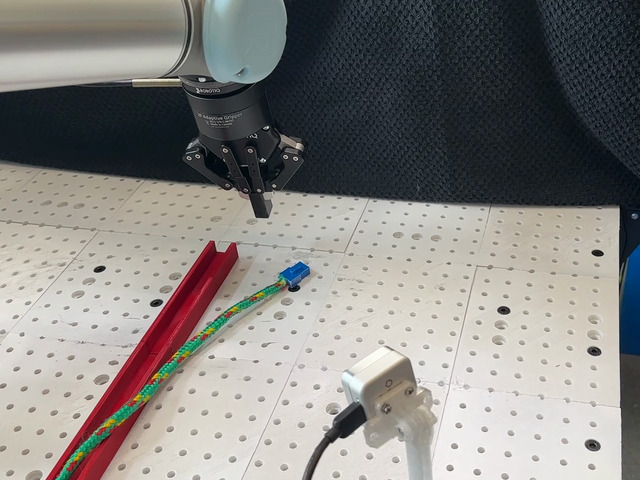} \\[2pt]
    (a) Missed grasp & (b) Wire beside clip & (c) Missed grasp \\[5pt]
    \includegraphics[width=0.324\textwidth]{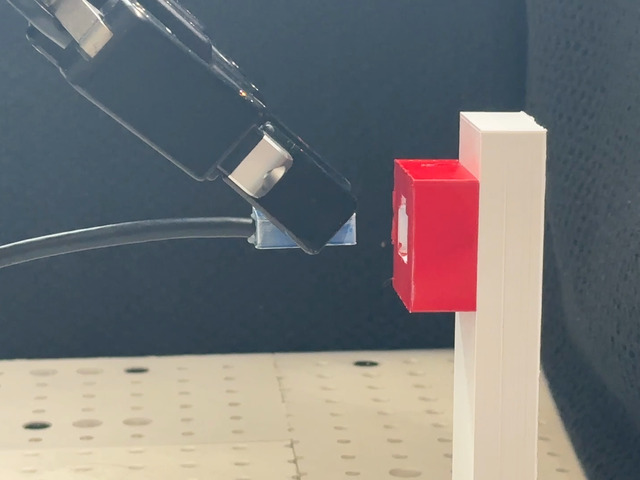} &
    \includegraphics[width=0.324\textwidth]{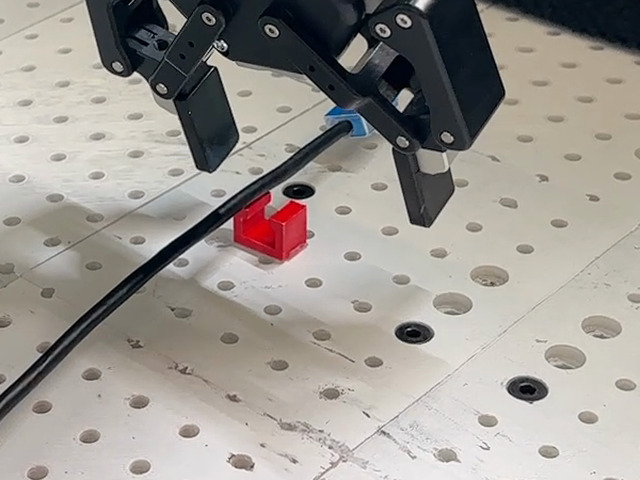} &
    \includegraphics[width=0.324\textwidth]{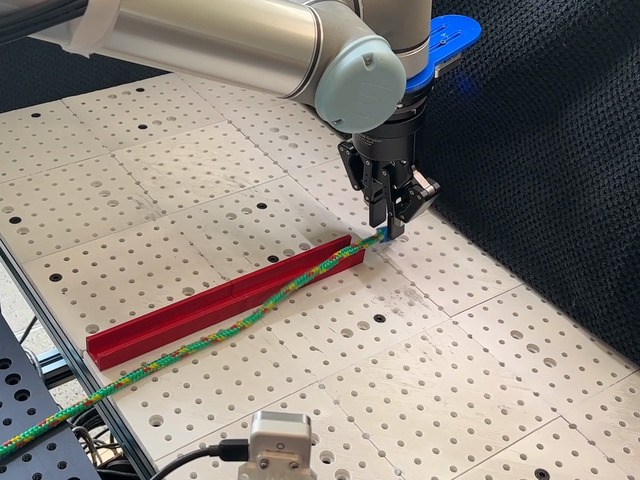} \\[2pt]
    (d) Unseated connector & (e) Wire on clip rim & (f) Wire outside channel \\
    \end{tabular}
    \caption{\textbf{Representative hardware failures in insertion (left), clip routing (middle), and channel seating (right).}
    Panels show (a,c) missed grasps, (b) wire misplacement beside the clip, (d) reaching without connector seating, (e) wire resting on the clip rim, and (f) wire placement outside the channel.}
    \label{fig:real_failure_cases}
\end{figure*}

\paragraph{Limitations.}
The hardware study covers three tasks on one UR5 with restricted initialization ranges and 20 trials per configuration. Indistinguishable difference in count of successful trials cannot lead to a definitive ranking of data mixtures. Source composition and dataset size vary together, so these comparisons measure data supplementation without isolating source diversity from data quantity.

\section{Teleoperation Data Collection}
\label{appendix:teleoperation}

WireCraft teleoperation collects demonstrations by having a human operator drive
the UR5 end-effector in Isaac Sim through a 6-DoF input device, with episodes
recorded as supervised trajectories for IL and VLA training. Operator input is
mapped to an end-effector delta pose plus a binary gripper command,
\[
a_{\mathrm{cart}} = [\Delta x, \Delta y, \Delta z, \Delta r_x, \Delta r_y, \Delta r_z, g],
\]
which is resolved by differential IK into a UR5 joint-position target. Trajectories
are stored under the same observation--action schema as scripted demonstrations,
so the two sources can be mixed without modifying the policy interface. The key
behavioral difference is that scripted policies produce smooth, phase-structured
rollouts from privileged simulation state, whereas teleoperation produces
visually conditioned rollouts that include pauses, corrections, and re-grasps
useful for contact-rich DLO manipulation.

\textbf{Control interfaces.}
We support three input modes, all sharing the same Cartesian-delta action
interface: (i) a 3Dconnexion SpaceMouse with on-screen camera views (default);
(ii) Meta Quest VR with retargeted wrist motion and a pinch-to-grip gesture,
useful when fixed cameras provide poor depth cues; and (iii) a combined mode
using the SpaceMouse for control and the VR headset for visualization. All
modes record only simulator observations and IK-resolved joint targets; raw
button events and hand-tracking signals are not stored.

\textbf{Recorded schema.} The teleoperation pipeline follows the shared WireCraft trajectory schema, storing synchronized third-person and wrist RGB streams (\texttt{observation.images.cam\_side2}, \texttt{observation.images.cam\_color}), UR5 joint and gripper state (\texttt{observation.state}), end-effector pose (\texttt{observation.cartesian\_state}), the applied joint-position target (\texttt{action}), \texttt{timestamp}, \texttt{episode\_index}, and a \texttt{task} label. The recorded \texttt{action} is the IK-resolved joint target rather than the raw operator Cartesian delta, allowing teleoperated trajectories to use the same action representation as scripted and policy-generated data. In this paper, we release the teleoperation interface and recording pipeline, but do not include simulated teleoperation trajectories in the reported training datasets. Systematic teleoperation-data collection, filtering, and use for policy training are left to future work.

\begin{figure*}[t]
    \centering
    \setlength{\tabcolsep}{2pt}
    \renewcommand{\arraystretch}{0.85}
    \begin{tabular}{@{}c@{\hspace{2pt}}c@{\hspace{2pt}}c@{\hspace{2pt}}c@{}}
    \begin{overpic}[width=0.24\textwidth]{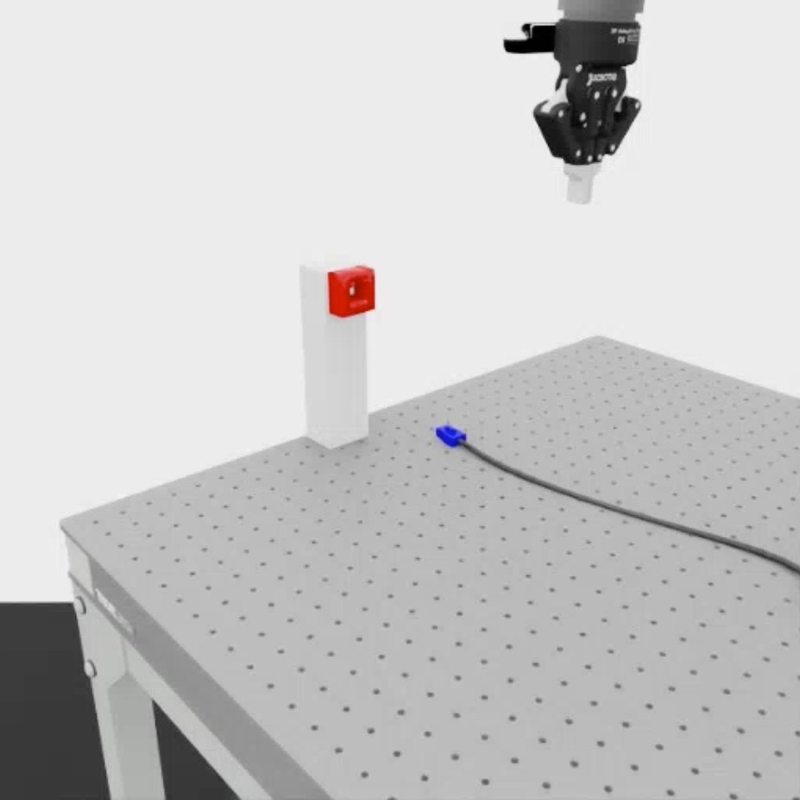}
        \put(3,3){\small\textbf{\textcolor{white}{(1) Initialize}}}
    \end{overpic} &
    \begin{overpic}[width=0.24\textwidth]{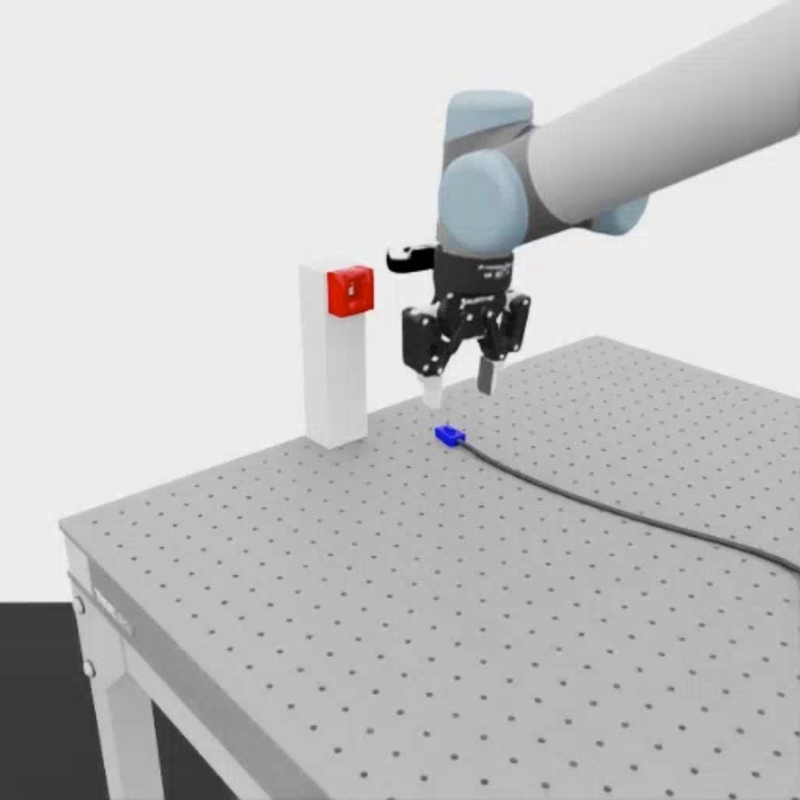}
        \put(3,3){\small\textbf{\textcolor{white}{(2) Approach}}}
    \end{overpic} &
    \begin{overpic}[width=0.24\textwidth]{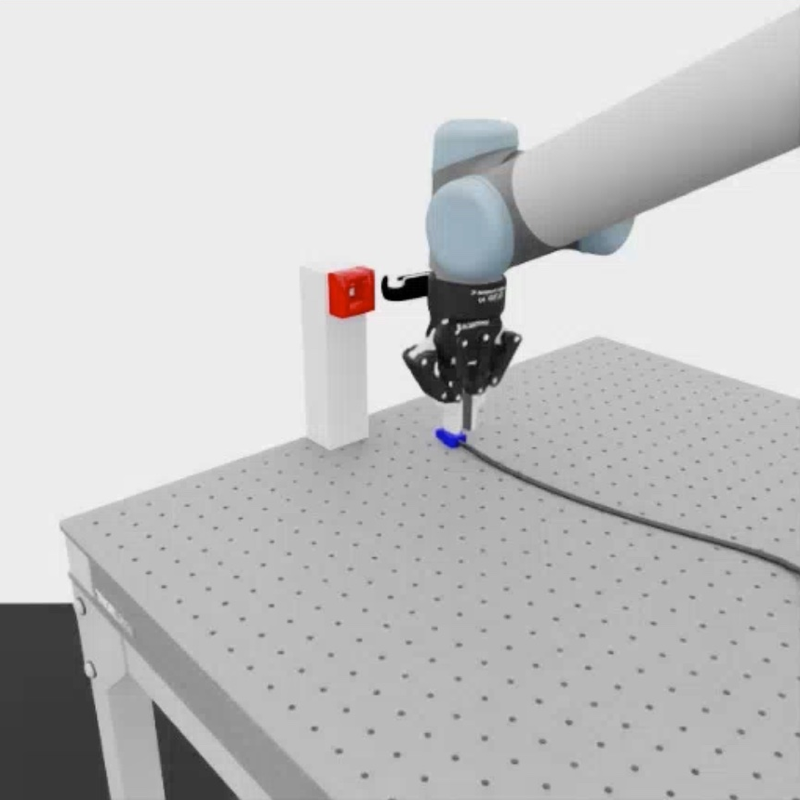}
        \put(3,3){\small\textbf{\textcolor{white}{(3) Align grasp}}}
    \end{overpic} &
    \begin{overpic}[width=0.24\textwidth]{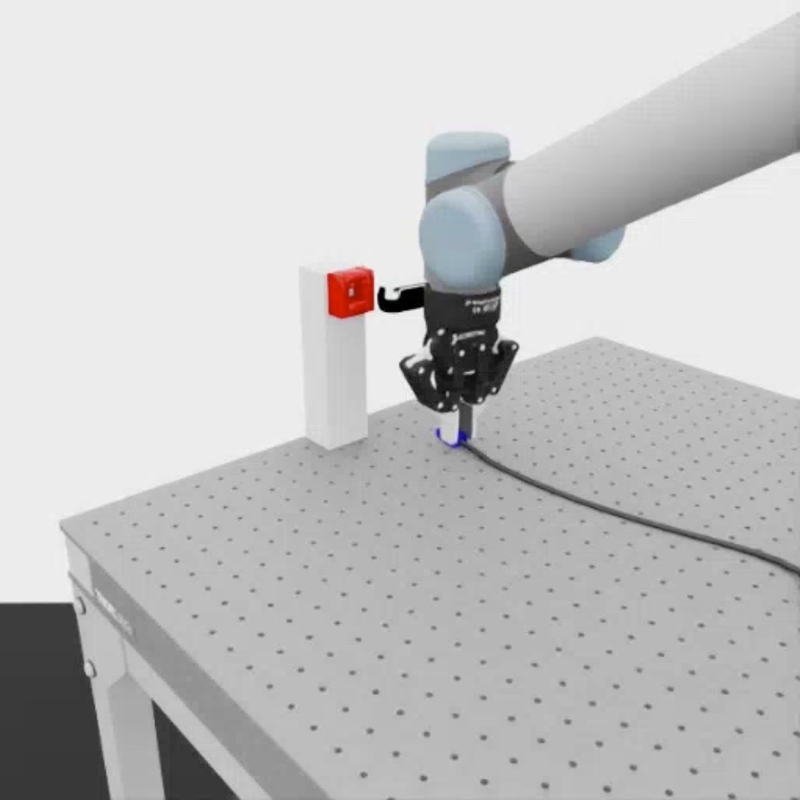}
        \put(3,3){\small\textbf{\textcolor{white}{(4) Grasp}}}
    \end{overpic}
    \\[2pt]
    \begin{overpic}[width=0.24\textwidth]{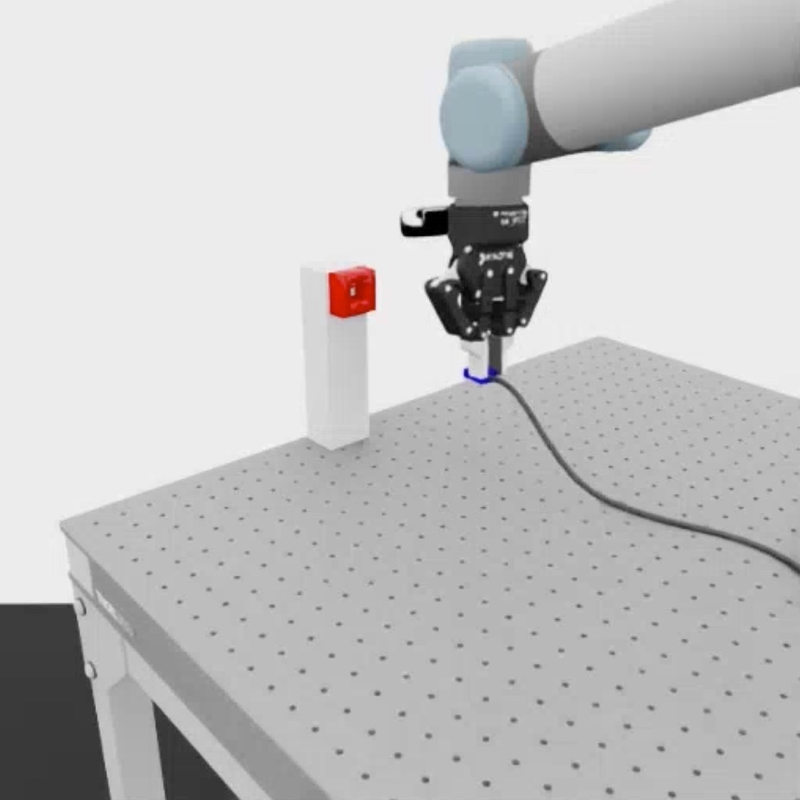}
        \put(3,3){\small\textbf{\textcolor{white}{(5) Lift}}}
    \end{overpic} &
    \begin{overpic}[width=0.24\textwidth]{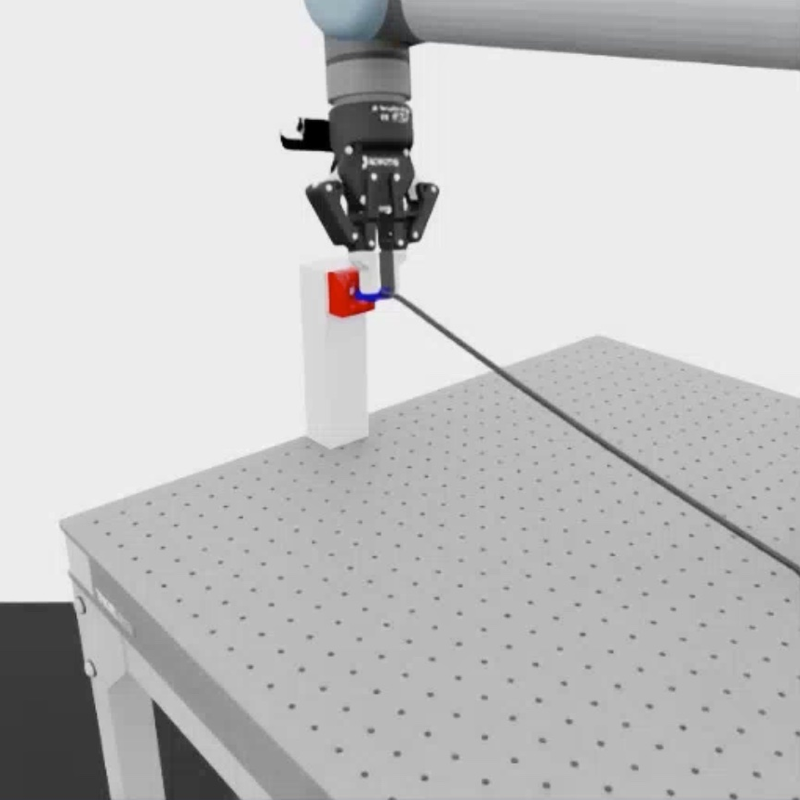}
        \put(3,3){\small\textbf{\textcolor{white}{(6) Move to socket}}}
    \end{overpic} &
    \begin{overpic}[width=0.24\textwidth]{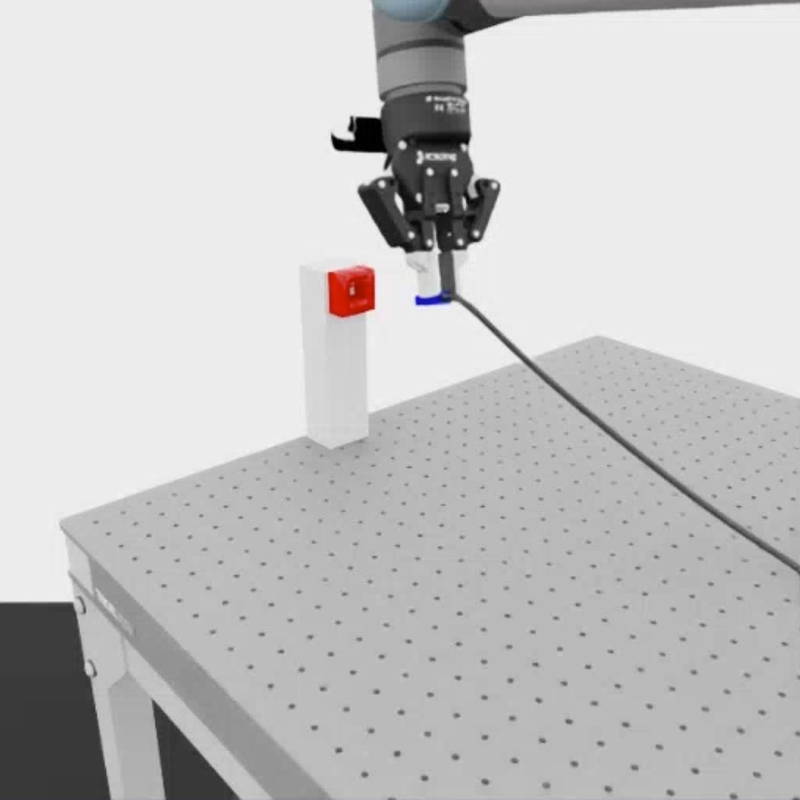}
        \put(3,3){\small\textbf{\textcolor{white}{(7) Align}}}
    \end{overpic} &
    \begin{overpic}[width=0.24\textwidth]{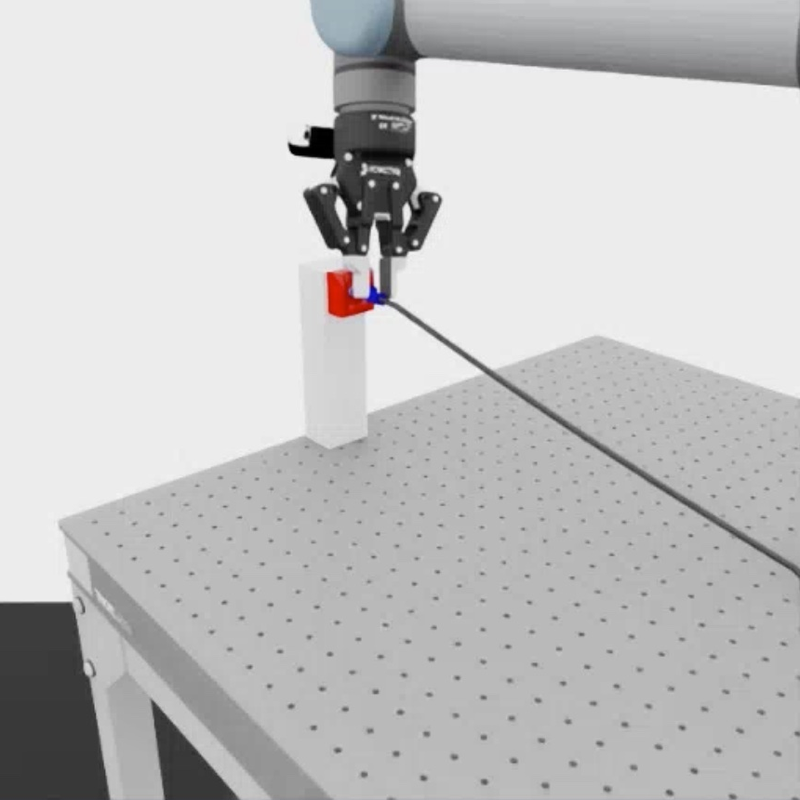}
        \put(3,3){\small\textbf{\textcolor{white}{(8) Insert}}}
    \end{overpic}
    \\
    \end{tabular}
    \caption{\textbf{A teleoperated connector-insertion trajectory in simulation.} A human operator drives the UR5 end-effector through eight representative stages: (1) initialization with the connector resting on the table; (2) approaching the connector; (3) aligning the gripper above it; (4) grasping the connector, typically after several attempts, since the operator must re-position and re-close the gripper to secure a firm grasp; (5) lifting the connector clear of the surface; (6) moving toward the socket while pulling the trailing wire taut to reduce its disturbance during insertion; (7) aligning the connector with the socket opening; and (8) seating it, often requiring repeated insertion attempts to correct small misalignments. Because the operator has no haptic feedback and only limited visual cues for fine alignment, both grasping and insertion rely on trial-and-error retries, making teleoperated episodes substantially longer than scripted or policy rollouts.}
    \label{fig:teleop_sequence}
\end{figure*}

\end{document}